\documentclass[10pt,twocolumn,letterpaper]{article}

\usepackage{cvpr}
\usepackage{times}
\usepackage{epsfig}
\usepackage{graphicx}
\usepackage{amsmath}
\usepackage{amssymb}
\usepackage{graphics}
\usepackage{pifont}
\usepackage{xcolor}
\usepackage{booktabs}
\usepackage{multirow}
\usepackage[font=normalsize]{subfig}
\usepackage{gensymb}

\usepackage[pagebackref=true,breaklinks=true,letterpaper=true,colorlinks,bookmarks=false]{hyperref}

\cvprfinalcopy 

\newcommand{\ourmodel}{UPSNet}

\ifcvprfinal\fi
\begin{document}

\title{UPSNet: A Unified Panoptic Segmentation Network}

\author{Yuwen Xiong$^{1, 2}$\thanks{Equal contribution.} \quad Renjie Liao$^{1,2*}$ \quad Hengshuang Zhao$^{3*}$\thanks{This work was done when Hengshuang Zhao
was an intern at Uber ATG.}, \\Rui Hu$^{1}$ \quad  Min Bai$^{1, 2}$ \quad  Ersin Yumer$^{1}$ \quad  Raquel Urtasun$^{1, 2}$\\
$^1$Uber ATG\quad $^2$University of Toronto \quad $^3$The Chinese University of Hong Kong\\
{\small \texttt{\{yuwen, rjliao, rui.hu, mbai3, yumer, urtasun\}@uber.com }}\\
{\small \texttt{hszhao@cse.cuhk.edu.hk}}
\\
}

\maketitle


\begin{abstract}
In this paper, we propose a unified panoptic segmentation network (UPSNet) for tackling the newly proposed panoptic segmentation task.
On top of a single backbone residual network, we first design a deformable convolution based semantic segmentation head and a Mask R-CNN style instance segmentation head which solve these two subtasks simultaneously.
More importantly, we introduce a parameter-free panoptic head which solves the panoptic segmentation via pixel-wise classification.
It first leverages the logits from the previous two heads and then innovatively expands the representation for enabling prediction of an extra unknown class which helps better resolve the conflicts between semantic and instance segmentation.
Additionally, it handles the challenge caused by the varying number of instances and permits back propagation to the bottom modules in an end-to-end manner.
Extensive experimental results on Cityscapes, COCO and our internal dataset demonstrate that our UPSNet achieves state-of-the-art performance with much faster inference. Code has been made available at: \url{https://github.com/uber-research/UPSNet}.
\end{abstract}

\section{Introduction}

Relying on the advances in deep learning, computer vision systems have been substantially improved, especially in tasks such as semantic segmentation~\cite{zhao2017pyramid} and instance segmentation~\cite{he2017mask}.
The former focuses on segmenting amorphous image regions which share similar texture or material such as grass, sky and road, whereas the latter focuses on segmenting countable objects such as people, bicycle and car.
Since both tasks aim at understanding the visual scene at the pixel level, a shared model or representation could arguably be beneficial.
However, the dichotomy of these two tasks lead to very different modeling strategies despite the inherent connections between them.
For example, fully convolutional neural networks~\cite{long2015fully} are often adopted for semantic segmentation while proposal based detectors~\cite{ren2015faster} are frequently exploited for instance segmentation.

As an effort to leverage the possible complementariness of these two tasks and push the segmentation systems further towards real-world application, Kirillov~\etal~\cite{kirillov2018panoptic} unified them and proposed the so-called \textit{panoptic segmentation} task.
It is interesting to note that tasks with the same spirit have been studied under various names before deep learning became popular. 
Notable ones include image parsing~\cite{tu2005image}, scene parsing~\cite{tu2005image} and holistic scene understanding~\cite{yao2012describing}.
In panoptic segmentation, countable objects (those that map to instance segmentation tasks well) are called \textit{things} whereas amorphous and uncountable regions (those that map to semantic segmentation tasks better) are called \textit{stuff}.
For any pixel, if it belongs to \textit{stuff}, the goal of a panoptic segmentation system is simply to predict its class label within the \textit{stuff} classes.
Otherwise the system needs to decide which instance it belongs to as well as which \textit{thing} class it belongs to.
The challenge of this task lies in the fact that the system has to give a unique answer for each pixel.

In this paper, we propose a unified panoptic segmentation network ({\ourmodel}) to approach the panoptic segmentation problem as stated above.
Unlike previous methods~\cite{kirillov2018panoptic,li2018weakly} which have two separate branches designed for semantic and instance segmentation individually, our model exploits a single network as backbone to provide shared representations.
We then design two heads on top of the backbone for solving these tasks simultaneously.
Our semantic head builds upon deformable convolution~\cite{dai2017deformable} and leverages multi-scale information from feature pyramid networks (FPN)~\cite{lin2017feature}.
Our instance head follows the Mask R-CNN~\cite{he2017mask} design and outputs mask segmentation, bounding box and its associated class.
As shown in experiments, these two lightweight heads along with the single backbone provide good semantic and instance segmentations which are comparable to separate models.
More importantly, we design a panoptic head which predicts the final panoptic segmentation via pixel-wise classification of which the number of classes per image could vary.
It exploits the logits from the above two heads and adds a new channel of logits corresponding to an extra \textit{unknown} class. 
By doing so, it provides a better way of resolving the conflicts between semantic and instance segmentation.
Moreover, our parameter-free panoptic head is very lightweight and could be used with various backbone networks.
It facilitates end-to-end training which is not the case for previous methods~\cite{kirillov2018panoptic,li2018weakly}.
To verify the effectiveness of our {\ourmodel}, we perform extensive experiments on two public datasets: Cityscapes~\cite{cordts2016cityscapes} and COCO~\cite{lin2014microsoft}.
Furthermore, we test it on our internal dataset which is similar in spirit to Cityscapes (\ie, images are captured from ego-centric driving scenarios) but with significantly larger ($\approx 3\times$) size.
Results on these three datasets manifest that our {\ourmodel} achieves state-of-the-art performances and enjoys much faster inference compared to recent competitors.


\section{Related Work}

\textbf{Semantic Segmentation:}
Semantic segmentation is one of the fundamental computer vision tasks that has a long history. 
Earlier work~\cite{everingham2010pascal,mottaghi2014role} focused on introducing datasets for this task and showed the importance of global context by demonstrating the gains in bayesian frameworks, whether structured or free-form. 
Recent semantic segmentation methods that exploit deep convolutional feature extraction mainly approach this problem from either multi-scale feature aggregation~\cite{zhao2017pyramid,long2015fully,chen2016attention,hariharan2015hypercolumns}, or end-to-end structured prediction~\cite{chen2014semantic,zheng2015conditional,arnab2016higher,liu2015semantic,chen2017rethinking} perspectives. 
As context is crucial for semantic segmentation, one notable improvement to most convolutional models emerged from dilated convolutions~\cite{yu2016multiscale,yu2017dilated} which allows for a larger receptive field without the need of more free parameters. 
Pyramid scene parsing network (PSPNet)~\cite{zhao2017pyramid} that uses dilated convolutions in its backbone, and its faster variant~\cite{zhao2018icnet} for real-time applications are widely utilized in practical applications.  
Based on FPN and PSPNet, a multi-task framework is proposed in~\cite{xiao2018unified} and demonstrated to be versatile in segmenting a wide range of visual concepts. 

\textbf{Instance Segmentation:}
Instance segmentation deals not only with identifying the semantic class a pixel is associated with, but also the specific object instance that it belongs to. 
Beginning with the introduction of region-based CNN (R-CNN)~\cite{girshick2014rich}, many early deep learning approaches to instance segmentation attacked the problem by casting the solution to instance segmentation as a two stage approach where a number of segment proposals are made, which is then followed by a voting between those proposals to choose the best~\cite{uijlings2013selective,dai2016instance,dai2015convolutional,hariharan2014simultaneous,hariharan2015hypercolumns,ren17recattend}.
The common denominator for these methods is that the segmentation comes before classification, and are therefore slower. 
Li~\etal~\cite{li2017fully} proposed a fully convolutional instance-aware segmentation method, where instance mask proposals~\cite{dai2016instance} are married with fully convolutional networks~\cite{long2015fully}. 
Most recently, Mask R-CNN~\cite{he2017mask} introduced a joint approach to both mask prediction and recognition where one of the two parallel heads are dedicated to each task.

\textbf{Panoptic Segmentation:}
Instance segmentation methods that focus on detection bounding box proposals, as mentioned above, ignore the classes that are not well suited for detection, \eg, sky, street. 
On the other hand, semantic segmentation does not provide instance boundaries for classes like pedestrian and bicycle in a given image. 
Panoptic segmentation task, first coined by Kirillov~\etal~\cite{kirillov2018panoptic} unifies these tasks and defines an ideal output for \textit{thing} classes as instance segmentations, as well as for \textit{stuff} classes as semantic segmentation. 
The baseline panoptic segmentation method introduced in~\cite{kirillov2018panoptic} processes the input independently for semantic segmentation via a PSPNet, and for instance segmentation utilizing a Mask R-CNN~\cite{he2017mask}, followed by simple heuristic decisions to produce a single void, stuff, or thing instance label per pixel. 
Recently, Li~\etal~\cite{li2018weakly} introduced a weakly- and semi-supervised panoptic segmentation method where they relieve some of the ground truth constraints by supervising thing classes using bounding boxes, and stuff classes by utilizing image level tags. 
De Gaus~\etal~\cite{de2018panoptic} uses a single feature extraction backbone for the pyramid semantic segmentation head~\cite{zhao2017pyramid}, and the instance segmentation head~\cite{he2017mask}, followed by heuristics for merging pixel level annotations, effectively introducing an end-to-end version of~\cite{kirillov2018panoptic} due to the shared backbone for the two task networks.
Li~\etal~\cite{li2018attention} propose the attention-guided unified network (AUNet) which leverages proposal and mask level attention to better segment the background. 
Similar post-processing heuristics as in~\cite{kirillov2018panoptic} are used to generate the final panoptic segmentation. 
Li~\etal~\cite{li2018learning} propose things and stuff consistency network (TASCNet) which constructs a binary mask predicting things vs. stuff for each pixel. 
An extra loss is added to enforce the consistency between things and stuff prediction.

In contrast to most of these methods, we use a single backbone network to provide both semantic and instance segmentation results. 
More importantly, we develop a simple yet effective panoptic head which helps accurately predict the instance and class label. 


\begin{figure*}[t]
    \begin{center}
        \includegraphics[width=0.9\linewidth]{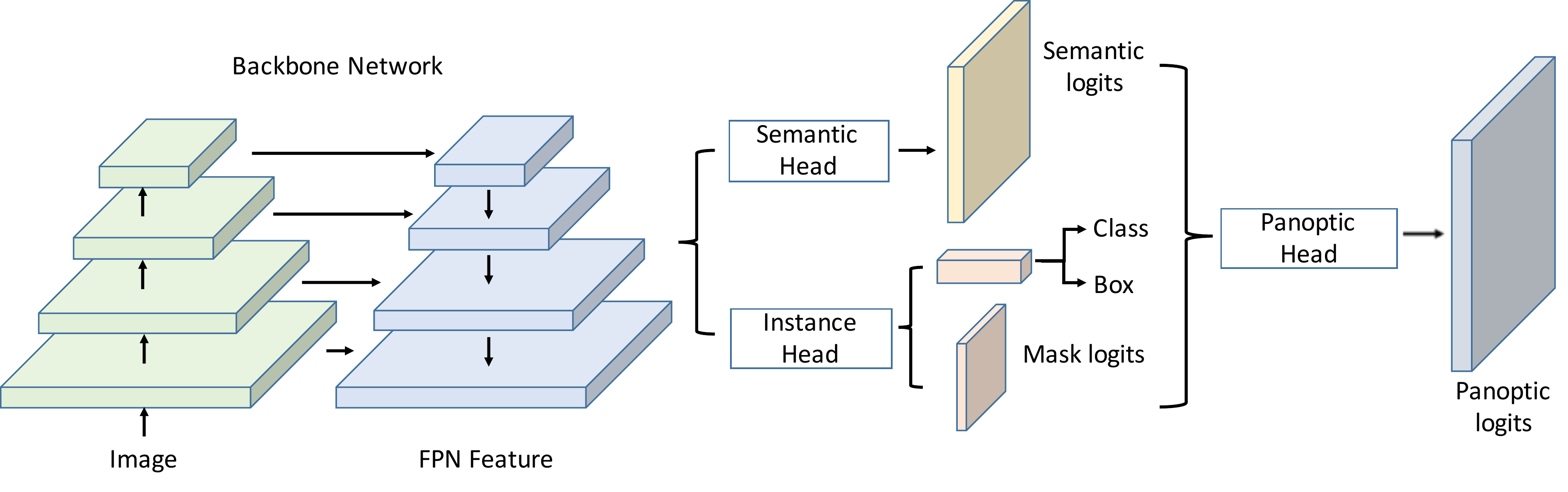}
    \end{center}
    \vspace{-0.5cm}
    \caption{Overall architecture of our {\ourmodel}.}
    \vspace{-0.5cm}
    \label{fig:model}
\end{figure*}

\section{Unified Panoptic Sementation Network}\label{sect:model}

In this section, we first introduce our model and then explain the implementation details.
Following the convention of~\cite{kirillov2018panoptic}, we divide the semantic class labels into \textit{stuff} and \textit{thing}. 
Specifically, \textit{thing} refers to the set of labels of instances (\eg pedestrian, bicycle), whereas \textit{stuff} refers to the rest of the labels that represent semantics without clear instance boundaries (\eg street, sky).
We denote the number of \textit{stuff} and \textit{thing} classes as $N_{\text{stuff}}$ and $N_{\text{thing}}$ respectively.

\subsection{{\ourmodel} Architecture}\label{sect:model_arch}

\ourmodel~consists of a shared convolutional feature extraction backbone and multiple heads on top of it.
Each head is a sub-network which leverages the features from the backbone and serves a specific design purpose that is explained in further detail below. 
The overall model architecture is shown in Fig.~\ref{fig:model}.

\vspace{-0.3cm}
\paragraph{Backbone:}

We adopt the original Mask R-CNN~\cite{he2017mask} backbone as our convolutional feature extraction network.
This backbone exploits a deep residual network (ResNet)~\cite{he2016deep} with a feature pyramid network (FPN)~\cite{lin2017feature}.

\vspace{-0.3cm}
\paragraph{Instance Segmentation Head:}

The instance segmentation head follows the Mask R-CNN design with a bounding box regression output, a classification output, and a segmentation mask output. 
The goal of the instance head is to produce instance aware representations that could identify \textit{thing} classes better.
Ultimately these representations are passed to the panoptic head to contribute to the logits for each instance. 

\vspace{-0.3cm}
\paragraph{Semantic Segmentation Head:}

\begin{figure}
    \begin{center}
        \includegraphics[width=0.99\linewidth]{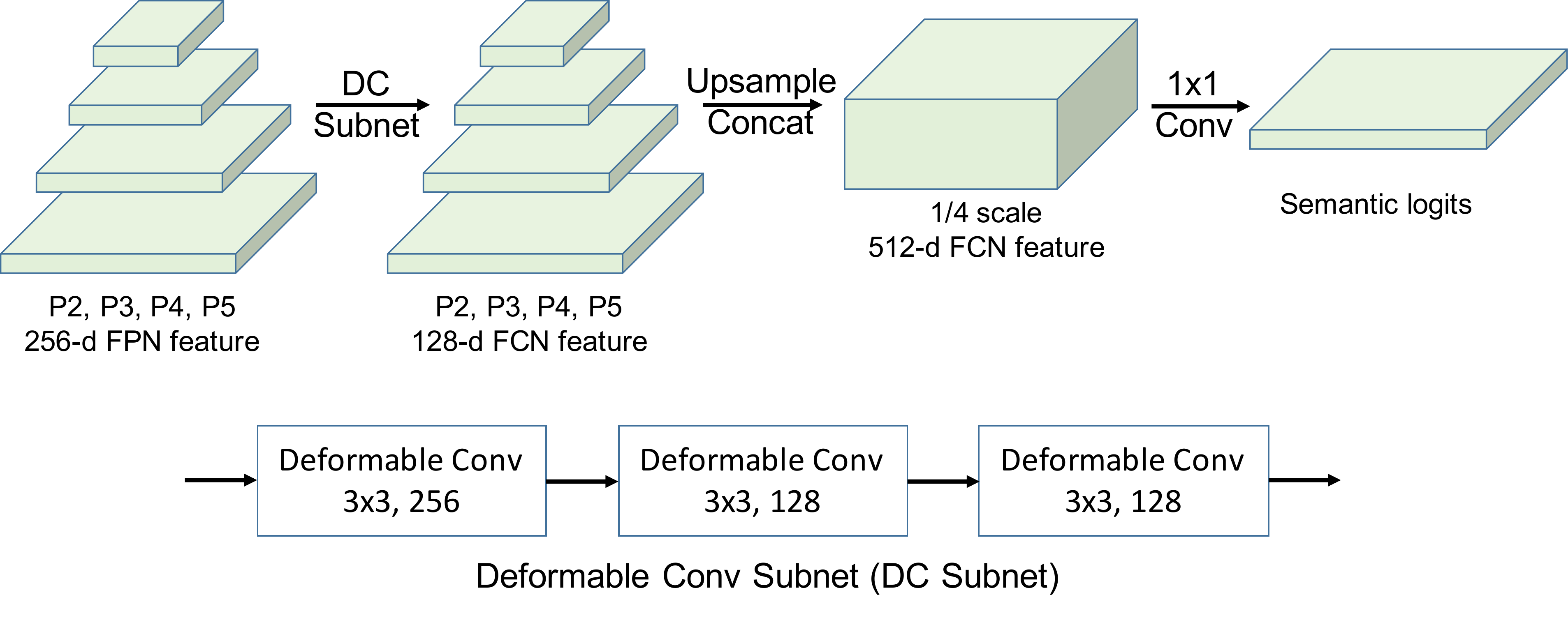}
    \end{center}
    \vspace{-0.5cm}
    \caption{Architecture of our semantic segmentation head.}
    \vspace{-0.5cm}
    \label{fig:semantic_head}
\end{figure}

The goal of the semantic segmentation head is to segment all semantic classes without discriminating instances.
It could help improving instance segmentation where it achieves good results of \textit{thing} classes.
Our semantic head consists of a deformable convolution~\cite{dai2017deformable} based sub-network which takes the multi-scale feature from FPN as input. 
In particular, we use $P_2$, $P_3$, $P_4$ and $P_5$ feature maps of FPN which contain $256$ channels and are $1/4$, $1/8$, $1/16$ and $1/32$ of the original scale respectively.
These feature maps first go through the same deformable convolution network independently and are subsequently upsampled to the $1/4$ scale.
We then concatenate them and apply $1 \times 1$ convolutions with softmax to predict the semantic class.
The architecture is shown in Fig.~\ref{fig:semantic_head}.
As will be experimentally verified later, the deformable convolution along with the multi-scale feature concatenation provide semantic segmentation results as good as a separate model, \eg, a PSPNet adopted in~\cite{kirillov2018panoptic}.
Semantic segmentation head is associated with the regular pixel-wise cross entropy loss.
To put more emphasis on the foreground objects such as pedestrians, we also incorporate a RoI loss.
During training, we use the ground truth bounding box of the instance to crop the logits map after the $1 \times 1$ convolution and then resize it to $28 \times 28$ following Mask R-CNN.
The RoI loss is then the cross entropy computed over $28 \times 28$ patch which amounts to penalizing more on the pixels within instances for incorrect classification.
As demonstrated in the ablation study later, we empirically found that this RoI loss helps improve the performance of panoptic segmentation without harming the semantic segmentation.

\vspace{-0.3cm}
\paragraph{Panoptic Segmentation Head:}

Given the semantic and instance segmentation results from the above described two heads, we combine their outputs (specifically per pixel logits) in the panoptic segmentation head. 

The logits from semantic head is denoted as $X$ of which the channel size, height and width are $N_{\text{stuff}} + N_{\text{thing}}$, $H$ and $W$ respectively.
$X$ can then be divided along channel dimension into two tensors $X_{\text{stuff}}$ and $X_{\text{thing}}$ which are logits corresponding to \textit{stuff} and \textit{thing} classes.
For any image, we determine the number of instances $N_{\text{inst}}$ according to the number of ground truth instances during training. 
During inference, we rely on a mask pruning process to determine $N_{\text{inst}}$ which is explained in Section ~\ref{sect:implement_detail}.
$N_{\text{stuff}}$ is fixed since number of \textit{stuff} classes is constant across different images, whereas $N_{\text{inst}}$ is not constant since the number of instances per image can be different.
The goal of our panoptic segmentation head is to first produce a logit tensor $Z$ which is of size $(N_{\text{stuff}} + N_{\text{inst}}) \times H \times W$ and then uniquely determine both the class and instance ID for each pixel.

\begin{figure}
    \begin{center}
        \includegraphics[width=0.99\linewidth]{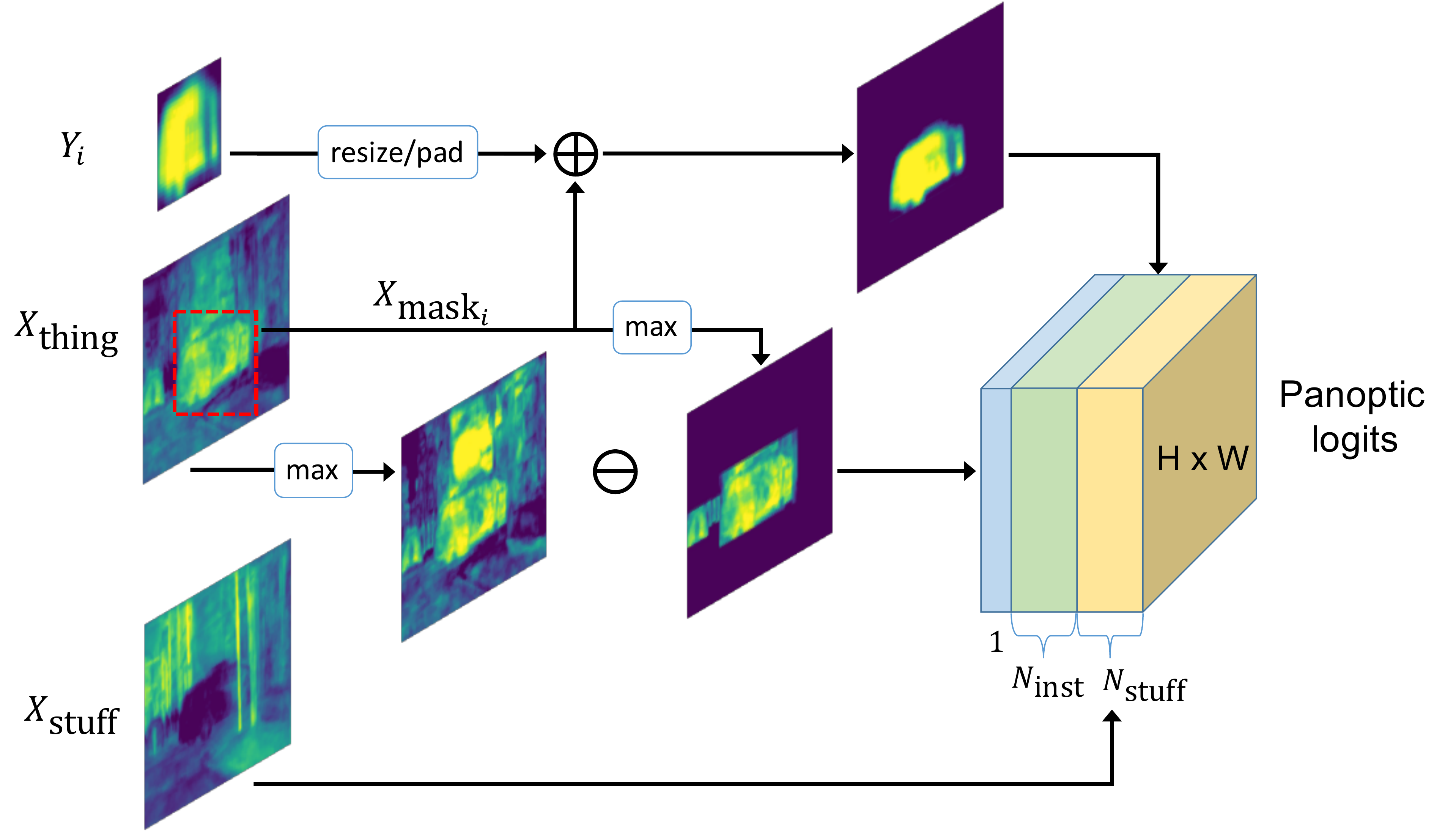}
    \end{center}
    \vspace{-0.5cm}
    \caption{Architecture of our panoptic segmentation head.}
    \vspace{-0.5cm}
    \label{fig:panoptic_head}
\end{figure}

We first assign $X_{\text{stuff}}$ to the first $N_{\text{stuff}}$ channels of $Z$ to provide the logits for classifying \textit{stuffs}. 
For any instance $i$, we have its mask logits $Y_i$ from the instance segmentation head which is of size $28 \times 28$.
We also have its box $B_i$ and class ID $C_i$.
During training $B_i$ and $C_i$ are ground truth box and class ID whereas during inference they are predicted by Mask R-CNN.
Therefore, we can obtain another representation of $i$-th instance from semantic head $X_{\text{mask}_i}$ by only taking the values inside box $B_i$ from the channel of $X_{\text{thing}}$ corresponding to $C_i$.
$X_{\text{mask}_i}$ is of size $H \times W$ and its values outside box $B_i$ are zero.
We then interpolate $Y_i$ back to the same scale as $X_{\text{mask}_i}$ via bilinear interpolation and pad zero outside the box to achieve a compatible shape with $X_{\text{mask}_i}$, denoted as $Y_{\text{mask}_i}$.
The final representation of $i$-th instance is $Z_{N_{\text{stuff}} + i} = X_{\text{mask}_i} + Y_{\text{mask}_i}$.
Once we fill in $Z$ with representations of all instances, we perform a softmax along the channel dimension to predict the pixel-wise class.
In particular, if the maximum value falls into the first $N_{\text{stuff}}$ channel, then it belongs to one of stuff classes.
Otherwise the index of the maximum value tells us the instance ID.
The architecture is shown in Fig.~\ref{fig:panoptic_head}.
During training, we generate the ground truth instance ID following the order of the ground truth boxes we used to construct the panoptic logits.
The panoptic segmentation head is then associated with the standard pixel-wise cross entropy loss.

During inference, once we predict the instance ID following the above procedure, we still need to determine the class ID of each instance.
One can either use the class ID $C_{\text{inst}}$ predicted by Mask R-CNN or the one predicted by the semantic head $C_{\text{sem}}$.
As shown later in the ablation study, we resort to a better heuristic rule.
Specifically, for any instance, we know which pixels correspond to it, \ie, those of which the \textit{argmax} of $Z$ along channel dimension equals to its instance ID. 
Among these pixels, we first check whether $C_{\text{inst}}$ and $C_{\text{sem}}$ are consistent.
If so, then we assign the class ID as $C_{\text{inst}}$.
Otherwise, we compute the mode of their $C_{\text{sem}}$, denoting as $\hat{C}_{\text{sem}}$.
If the frequency of the mode is larger than $0.5$ and $\hat{C}_{\text{sem}}$ belongs to \textit{stuff}, then the predicted class ID is $\hat{C}_{\text{sem}}$.
Otherwise, we assign the class ID as $C_{\text{inst}}$.
In short, while facing inconsistency, we trust the majority decision made by the semantic head only if it prefers a \textit{stuff} class.
The justification of such a conflict resolution heuristic is that semantic head typically achieves very good segmentation results over \textit{stuff} classes.

\vspace{-0.3cm}
\paragraph{Unknown Prediction:}


In this section, we explain a novel mechanism to allow \ourmodel~to classify a pixel as the \textit{unknown} class instead of making a wrong prediction.
To motivate our design, we consider a case where a pedestrian is instead predicted as a bicycle.
Since the prediction missed the pedestrian, the false negative (FN) value of pedestrian class will be increased by $1$. 
On the other hand, predicting it as a bicycle will be increasing the false positive (FP) of bicycle class also by $1$.
Recall that, the panoptic quality (PQ)~\cite{kirillov2018panoptic} metric for panoptic segmentation is defined as, 
\begin{align}
PQ = \underbrace{\frac{\sum_{(p, g) \in \text{TP}} \text{IoU}(p, g) }{ \vert \text{TP} \vert }}_{\text{SQ}} \underbrace{\frac{ \vert \text{TP} \vert }{ \vert \text{TP} \vert + \frac{1}{2} \vert \text{FP} \vert + \frac{1}{2} \vert \text{FN} \vert }}_{\text{RQ}}, \nonumber
\end{align}
which consist of two parts: recognition quality (RQ) and semantic quality (SQ).
It is clear that increasing either FN or FP degrades this measurement.
This phenomena extends to wrong predictions of the stuff classes as well.
Therefore, if a wrong prediction is inevitable, predicting such pixel as \textit{unknown} is preferred since it will increase FN of one class by $1$ without affecting FP of the other class.

To alleviate the issue, we compute the logits of the extra unknown class as $Z_{\text{unknown}} = \max \left( X_{\text{thing}} \right) - \max \left( X_{\text{mask}} \right)$ where $X_{\text{mask}}$ is the concatenation of $X_{\text{mask}_i}$ of all masks along channel dimension and of shape $N_{\text{inst}} \times H \times W$.
The maximum is taken along the channel dimension.
The rationale behind this is that for any pixel if the maximum of $X_{\text{thing}}$ is larger than the maximum of $X_{\text{mask}}$, then it is highly likely that we are missing some instances (FN).
The construction of the logits is shown in Fig.~\ref{fig:panoptic_head}.
To generate the ground truth for the \textit{unknown} class, we randomly sample $30\%$ ground truth masks and set them as \textit{unknown} during training.
In evaluating the metric, any pixel belonging to \textit{unknown} is ignored, \ie, setting to \textit{void} which will not contribute to the results.

\subsection{Implementation Details}\label{sect:implement_detail}

In this section, we explain the implementation details of {\ourmodel}.
We follow most of settings and hyper-parameters of Mask R-CNN which will be introduced in the supplementary material.
Hereafter, we only explain those which are different.

\vspace{-0.3cm}
\paragraph{Training: }

We implement our model in PyTorch~\cite{paszke2017automatic} and train it with $16$ GPUs using the distributed training framework Horovod~\cite{sergeev2018horovod}.
Images are preprocessed following~\cite{he2017mask}.
Each mini-batch has $1$ image per GPU.
As mentioned, we use ground truth box, mask and class label to construct the logits of panoptic head during training.
Our region proposal network (RPN) is trained end-to-end with the backbone whereas it was trained separately in \cite{he2017mask}.
Due to the high resolution of images, \eg, $1024 \times 2048$ in Cityscapes, logits from semantic head and panoptic head are downsampled to $1/4$ of the original resolution.
Although we do not fine-tune batch normalization (BN) layers within the backbone for simplicity, we still achieve comparable results with the state-of-the-art semantic segmentation networks like PSPNet. 
Based on common practice in semantic~\cite{chen2017rethinking,zhao2017pyramid} and instance segmentation~\cite{peng2017megdet,liu2018path}, we expect the performance to be further improved with BN layers fine-tuned.
Our {\ourmodel} contains $8$ loss functions in total: semantic segmentation head (whole image and RoI based pixel-wise classification losses), panoptic segmentation head (whole image based pixel-wise classification loss), RPN (box classification, box regression) and instance segmentation head (box classification, box regression and mask segmentation).
Different weighting schemes on these multi-task loss functions could lead to very different training results.
As shown in the ablation study, we found the loss balance strategy, \ie, assuring the scales of all losses are roughly on the same order of magnitude, works well in practice.

\vspace{-0.3cm}
\paragraph{Inference: }

During inference, once we obtained output boxes, masks and predicted class labels from the instance segmentation head, we apply a mask pruning process to determine which mask will be used for constructing the panoptic logits. 
In particular, we first perform the class-agnostic non-maximum suppression with the box IoU threshold as $0.5$ to filter out some overlapping boxes.
Then we sort the predicted class probabilities of the remaining boxes and keep those whose probability are larger than $0.6$.
For each class, we create a canvas which is of the same size as the image.
Then we interpolate masks of that class to the image scale and paste them onto the corresponding canvas one by one following the decreasing order of the probability.
Each time we copy a mask, if the intersection between the current mask and those already existed over the size of the current mask is larger than a threshold, we discard this mask. 
Otherwise we copy the non-intersecting part onto the canvas.
The threshold of this intersection over itself is set to $0.3$ in our experiments.
Logits from the semantic segmentation head and panoptic segmentation head are of the original scale of the input image during inference.



\section{Experiments}

In this section, we present the experimental results on COCO~\cite{lin2014microsoft}, Cityscapes~\cite{cordts2016cityscapes} and our internal dataset.

\textbf{COCO}
We follow the setup of COCO 2018 panoptic segmentation task which consists of $80$ and $53$ classes for thing and stuff respectively.
We use train2017 and val2017 subsets which contain approximately $118$k training images and $5$k validation images.

\textbf{Cityscapes}
Cityscapes has $5000$ images of ego-centric driving scenarios in urban settings which are split into $2975$, $500$ and $1525$ for training, validation and testing respectively.
It consists of $8$ and $11$ classes for \textit{thing} and \textit{stuff}.

\textbf{Our Dataset}
We also use an internal dataset which is similar to Cityscapes and consists of $10235$ training, $1139$ validation and $1186$ test images of ego-centric driving scenarios.
Our dataset consists of $10$ and $17$ classes for \textit{thing} (\eg, car, bus) and \textit{stuff} (\eg, building, road) respectively.

\begin{table}[t]
\centering
\resizebox{\linewidth}{!}{%
\setlength{\tabcolsep}{3pt}
\begin{tabular}{@{}c|ccc|cccc@{}}
\hline
\toprule
Models & PQ & SQ & RQ & $\text{PQ}^{\text{Th}}$ & $\text{PQ}^{\text{St}}$ & mIoU & AP \\
\midrule
\midrule
JSIS-Net~\cite{de2018panoptic} & 26.9 & 72.4 & 35.7 & 29.3 & 23.3 & - & - \\
RN50-MR-CNN & 38.6 & 76.4 & 47.5 & 46.2 & 27.1 & - & - \\
MR-CNN-PSP & 41.8 & \textbf{78.4} & 51.3 & 47.8 & 32.8 & 53.9 & 34.2 \\
Ours & \textbf{42.5} & 78.0 & \textbf{52.4} & \textbf{48.5} & \textbf{33.4} & \textbf{54.3} & \textbf{34.3} \\
\toprule
Multi-scale & PQ & SQ & RQ & $\text{PQ}^{\text{Th}}$ & $\text{PQ}^{\text{St}}$ & mIoU & AP \\
\midrule
\midrule
MR-CNN-PSP-M & 42.2 & 78.5 & 51.7 & 47.8 & 33.8 & 55.3 & 34.2 \\
Ours-M & \textbf{43.2} & \textbf{79.2} & \textbf{52.9} & \textbf{49.1} & \textbf{34.1} & \textbf{55.8} & \textbf{34.3} \\
\bottomrule
\end{tabular}
}
\vspace{-0.3cm}
\caption{Panoptic segmentation results on COCO. Superscripts \text{Th} and \text{St} stand for thing and stuff. `-` means inapplicable.}
\vspace{-0.5cm}
\label{table:coco}
\end{table}

\begin{table*}[t]
\centering
{%
\setlength{\tabcolsep}{3pt}
\begin{tabular}{@{}c|c|ccc|ccc|ccc@{}}
\hline
\toprule
Models & backbone & PQ & SQ & RQ & $\text{PQ}^{\text{Th}}$ & $\text{SQ}^{\text{Th}}$ & $\text{RQ}^{\text{Th}}$ & $\text{PQ}^{\text{St}}$ & $\text{SQ}^{\text{St}}$ & $\text{RQ}^{\text{St}}$ \\
\midrule
\midrule
Megvii (Face++) & ensemble model & 53.2 & 83.2 & 62.9 & 62.2 & 85.5 & 72.5 & 39.5 & 79.7 & 48.5 \\
Caribbean & ensemble model & 46.8 & 80.5 & 57.1 & 54.3 & 81.8 & 65.9 & 35.5 & 78.5 & 43.8 \\
PKU\_360 & ResNeXt-152-FPN & 46.3 & 79.6 & 56.1 & 58.6 & 83.7 & 69.6 & 27.6 & 73.6 & 35.6 \\
\midrule
JSIS-Net~\cite{de2018panoptic} & ResNet-50 & 27.2 & 71.9 & 35.9 & 29.6 & 71.6 & 39.4 & 23.4 & 72.3 & 30.6 \\
AUNet~\cite{li2018attention} & ResNeXt-152-FPN & 46.5 & \textbf{81.0} & 56.1 & \textbf{55.9} & \textbf{83.7} & \textbf{66.3} & 32.5 & 77.0 & 40.7 \\
Ours & ResNet-101-FPN & \textbf{46.6} & 80.5 & \textbf{56.9} & 53.2 & 81.5 & 64.6 & \textbf{36.7} & \textbf{78.9} & \textbf{45.3} \\
\bottomrule
\end{tabular}
}
\vspace{-0.3cm}
\caption{Panoptic segmentation results on MS-COCO 2018 \emph{test-dev}. The top $3$ rows contain results of top $3$ models taken from the official leadboard.}
\vspace{-0.5cm}
\label{table:coco_test_dev}
\end{table*}

\textbf{Experimental Setup}
For all datasets, we report results on the validation set.
To evaluate the performance, we adopt panoptic quality (PQ), recognition quality (RQ) and semantic quality (SQ)~\cite{kirillov2018panoptic} as the metrics.
We also report average precision (AP) of mask prediction, mean IoU of semantic segmentation on both \textit{stuff} and \textit{thing} and the inference run-time for comparison.
At last, we show results of ablation study on various design components of our model.
Full results with all model variants are shown in the supplementary material.

We set the learning rate and weight decay as $0.02$ and $0.0001$ for all datasets.
For COCO, we train for $90$K iterations and decay the learning rate by a factor of $10$ at $60$K and $80$K iterations.
For Cityscapes, we train for $12$K iterations and apply the same learning rate decay at $9$K iterations.
For our dataset, we train for $36$K iterations and apply the same learning rate decay at $24$K and $32$K iterations.
Loss weights of semantic head are $0.2$, $1.0$ and $1.0$ on COCO, Cityscapes and ours respectively.
RoI loss weights are one fifth of those of semantic head.
Loss weights of panoptic head are $0.1$, $0.5$ and $0.3$ on COCO, Cityscapes and ours respectively.
All other loss weights are set to $1.0$.

We mainly compare with the combined method used in~\cite{kirillov2018panoptic}.  
For a fair comparison, we adopt the model which uses a Mask R-CNN with a ResNet-50-FPN and a PSPNet with a ResNet-50 as the backbone and apply the combine heuristics to compute the panoptic segmentation.
We denote our implementation of the combined method as `MR-CNN-PSP' and its multi-scale testing version as `MR-CNN-PSP-M'. 
Unless specified otherwise, the combined method hereafter refers to `MR-CNN-PSP'.
For PSPNet, we use `poly' learning rate schedule as in~\cite{chen2018deeplab} and train $220$K, $18$K and $76$K on COCO, Cityscapes and our dataset with mini-batch size $16$.
We test all available models with the multi-scale testing.
Specifically, we average the multi-scale logits of PSPNet for the combined method and the ones of semantic segmentation head for our {\ourmodel}.
For simplicity, we just use single scale testing on Mask R-CNN of the combined method and our instance segmentation head.
During evaluation, due to the sensitivity of PQ with respect to RQ, we predict all \textit{stuff} segments of which the areas are smaller than a threshold as \textit{unknown}.
The thresholds on COCO, Cityscapes and our dataset are $4096$, $2048$ and $2048$ respectively.
To be fair, we apply this area thresholding to all methods.

\begin{table}[t]
\centering
\resizebox{\linewidth}{!}{%
\setlength{\tabcolsep}{3pt}
\begin{tabular}{@{}c|ccc|cccc@{}}
\hline
\toprule
Models & PQ & SQ & RQ & $\text{PQ}^{\text{Th}}$ & $\text{PQ}^{\text{St}}$ & mIoU & AP \\
\midrule
\midrule
Li \textit{et al.}~\cite{li2018weakly} & 53.8 & - & - &  42.5 & 62.1 & 71.6 & 28.6 \\
MR-CNN-PSP & 58.0 & 79.2 & 71.8 & 52.3 & 62.2 & 75.2 & 32.8 \\
TASCNet~\cite{li2018learning} & 55.9 & - & - & 50.5 & 59.8 & - & - \\
Ours & \textbf{59.3} & \textbf{79.7} & \textbf{73.0} & \textbf{54.6} & \textbf{62.7} & \textbf{75.2} & \textbf{33.3} \\
\midrule
TASCNet-COCO~\cite{li2018learning} & 59.2 & - & - & 56.0 & 61.5 & - & - \\
Ours-COCO & \textbf{60.5} & \textbf{80.9} & \textbf{73.5} & \textbf{57.0} & \textbf{63.0} & \textbf{77.8} & \textbf{37.8} \\
\toprule
Multi-scale & PQ & SQ & RQ & $\text{PQ}^{\text{Th}}$ & $\text{PQ}^{\text{St}}$ & mIoU & AP \\
\midrule
\midrule
Kirillov \textit{et al.}~\cite{kirillov2018panoptic} & 61.2 & 80.9 & 74.4 & 54.0 & \textbf{66.4} & \textbf{80.9} & 36.4 \\
MR-CNN-PSP-M & 59.2 & 79.7 & 73.0 & 52.3 & 64.2 & 76.9 & 32.8 \\
Ours-M & 60.1 & 80.3 & 73.5 & 55.0 & 63.7 & 76.8 & 33.3\\
Ours-101-M-COCO & \textbf{61.8} & \textbf{81.3} & \textbf{74.8} & \textbf{57.6} & 64.8 & 79.2 & \textbf{39.0} \\
\bottomrule
\end{tabular}
}
\vspace{-0.3cm}
\caption{Panoptic segmentation results on Cityscapes. '-COCO' means the model is pretrained on COCO. `-101` means the model uses ResNet-101 as the backbone. Unless specified, all models use ResNet-50 as the backbone and are pretrained on ImageNet.}
\vspace{-0.3cm}
\label{table:cityscapes}
\end{table}

\subsection{COCO}

We compare with several recent published methods including JSIS-Net~\cite{de2018panoptic}, RN50-MR-CNN~\footnote{\url{https://competitions.codalab.org/competitions/19507\#results}} and the combined method~\cite{kirillov2018panoptic} on COCO.
Since authors in~\cite{kirillov2018panoptic} do not report results on COCO, we use our MR-CNN-PSP model as the alternative to do the experiments.
JSIS-Net uses a ResNet-50 wheres RN50-MR-CNN uses two separate ResNet-50-FPNs as the backbone.
Our {\ourmodel} adopts a ResNet-50-FPN as the backbone.
In order to better leverage context information, we use a global average pooling over the feature map of the last layer in the 5-th stage of
ResNet (`res5'), reduce its dimension to $256$ and add back to FPN before producing $P_5$ feature map.
Table~\ref{table:coco} shows the results of all metrics.
The mIoU metric is computed over the $133$ classes of stuff and thing in the COCO 2018 panoptic segmentation task which is different from previous $172$ classes of COCO-Stuff. 
We are among the first to evaluate mIoU on this $133$-class subset.
From the table, we can see that our {\ourmodel} achieves better performance in all metrics except the SQ.
It is typically the case that the an increase in RQ leads to the slight decrease of SQ since we include more TP segments which could have possibly lower IoU.
Note that even with multi-scale testing, MR-CNN-PSP is still worse than ours on PQ.
Moreover, from the mIoU column, we can see that the performance of our semantic head is even better than a separate PSPNet which verifies its effectiveness.
With multi-scale testing, both MR-CNN-PSP and {\ourmodel} are improved and ours is still better.
We also add the comparisons on the \emph{test-dev} of MS-COCO 2018 in Table~\ref{table:coco_test_dev}.
Although we just use ResNet-101 as the backbone, we achieve slightly better results compared to the recent AUNet~\cite{li2018attention} which uses ResNeXt-152.
We also list the top three results on the leadboard which uses ensemble and other tricks.
It is clear from the table that we are on par with the second best model without using any such tricks.
In terms of the model size, RN50-MR-CNN, MR-CNN-PSP and {\ourmodel} consists of $71.2$M, $91.6$M and $46.1$M parameters respectively.
Therefore, our model is significantly lighter.
We show visual examples of panoptic segmentation on this dataset in the first two rows of Fig.~\ref{fig:results}.
From the $1$-st row of the figure, we can see that the combined method completely ignores the cake and other objects on the table whereas ours successfully segments them out.
This is due to the inherent limitations of the combine heuristic which first pastes the high confidence segment, \ie, table, and then ignores all highly overlapped objects thereafter.

\begin{table}[t]
\centering
\resizebox{\linewidth}{!}{%
\setlength{\tabcolsep}{3pt}
\begin{tabular}{@{}c|ccc|cccc@{}}
\hline
\toprule
Models & PQ & SQ & RQ & $\text{PQ}^{\text{Th}}$ & $\text{PQ}^{\text{St}}$ & mIoU & AP \\
\midrule
\midrule
MR-CNN-PSP & 45.5 & 77.1 & 57.6 & 40.1 & 48.7 & 70.5 & 29.6 \\
Ours & \textbf{47.1} & \textbf{77.1} & \textbf{59.4} & \textbf{43.8} & \textbf{49.0} & \textbf{70.8} & \textbf{30.4} \\
\bottomrule
\end{tabular}
}
\vspace{-0.3cm}
\caption{Panoptic segmentation results on our dataset.}
\vspace{-0.3cm}
\label{table:our_dataset}
\end{table}

\subsection{Cityscapes}

We compare our model on Cityscapes with Li \textit{et al.}~\cite{li2018weakly}, the combined method~\cite{kirillov2018panoptic} and TASCNet~\cite{li2018learning}.
Note that the method in~\cite{li2018weakly} uses a ResNet-101 as the backbone whereas all other reported methods use ResNet-50 within their backbones.
We use $2$ deformable convolution layers for the semantic head.
The results are reported in Table~\ref{table:cityscapes}.
It is clear from the table that both our {\ourmodel} and MR-CNN-PSP significantly outperform the method in~\cite{li2018weakly}, especially on $\text{PQ}^{\text{Th}}$.
This may possibly be caused by the fact that their CRF based instance subnetwork performs worse compared to Mask R-CNN on instance segmentation. 
Under the same single scale testing, our model achieves better performance than the combined method.
Although multi-scale testing significantly improves both the combined method and {\ourmodel}, ours is still slightly better.
Results reported in~\cite{kirillov2018panoptic} are different from the ones of our MR-CNN-PSP-M since: 1) they use ResNet-101 as the backbone for PSPNet; 2) they pre-train Mask R-CNN on COCO and PSPNet on extra coarsely labeled data.
We also have a model variant, denoting as `Ours-101-M', which adopts ResNet-101 as the backbone and pre-trains on COCO.
As you can see, it outperforms the reported metrics of the combined method. 
We show the visual examples of panoptic segmentation on this dataset in the $3$-rd and $4$-th rows of Fig.~\ref{fig:results}.
From the $3$-rd row of the figure, we can see that the combined method tends to produce large black area, \ie, \textit{unknown}, whenever the instance and semantic segmentation conflicts with each other.
In contrast, our {\ourmodel} resolves the conflicts better.
Moreover, it is interesting to note that some \textit{unknown} prediction of our model has the vertical or horizontal boundaries.
This is caused by the fact that instance head predicts nothing whereas semantic head predicts something for these out-of-box areas.
The logits of \textit{unknown} class will then stand out by design to avoid contributing to both FP and FN as described in section~\ref{sect:model_arch}.

\subsection{Our Dataset}

\begin{figure*}
	\centering
	\subfloat{\includegraphics[width = .249\linewidth]{imgs/vis/image/coco_1.png}} \hfill
	\subfloat{\includegraphics[width = .249\linewidth]{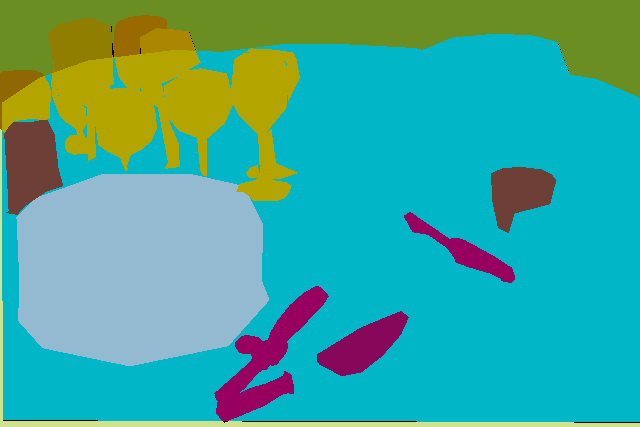}} \hfill
	\subfloat{\includegraphics[width = .249\linewidth]{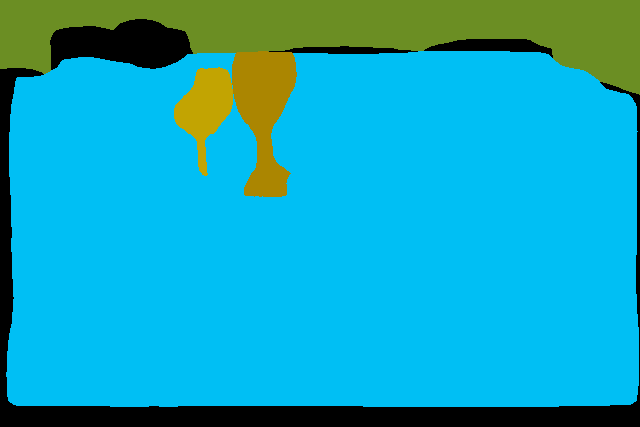}} \hfill
	\subfloat{\includegraphics[width = .249\linewidth]{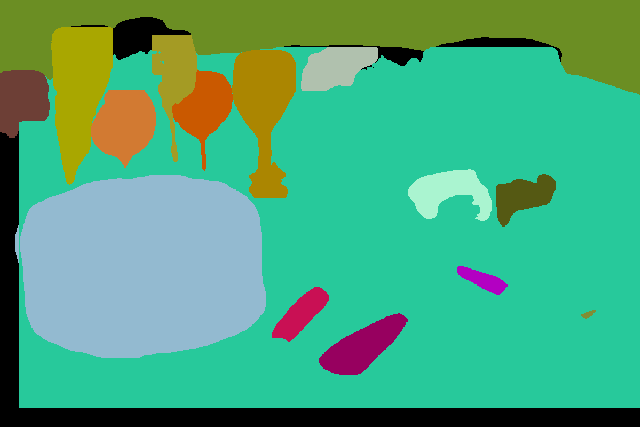}} \\ 
	\vspace{-0.35cm}
	\subfloat{\includegraphics[width = .249\linewidth]{imgs/vis/image/coco_2.png}} \hfill
	\subfloat{\includegraphics[width = .249\linewidth]{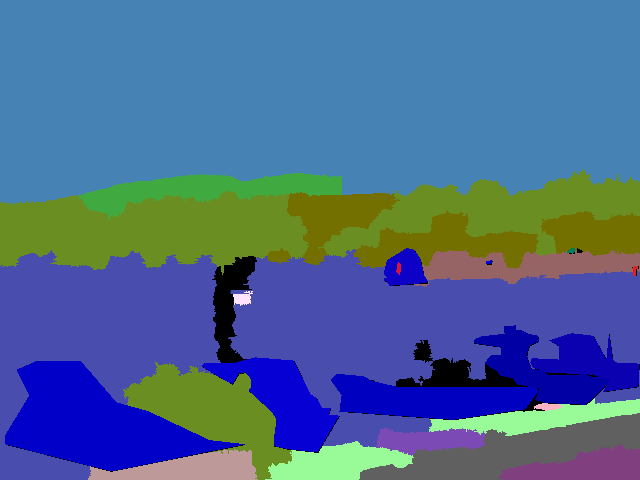}} \hfill
	\subfloat{\includegraphics[width = .249\linewidth]{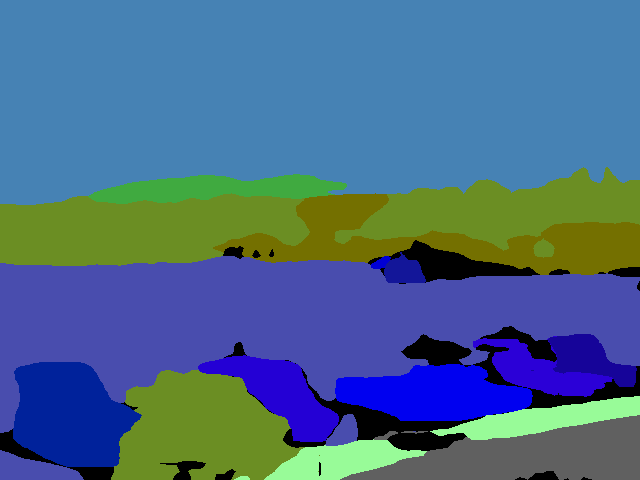}} \hfill
	\subfloat{\includegraphics[width = .249\linewidth]{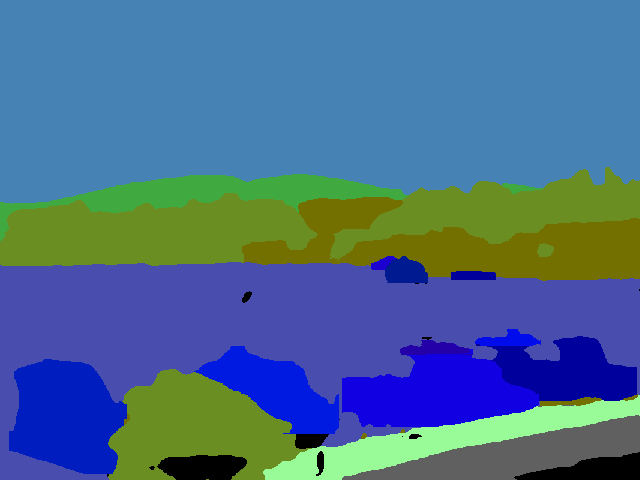}} \\ 
	\vspace{-0.35cm}
	\subfloat{\includegraphics[width = .249\linewidth]{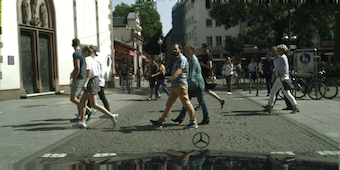}} \hfill
	\subfloat{\includegraphics[width = .249\linewidth]{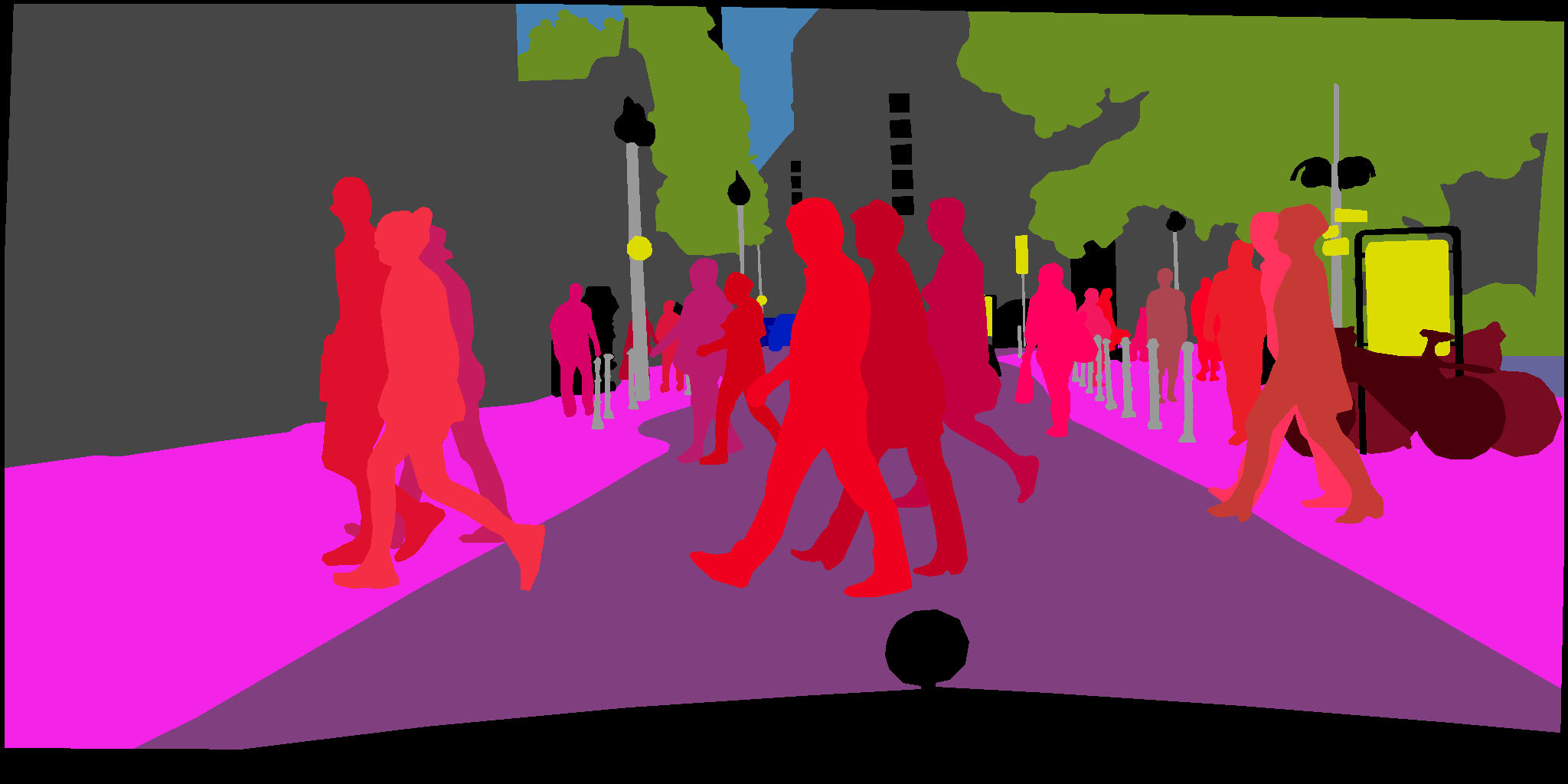}} \hfill
	\subfloat{\includegraphics[width = .249\linewidth]{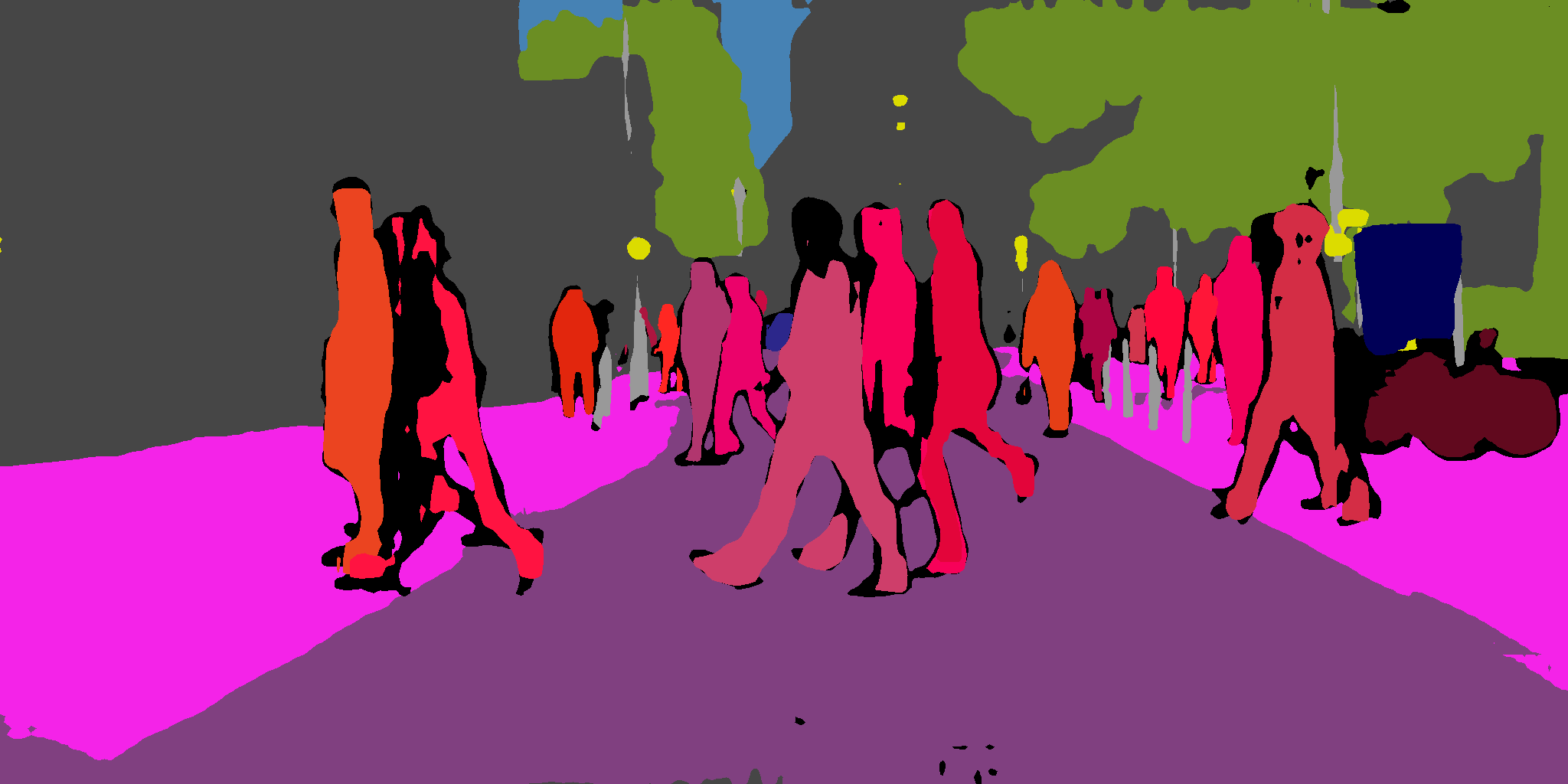}} \hfill
	\subfloat{\includegraphics[width = .249\linewidth]{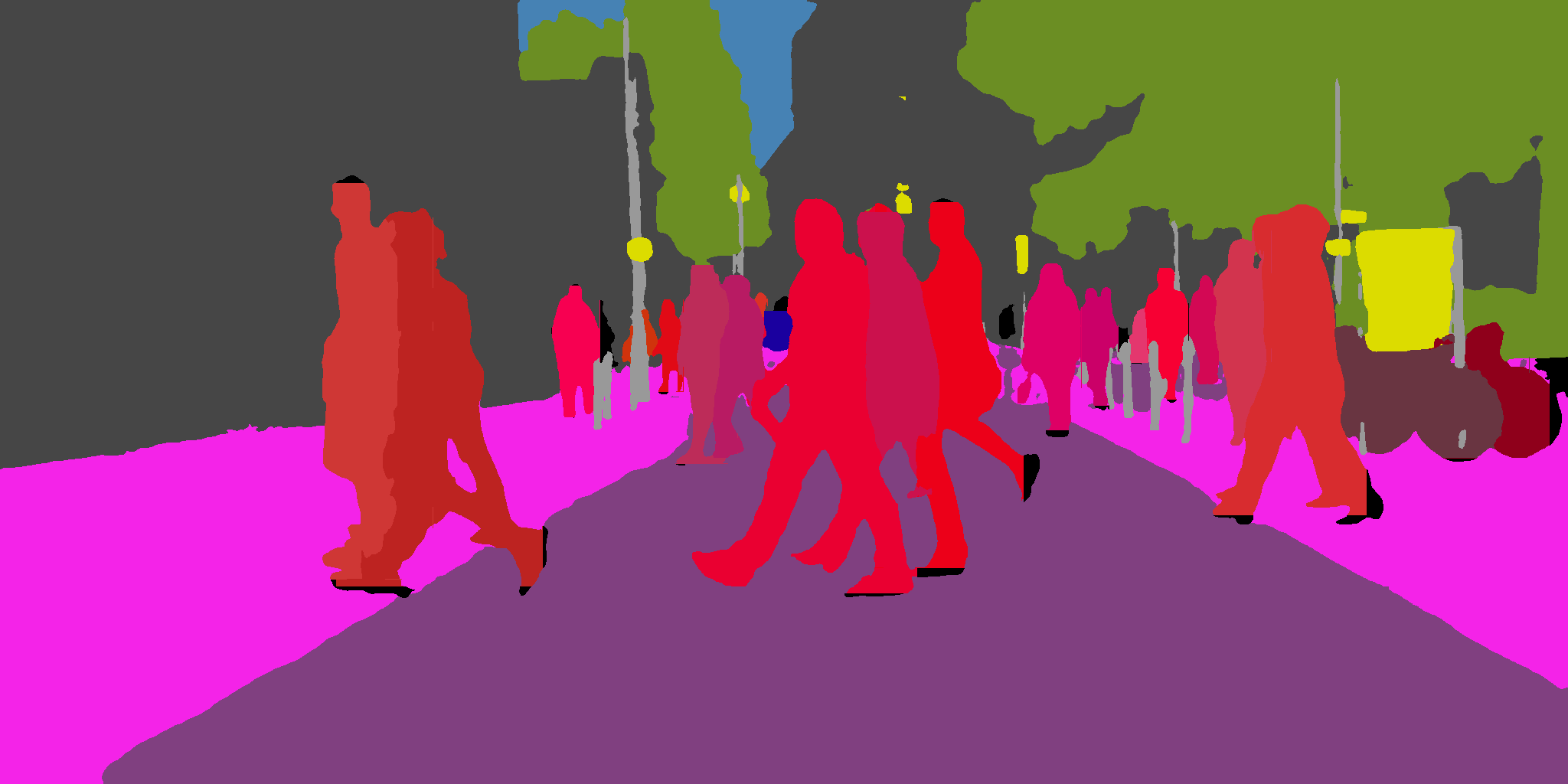}} \\ 
	\vspace{-0.35cm}
	\subfloat{\includegraphics[width = .249\linewidth]{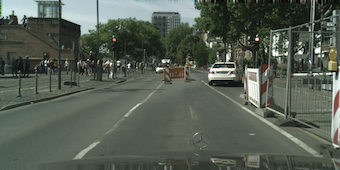}} \hfill
	\subfloat{\includegraphics[width = .249\linewidth]{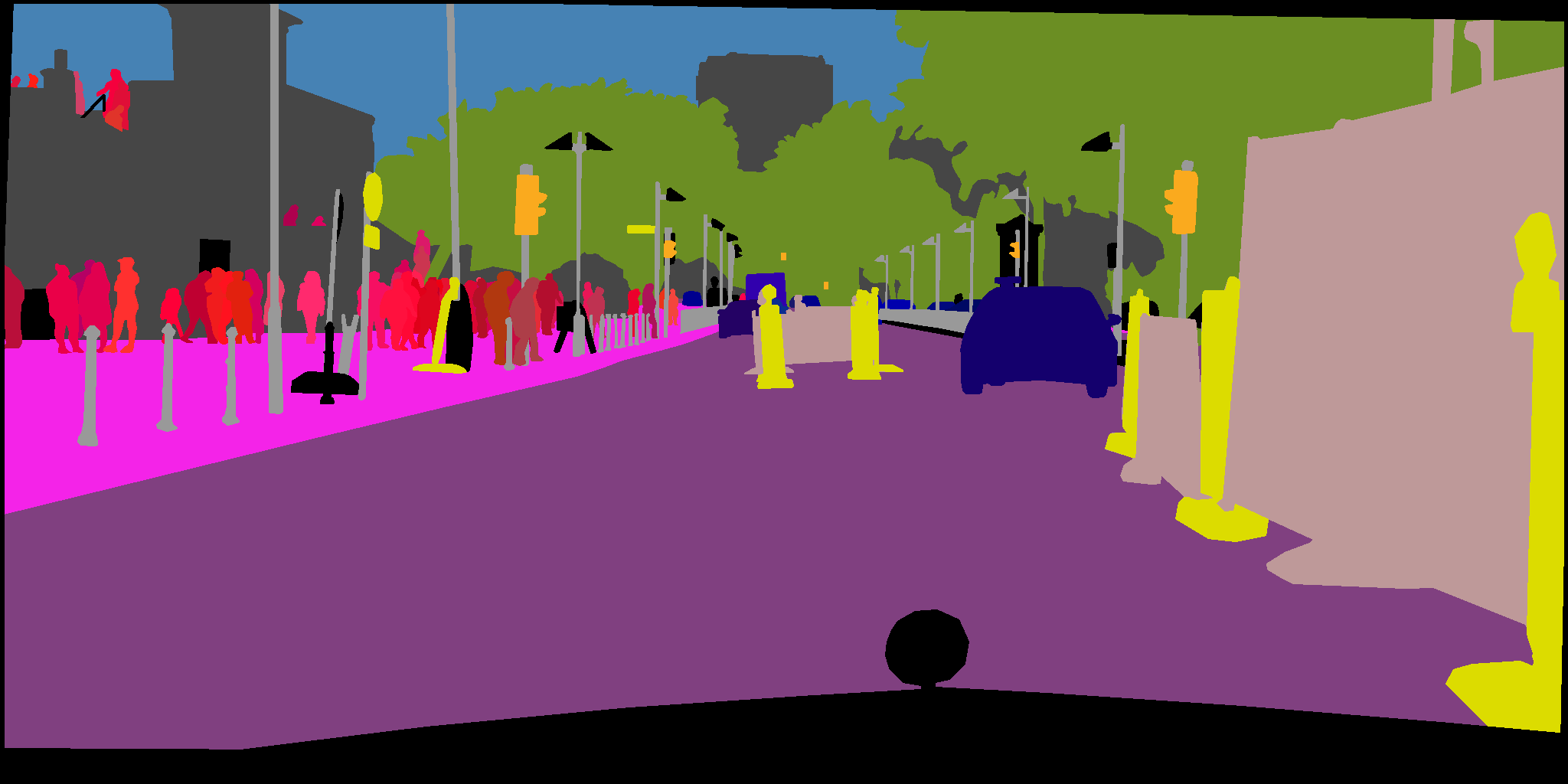}} \hfill
	\subfloat{\includegraphics[width = .249\linewidth]{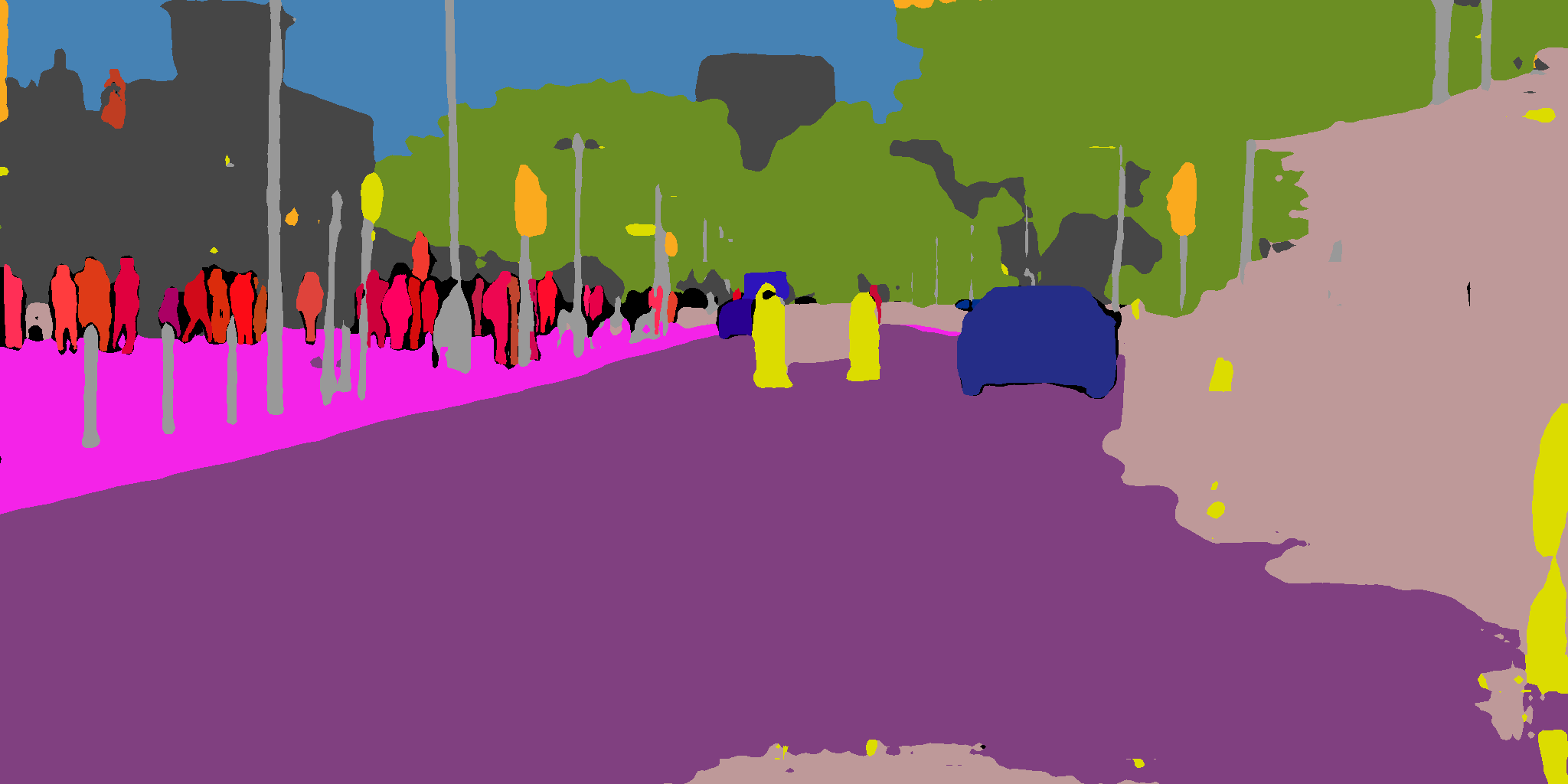}} \hfill
	\subfloat{\includegraphics[width = .249\linewidth]{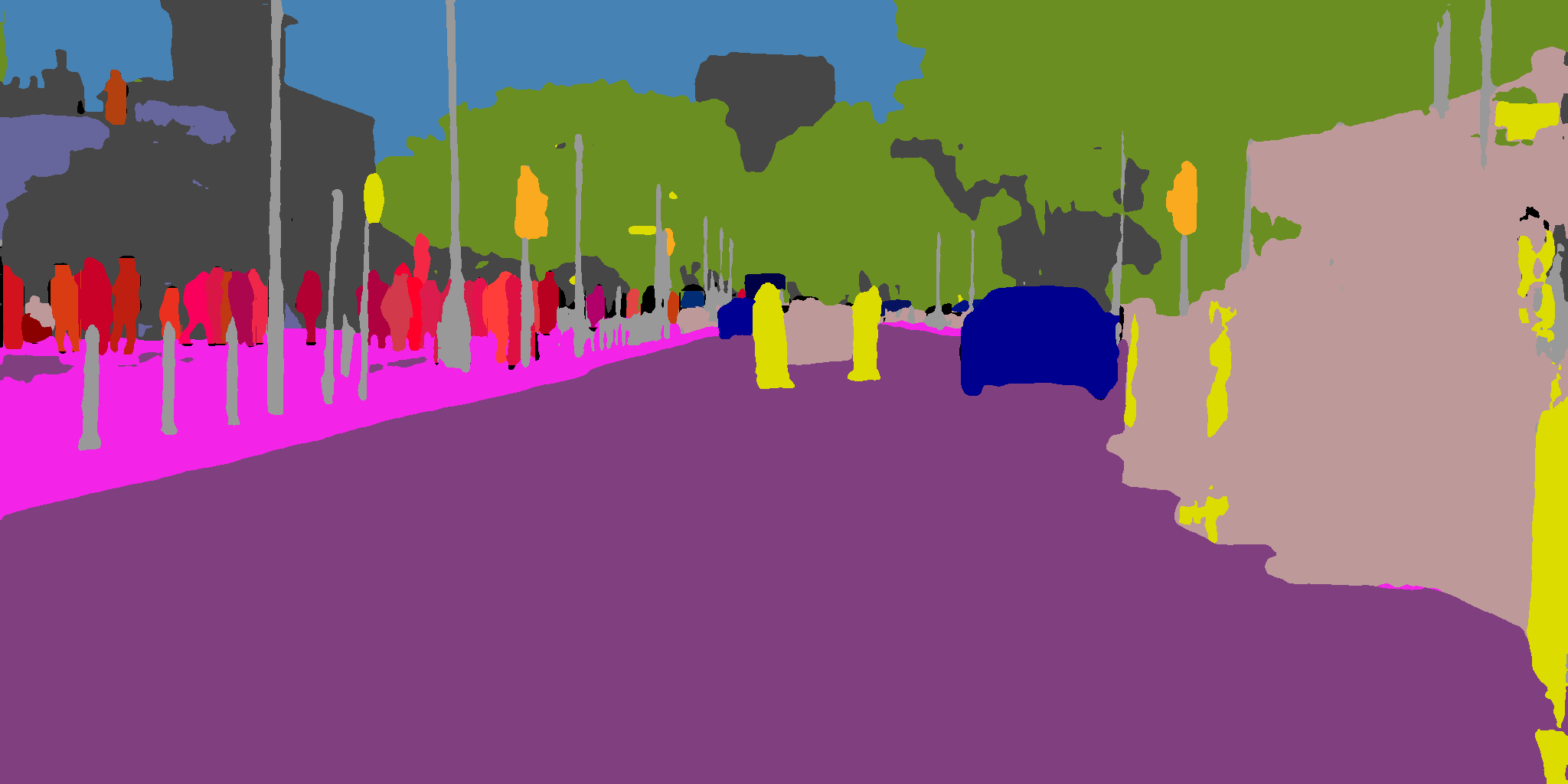}} \\ 	
	\vspace{-0.35cm}
	\subfloat{\includegraphics[width = .249\linewidth]{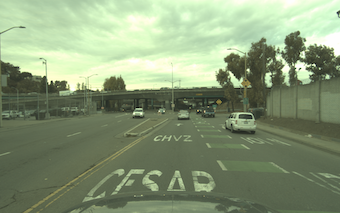}} \hfill
	\subfloat{\includegraphics[width = .249\linewidth]{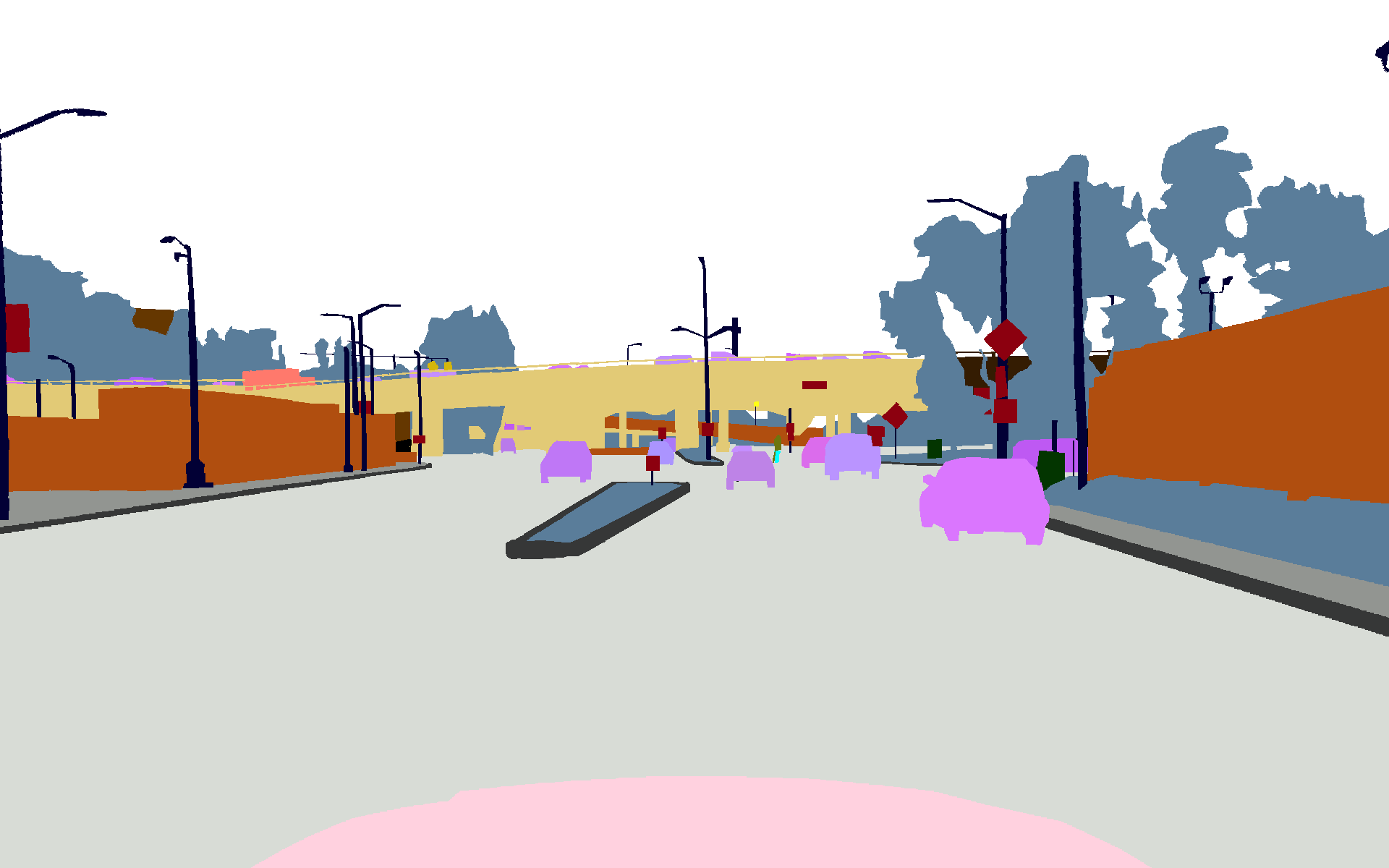}} \hfill
	\subfloat{\includegraphics[width = .249\linewidth]{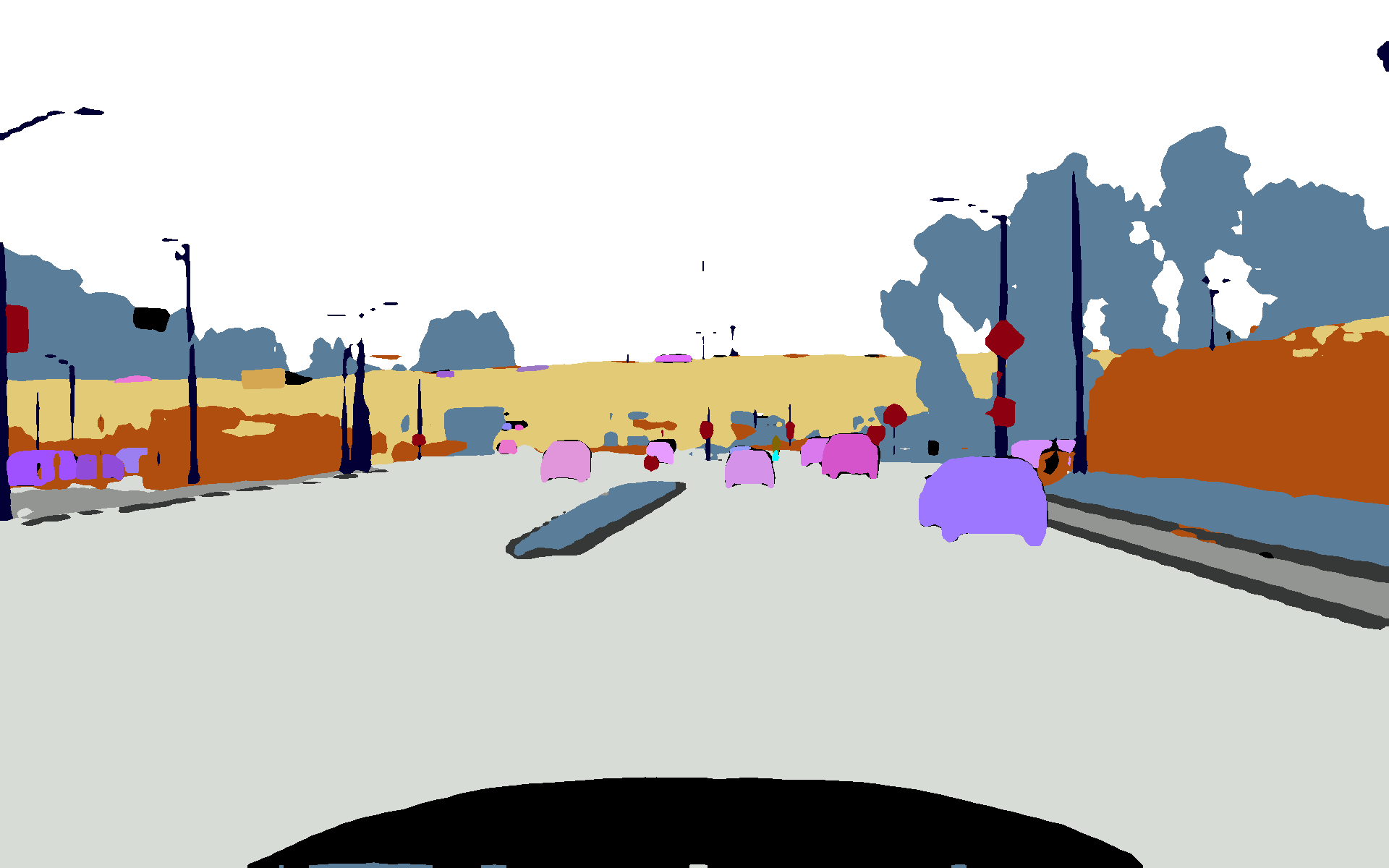}} \hfill
	\subfloat{\includegraphics[width = .249\linewidth]{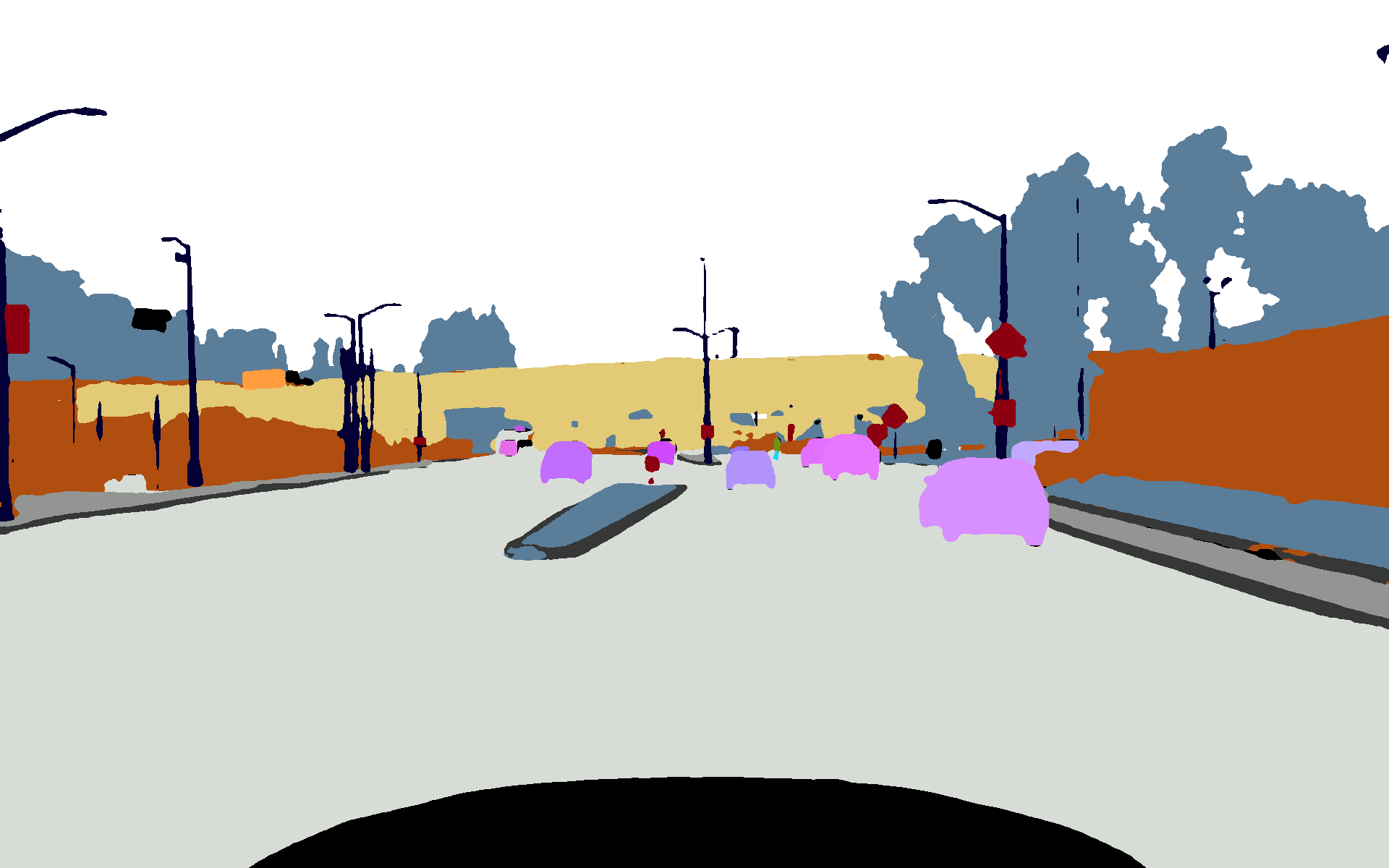}} \\
	\vspace{-0.35cm}
	\subfloat[Image]{\includegraphics[width = .249\linewidth]{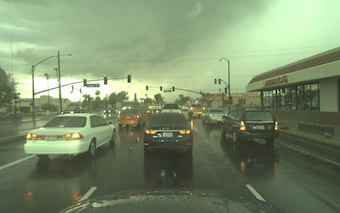}} \hfill
	\subfloat[Ground truth]{\includegraphics[width = .249\linewidth]{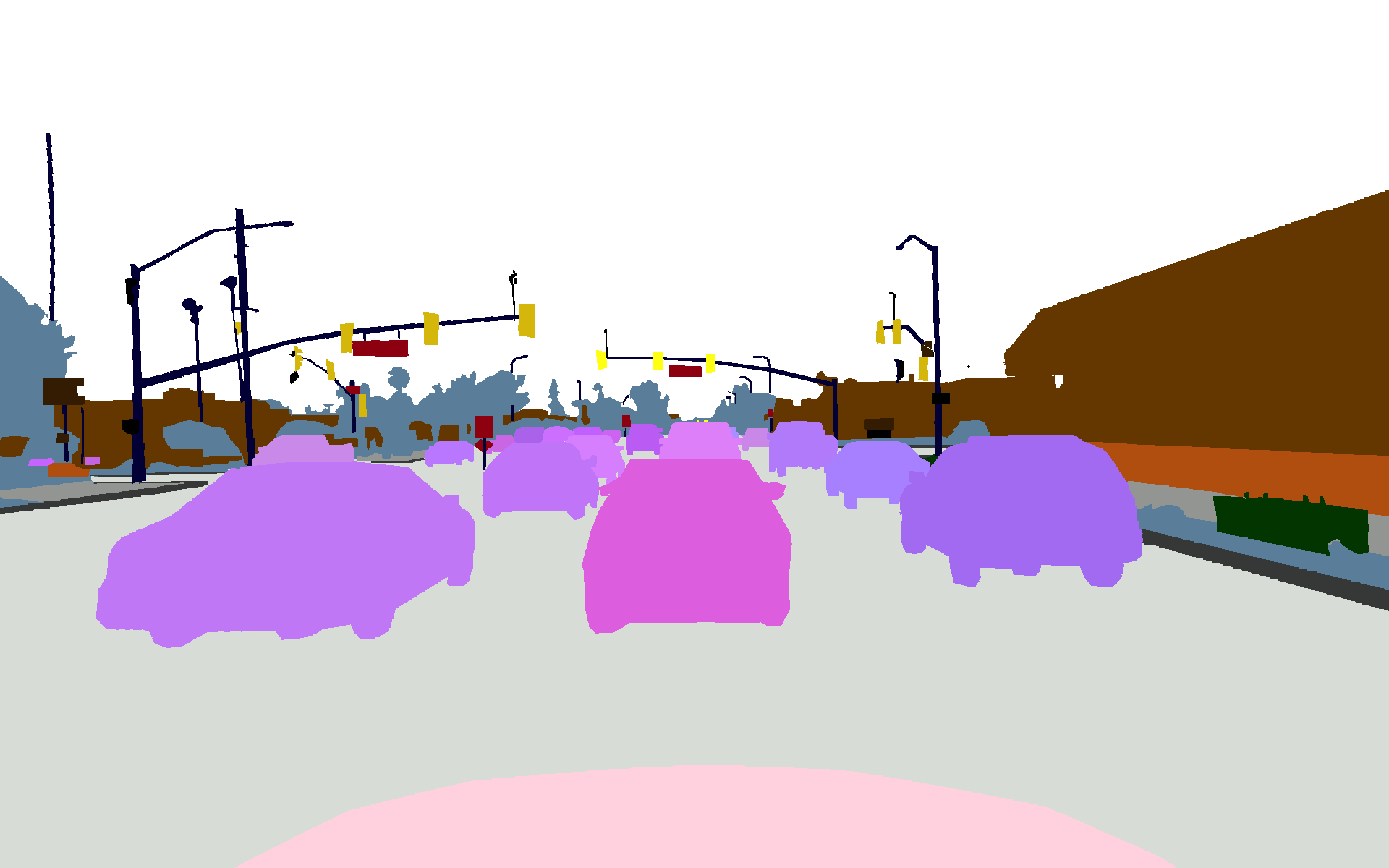}} \hfill
	\subfloat[Combined method~\cite{kirillov2018panoptic}]{\includegraphics[width = .249\linewidth]{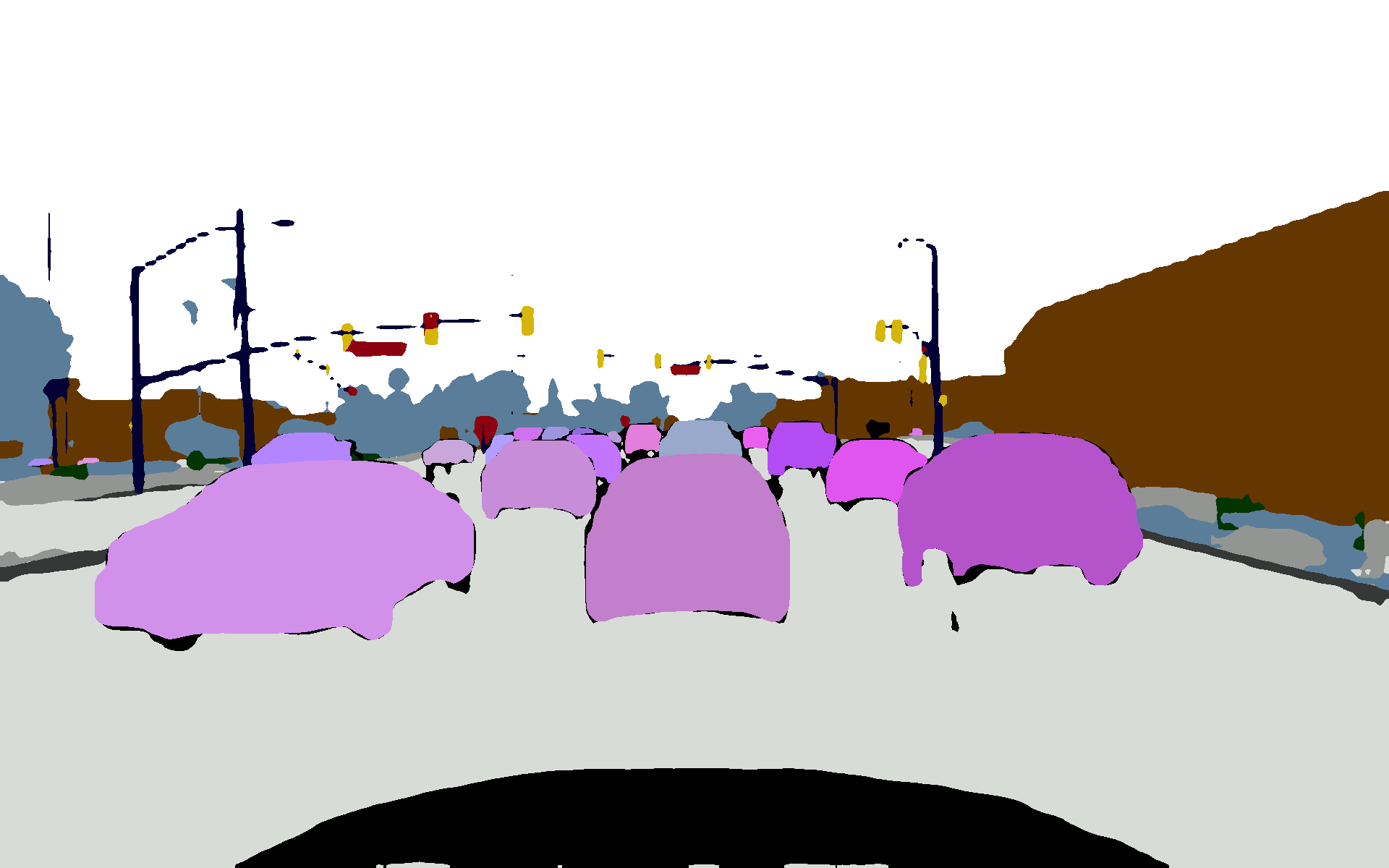}} \hfill
	\subfloat[Ours]{\includegraphics[width = .249\linewidth]{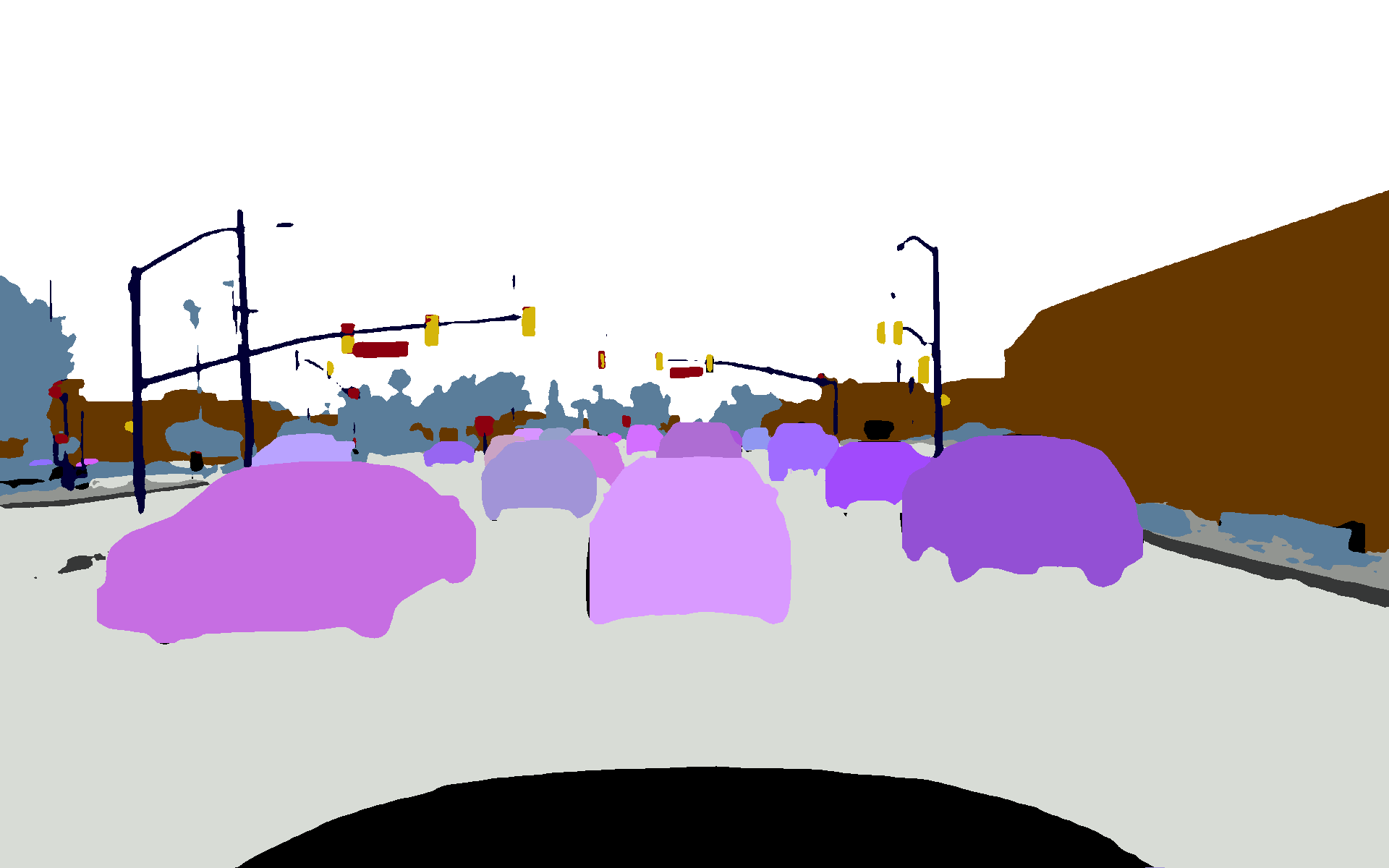}} \\
	\vspace{-0.3cm}
    \caption{Visual examples of panoptic segmentation. $1$-st and $2$-nd rows are from COCO. $3$-rd and $4$-th rows are from Cityscapes. $5$-th and $6$-th rows are from our internal dataset.}
    \vspace{-0.5cm}
   	\label{fig:results}
\end{figure*}

Last but not least, we compare our model on our own dataset with the combined method~\cite{kirillov2018panoptic}.
All reported methods use ResNet-50 within their backbones.
We use $2$ deformable convolution layers for the semantic head.
The results are reported in Table~\ref{table:our_dataset}.
We can observe that similar to COCO, our model performs significantly better than the combined method on all metrics except SQ.
We show the visual examples of panoptic segmentation on this dataset in the last two rows of Fig.~\ref{fig:results}.
From the examples, similar observations as COCO and Cityscapes are found.

\begin{table}[t]
\centering
\resizebox{\linewidth}{!}{%
\setlength{\tabcolsep}{3pt}
\begin{tabular}{@{}c|ccc@{}}
\hline
\toprule
Model & COCO & Cityscapes & Our Dataset \\
\midrule
\midrule
Img. Size & $800 \times 1300$ & $1024 \times 2048$ & $1200 \times 1920$ \\
\midrule
MR-CNN-PSP & \begin{tabular}{@{}c@{}}188 \\ (186 + 2) \\ \end{tabular} & \begin{tabular}{@{}c@{}}1105 \\ (583 + 522) \\ \end{tabular} & \begin{tabular}{@{}c@{}}1016 \\ (598 + 418) \\ \end{tabular} \\ \hline
Ours & \textbf{171} & \textbf{236} & \textbf{264} \\ \hline
Speedup & 10$\%$ $\times$ & 368$\%$ $\times$ & 285$\%$ $\times$ \\
\bottomrule
\end{tabular}
}
\vspace{-0.3cm}
\caption{Run time (ms) comparison. Note the bracket below the MR-CNN-PSP results contains their breakdown into the network inference (left) and the combine heurisitc (right).}
\vspace{-0.5cm}
\label{table:run_time}
\end{table}

\captionsetup[subfigure]{labelformat=empty}

\subsection{Run Time Comparison}

We compare the run time of inference on all three datasets in Table~\ref{table:run_time}.
We use a single NVIDIA GeForce GTX 1080 Ti GPU and an Intel Xeon E5-2687W CPU (3.00GHz).
All entries are averaged over $100$ runs on the same image with single scale test.
For COCO, the PSPNet within the combined method uses the original scale.
We also list the image size on each dataset.
It is clearly seen in the table that as the image size increases, our {\ourmodel} is significantly faster in run time. 
For example, the combined method takes about $3 \times$ time than ours.

\subsection{Ablation Study}

We perform extensive ablation studies on COCO dataset to verify our design choices as listed in Table~\ref{table:ablation}.
Empty cells in the table indicate the corresponding component is not used.
All evaluation metrics are computed over the output of the panoptic head on the validation set. 

\textbf{Panoptic Head:}
Since the inference of our panoptic head is applicable as long as we have outputs from both semantic and instance segmentation heads, we can first train these two heads simultaneously and then directly evaluate the panoptic head.
We compare the results with the ones obtained by training all three heads.
By doing so, we can verify the gain of training the panoptic head over treating it as a post processing procedure. 
From the first two rows of Table~\ref{table:ablation}, we can see that training the panoptic head does improve the PQ metric.

\textbf{Instance Class Assignment:}
Here, we focus on different alternatives of assigning instance class.
We compare our heuristic as described in section~\ref{sect:model_arch} with the one which only trusts the predicted class given by the instance segmentation head. 
As you can see from the $2$-nd and $3$-rd rows of the Table~\ref{table:ablation}, our instance class assignment is better.

\textbf{Loss Balance:}
We investigate the weighting scheme of loss functions. 
Recall that without the proposed RoI loss, our {\ourmodel} contains $7$ loss functions.
In order to weight them, we follow the principle of loss balance, \ie, making sure their values are roughly on the same order of magnitude.
In particular, with loss balance, we set the weights of semantic and panoptic losses as $0.2$ and $0.1$ and the rest ones as $1.0$.
Without loss balance, we set the weights of both semantic and panoptic losses as $0.1$ and the rest ones as $1.0$.
The $3$-rd and $4$-th rows of Table~\ref{table:ablation} show that introducing the loss balance improves the performance.

\textbf{RoI Loss \& Unknown Prediction:} 
Here, we investigate the effectiveness of our RoI loss function over the semantic head and the unknown prediction.
From $4$-th and $5$-th rows of Table~\ref{table:ablation}, one can conclude that adding such a new loss function does slightly boost the $\text{PQ}^{\text{St}}$.
From $5$-th and $6$-th rows of Table~\ref{table:ablation}, along with the RoI loss, predicting unknown class improves the metrics significantly.

\textbf{\textsc{Oracle} Results:}
We also explore the room for improvement of the current system by replacing some inference results with the ground truth (GT) ones.
Specifically, we study the box, instance class assignment and semantic segmentation results which correspond to GT Box, GT ICA and GT Seg. columns in Table~\ref{table:ablation}.
It is clear from the table that using GT boxes along with our predicted class probabilities improves PQ which indicates that better region proposals are required to achieve higher recall.
On top of the GT boxes, using the GT class assignment greatly improves the PQ, \eg, $\approx +7.0$. 
The imperfect $\text{PQ}^{\text{Th}}$ indicates that our mask segmentation is not good enough. 
Moreover, using the GT semantic segmentation gives the largest gain of PQ, \ie, $+29.5$, which highlights the importance of improving semantic segmentation.
$\text{PQ}^{\text{St}}$ is imperfect since we resize images during inference which causes the misalignment with labels. 
It is worth noticing that increasing semantic segmentation also boosts $\text{PQ}^{\text{Th}}$ for $10$ points.
This is because our model leverages semantic segments while producing instance segments.
However, it is not the case for the combined method as its predicted instance segments only relies on the instance segmentation network.

\begin{table}
\centering
\resizebox{\linewidth}{!}
{%
\setlength{\tabcolsep}{2.5pt}
\begin{tabular}{@{}cccccccc|ccc@{}}
\hline
\toprule
Pano. & \begin{tabular}{@{}c@{}}Our \\ ICA \end{tabular} & \begin{tabular}{@{}c@{}} Loss \\ Bal.\end{tabular} & \begin{tabular}{@{}c@{}}RoI \\ Loss\end{tabular} & Unk. & \begin{tabular}{@{}c@{}}GT \\ Box\end{tabular} & \begin{tabular}{@{}c@{}}GT \\ ICA\end{tabular} & \begin{tabular}{@{}c@{}}GT \\ Seg.\end{tabular} & PQ & $\text{PQ}^{\text{Th}}$ & $\text{PQ}^{\text{St}}$ \\
\midrule
\midrule
&  &  &  &  &  &  &  & 41.2 & 47.6 & 31.6 \\
\hline
\checkmark &  &  &  &  &  &  &  & 41.6 & 47.6 & 32.5 \\
\hline
\checkmark & \checkmark &  &  &  &  &  &  & 41.7 & 47.7 & 32.8 \\
\hline
\checkmark & \checkmark & \checkmark &  &  &  &  &  & 42.3 & 48.4 & 33.1 \\
\hline
\checkmark & \checkmark & \checkmark & \checkmark &  &  &  &  & 42.3 & 48.3 & 33.2 \\
\hline
\checkmark & \checkmark & \checkmark & \checkmark & \checkmark &  &  &  & 42.5 & 48.5 & 33.4 \\
\hline
\checkmark & \checkmark & \checkmark & \checkmark & \checkmark & \checkmark &  &  & 46.7 & 55.3 & 33.6 \\
\hline
\checkmark & & \checkmark & \checkmark &  & \checkmark & \checkmark &  & 53.0 & 64.6 & 35.5 \\
\hline
\checkmark & \checkmark & \checkmark & \checkmark & \checkmark &  &  & \checkmark & 72.0 & 58.6 & 92.3 \\
\bottomrule
\end{tabular}
}
\vspace{-0.3cm}
\caption{Ablation study on COCO dataset. `Pano.', `Loss Bal.', `Unk.' and `ICA' stand for training with panoptic loss, loss balance, \textit{unknown} prediction and instance class assignment respectively.}
\vspace{-0.3cm}
\label{table:ablation}
\end{table}


\section{Conclusion}

In this paper, we proposed the {\ourmodel} which provides a unified framework for panoptic segmentation.
It exploits a single backbone network and two lightweight heads to predict semantic and instance segmentation in one shot.
More importantly, our parameter-free panoptic head leverages the logits from the above two heads and has the flexibility to predict an extra \textit{unknown} class.
It handles the varying number of classes per image and enables back propagation for the bottom representation learning.
Empirical results on three large datasets show that our {\ourmodel} achieves state-of-the-art performance with significantly faster inference compared to other methods.
In the future, we would like to explore more powerful backbone networks and smarter parameterization of panoptic head.


{\small
\bibliographystyle{ieee}
\bibliography{panoptic}
}


\section{Supplementary Material}

We first explain the hyper-parameters and then provide full experimental results on all three datasets.

\subsection{Hyperparameters}

For all our experiments, we exploit a $1500$-iteration warm-up phase where the learning rate gradually increases from $0.002$ to $0.02$.   
We also initialize all models with ImageNet pre-trained weights released by MSRA~\footnote{https://github.com/KaimingHe/deep-residual-networks}.

\vspace{-0.25cm}
\paragraph{COCO:}
We resize the image such that the length of the shorter edge is $800$ and the length of the longer edge does not exceed $1333$. 
We do not utilize multi-scale training.
For testing, we feed multi-scale images for all models.
Specifically, we resize the images to multiple scales of which the shorter edge equals to $\{480, 544, 608, 672, 736, 800, 864, 928, 992, 1056, 1120\}$ respectively. 
We also add left-right flipping.
Finally, we average the semantic segmentation logits under different scales.
For PSPNet, we do sliding window test with $513 \times 513$ cropped image.

\vspace{-0.25cm}
\paragraph{Cityscapes:}
We do multi-scale training where we resize the input image in a way that the length of the shorter edge is randomly sampled from $[800, 1024]$.
For multi-scale testing, we use the same protocol as COCO except the set of scales is $\{704, 768, 832, 896, 960, 1024, 1088, 1152, 1216, 1280, 1344\}$.
For PSPNet, we do sliding window test with $713 \times 713$ cropped image.

\vspace{-0.25cm}
\paragraph{Our Dataset:}
We utilize multi-scale training where the length of the shorter edge is randomly sampled from $[800, 1200]$.
We do not perform multi-scale testing.

\subsection{Full Experimental Results}

We show the full experimental results including run time in Table~\ref{table:coco}, Table~\ref{table:cityscapes} and Table~\ref{table:our_dataset}.
We also add two more variants of our model as baselines, \ie, UPSNet-C and UPSNet-CP.
UPSNet-C is UPSNet without the panoptic head. 
We train it with just semantic and instance segmentation losses.
During test, we use the same combine heuristics as in~\cite{kirillov2018panoptic} to generate the final prediction.
UPSNet-CP has the same model as UPSNet.
We train it in the same way as UPSNet as well.
During test, we use the same combine heuristics as in~\cite{kirillov2018panoptic} to generate the final prediction.

\begin{table*}[t]
\centering
\setlength{\tabcolsep}{3pt}
\begin{tabular}{@{}c|ccc|ccc|ccc|cccc@{}}
\hline
\toprule
Models & PQ & SQ & RQ & $\text{PQ}^{\text{Th}}$ & $\text{SQ}^{\text{Th}}$ & $\text{RQ}^{\text{Th}}$ & $\text{PQ}^{\text{St}}$ & $\text{SQ}^{\text{St}}$ & $\text{RQ}^{\text{St}}$ & mIoU & $\text{AP}^{\text{box}}$ & $\text{AP}^{\text{mask}}$ & Run Time (ms) \\
\midrule
\midrule
JSIS-Net~\cite{de2018panoptic} & 26.9 & 72.4 & 35.7 & 29.3 & 72.1 & 39.2 & 23.3 & 73.0 & 30.4 & - & - & - & - \\
RN50-MR-CNN & 38.6 & 76.4 & 47.5 & 46.2 & 80.2 & 56.2 & 27.1 & 70.8 & 34.5 & - & - & - & - \\
MR-CNN-PSP & 41.8 & 78.4 & 51.3 & 47.8 & 81.3 & 58.0 & 32.8 & 74.1 & 41.1 & 53.9 & 38.1 & 34.2 & 186 \\
{\ourmodel}-C & 41.5 & 79.1 & 50.9 & 47.5 & 81.2 & 57.7 & 32.6 & \textbf{76.1} & 40.6 & \textbf{54.5} & \textbf{38.2} & \textbf{34.4} & 153 \\
{\ourmodel}-CP & 41.5 & \textbf{79.2} & 50.8 & 47.3 & \textbf{81.3} & 57.4 & 32.8 & 76.0 & 40.9 & 54.3 & 37.8 & 34.3 & $\ast$ \\
{\ourmodel} & \textbf{42.5} & 78.0 & \textbf{52.5} & \textbf{48.6} & 79.4 & \textbf{59.6} & \textbf{33.4} & 75.9 & \textbf{41.7} & 54.3 & 37.8 & 34.3 & 167 \\
\toprule
Multi-scale & PQ & SQ & RQ & $\text{PQ}^{\text{Th}}$ & $\text{SQ}^{\text{Th}}$ & $\text{RQ}^{\text{Th}}$ & $\text{PQ}^{\text{St}}$ & $\text{SQ}^{\text{St}}$ & $\text{RQ}^{\text{St}}$ & mIoU & $\text{AP}^{\text{box}}$ & $\text{AP}^{\text{mask}}$ & Run Time (ms) \\
\midrule
\midrule
MR-CNN-PSP-M & 42.2 & 78.5 & 51.7 & 47.8 & \textbf{81.3} & 58.0 & 33.8 & 74.3 & 42.2 & 55.3 & \textbf{38.1} & 34.2 & 3,624 \\
UPSNet-M & \textbf{43.0} & \textbf{79.1} & \textbf{52.8} & \textbf{48.9} & 79.7 & \textbf{59.7} & \textbf{34.1} & \textbf{78.2} & \textbf{42.3} & \textbf{55.7} & 37.8 & \textbf{34.3} & 2,433 \\
\bottomrule
\end{tabular}
\caption{Panoptic segmentation results on COCO. Superscripts \text{Th} and \text{St} stand for thing and stuff. `-` means inapplicable. `$\ast$` means the run time of {\ourmodel}-CP is the same with the one of {\ourmodel}-C.}
\label{table:coco}
\end{table*}

\begin{table*}[t]
\centering
\setlength{\tabcolsep}{3pt}
\begin{tabular}{@{}c|ccc|ccc|ccc|cccc@{}}
\hline
\toprule
Models & PQ & SQ & RQ & $\text{PQ}^{\text{Th}}$ & $\text{SQ}^{\text{Th}}$ & $\text{RQ}^{\text{Th}}$ & $\text{PQ}^{\text{St}}$ & $\text{SQ}^{\text{St}}$ & $\text{RQ}^{\text{St}}$ & mIoU & $\text{AP}^{\text{box}}$ & $\text{AP}^{\text{mask}}$ & Run Time (ms) \\
\midrule
\midrule
Li \textit{et al.}~\cite{li2018weakly} & 53.8 & - & - &  42.5 & - & - &  62.1 & - & - & 71.6 & - & 28.6 & - \\
MR-CNN-PSP & 58.0 & 79.2 & 71.8 & 52.3 & 77.9 & 66.9 & 62.2 & 80.1 & 75.4 & 75.2 & 37.7 & 32.8 & 583 \\
{\ourmodel}-C & 58.1 & 79.0 & 72.0 & 52.2 & 78.0 & 66.6 & 62.3 & 79.8 & 75.9 & \textbf{75.3} & 37.9 & 33.2 & 198 \\
{\ourmodel}-CP & 58.7 & 79.1 & 72.8 & 53.1 & 77.7 & 68.1 & 62.7 & 80.0 & \textbf{76.3} & 75.2 & 39.1 & 33.3 & $\ast$ \\
{\ourmodel} & \textbf{59.3} & \textbf{79.7} & \textbf{73.0} & \textbf{54.6} & \textbf{79.3} & \textbf{68.7} & \textbf{62.7} & \textbf{80.1} & 76.2 & 75.2 & \textbf{39.1} & \textbf{33.3} & 202 \\
\toprule
Multi-scale & PQ & SQ & RQ & $\text{PQ}^{\text{Th}}$ & $\text{SQ}^{\text{Th}}$ & $\text{RQ}^{\text{Th}}$ & $\text{PQ}^{\text{St}}$ & $\text{SQ}^{\text{St}}$ & $\text{RQ}^{\text{St}}$ & mIoU & $\text{AP}^{\text{box}}$ & $\text{AP}^{\text{mask}}$ & Run Time (ms) \\
\midrule
\midrule
Kirillov \textit{et al.}~\cite{kirillov2018panoptic} & 61.2 & 80.9 & 74.4 & 54.0 & - & - & \textbf{66.4} & - & - & \textbf{80.9} & - & 36.4 & \\
MR-CNN-PSP-M & 59.2 & 79.7 & 73.0 & 52.3 & 77.9 & 66.9 & 64.2 & 81.0 & 77.4 & 76.9 & 37.7 & 32.8 & 18,063 \\
UPSNet-50-M & 60.1 & 80.3 & 73.6 & 54.9 & 79.5 & 68.9 & 63.7 & 80.8 & 76.9 & 76.8 & 39.1 & 33.3 & 2,607 \\
UPSNet-101-M & \textbf{61.8} & \textbf{81.3} & \textbf{74.8} & \textbf{57.6} & \textbf{81.3} & \textbf{70.5} & 64.8 & \textbf{81.4} & \textbf{77.8} & 79.2 & \textbf{43.9} & \textbf{39.0} & 4,126 \\
\bottomrule
\end{tabular}
\caption{Panoptic segmentation results on Cityscapes.}
\label{table:cityscapes}
\end{table*}

\begin{table*}[t]
\centering
\setlength{\tabcolsep}{3pt}
\begin{tabular}{@{}c|ccc|ccc|ccc|cccc@{}}
\hline
\toprule
Models & PQ & SQ & RQ & $\text{PQ}^{\text{Th}}$ & $\text{SQ}^{\text{Th}}$ & $\text{RQ}^{\text{Th}}$ & $\text{PQ}^{\text{St}}$ & $\text{SQ}^{\text{St}}$ & $\text{RQ}^{\text{St}}$ & mIoU & $\text{AP}^{\text{box}}$ & $\text{AP}^{\text{mask}}$ & Run Time \\
\midrule
\midrule
MR-CNN-PSP & 45.5 & 77.1 & 57.6 & 40.1 & 77.2 & 52.1 & 48.7 & \textbf{77.1} & 60.9 & 70.5 & 32.2 & 29.6 & 598 \\
{\ourmodel}-C & 46.4 & 76.7 & 58.8 & 43.1 & 77.2 & 55.4 & 48.3 & 76.4 & 60.8 & 70.6 & 33.0 & \textbf{30.5} & 224\\
{\ourmodel}-CP & 46.7 & 76.8 & 59.1 & 42.9 & 77.1 & 55.1 & 48.9 & 76.7 & 61.5 & 70.8 & 33.2 & 30.4 & $\ast$\\
{\ourmodel} & \textbf{47.1} & \textbf{77.1} & \textbf{59.4} & \textbf{43.8} & \textbf{77.7} & \textbf{55.5} & \textbf{49.0} & 76.7 & \textbf{61.7} & \textbf{70.8} & \textbf{33.2} & 30.4 & 225 \\
\bottomrule
\end{tabular}
\caption{Panoptic segmentation results on our dataset.}
\label{table:our_dataset}
\end{table*}

\subsection{Visual Examples}

\begin{figure*}
	\centering
	\subfloat{\includegraphics[width = .249\linewidth]{imgs/vis/image/coco_3.png}} \hfill
	\subfloat{\includegraphics[width = .249\linewidth]{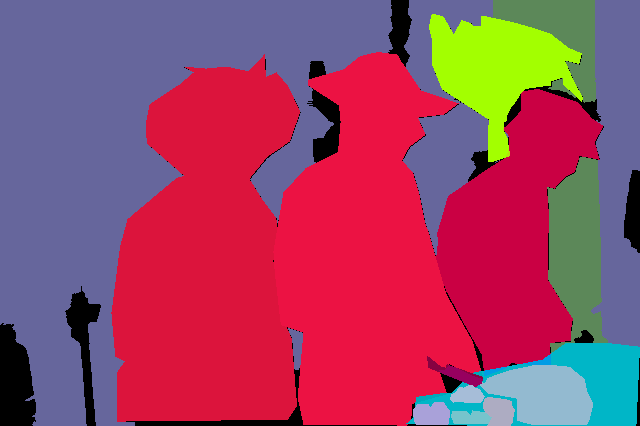}} \hfill
	\subfloat{\includegraphics[width = .249\linewidth]{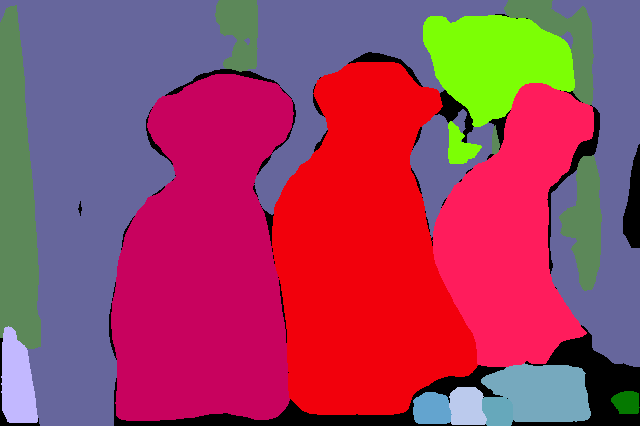}} \hfill
	\subfloat{\includegraphics[width = .249\linewidth]{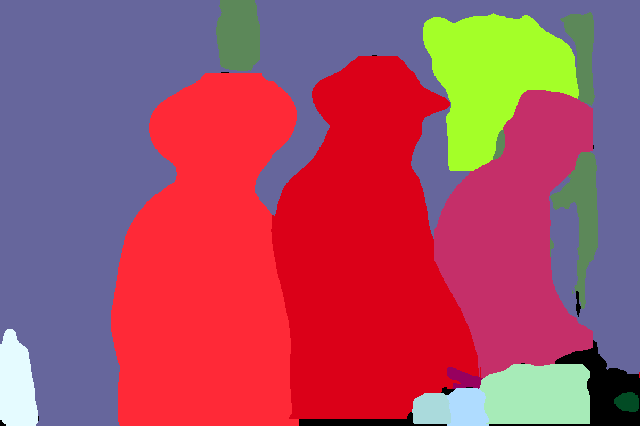}} \\ 
	\vspace{-0.35cm}
	\subfloat{\includegraphics[width = .249\linewidth]{imgs/vis/image/coco_4.png}} \hfill
	\subfloat{\includegraphics[width = .249\linewidth]{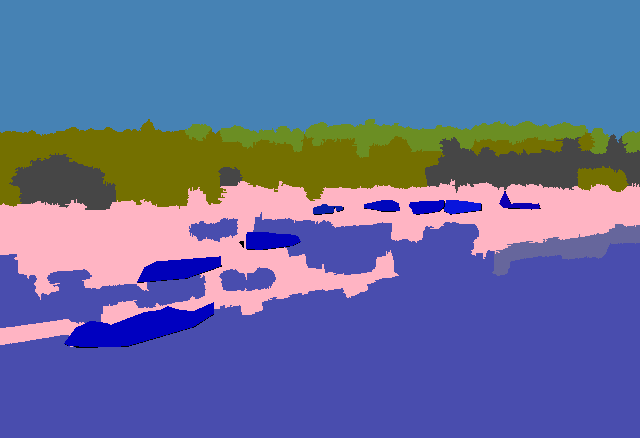}} \hfill
	\subfloat{\includegraphics[width = .249\linewidth]{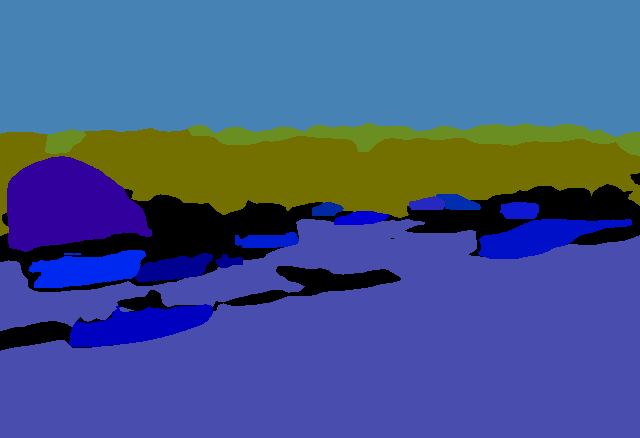}} \hfill
	\subfloat{\includegraphics[width = .249\linewidth]{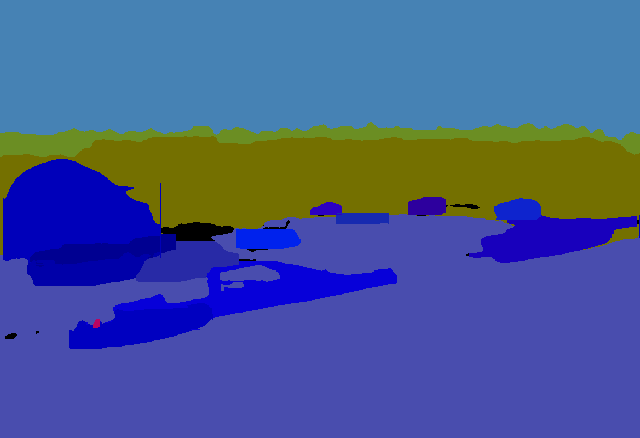}} \\ 
	\vspace{-0.35cm}
	\subfloat[Image]{\includegraphics[width = .249\linewidth]{imgs/vis/image/coco_5.png}} \hfill
	\subfloat[Ground truth]{\includegraphics[width = .249\linewidth]{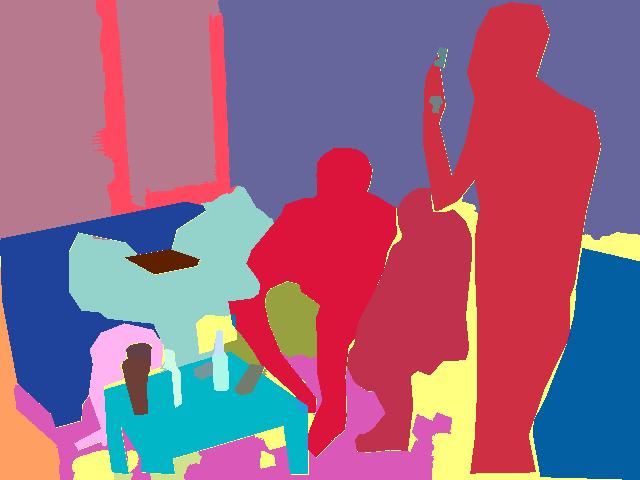}} \hfill
	\subfloat[Combined method~\cite{kirillov2018panoptic}]{\includegraphics[width = .249\linewidth]{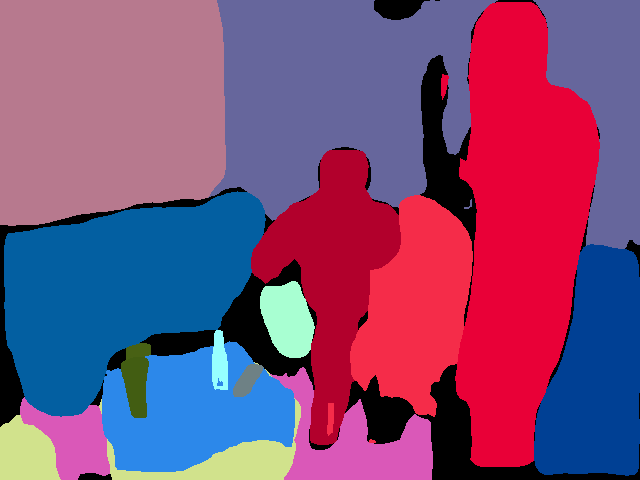}} \hfill
	\subfloat[Ours]{\includegraphics[width = .249\linewidth]{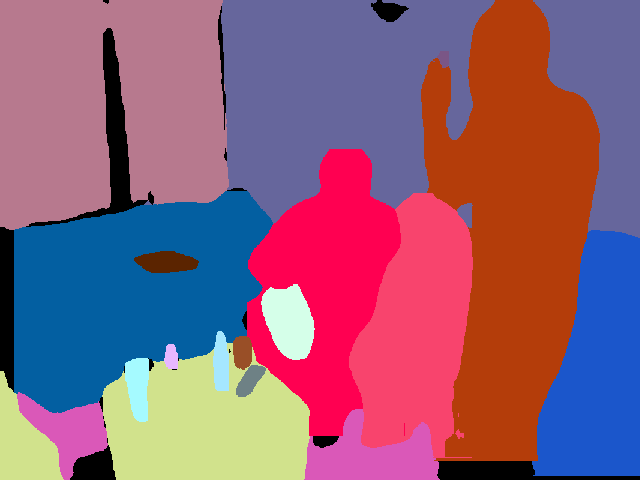}} \\
    \caption{Visual examples of panoptic segmentation on COCO.}
   	\label{fig:results_coco}
\end{figure*}

\begin{figure*}
	\centering	
	\subfloat{\includegraphics[width = .249\linewidth]{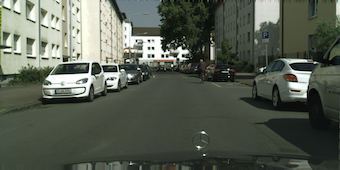}} \hfill
	\subfloat{\includegraphics[width = .249\linewidth]{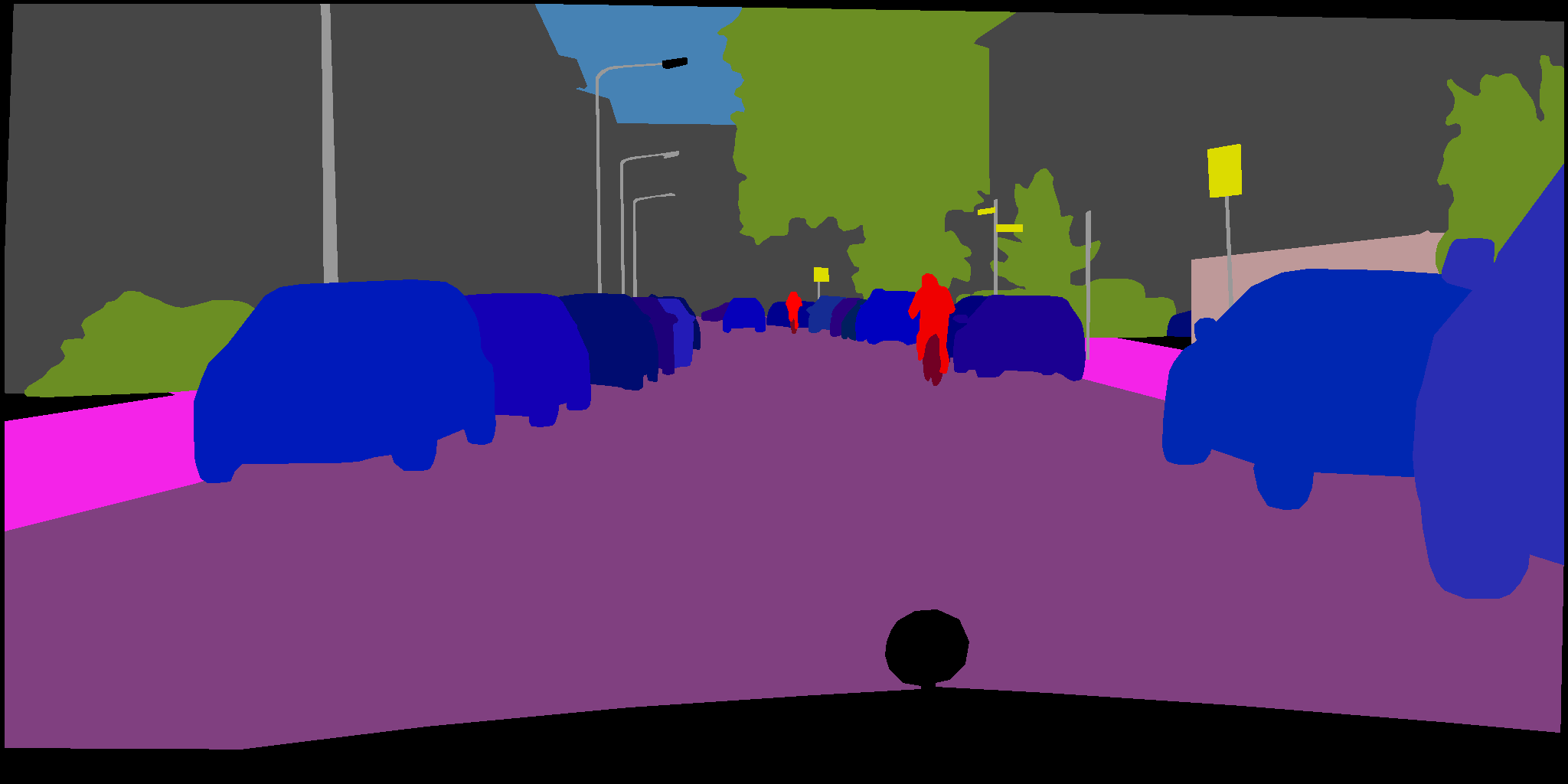}} \hfill
	\subfloat{\includegraphics[width = .249\linewidth]{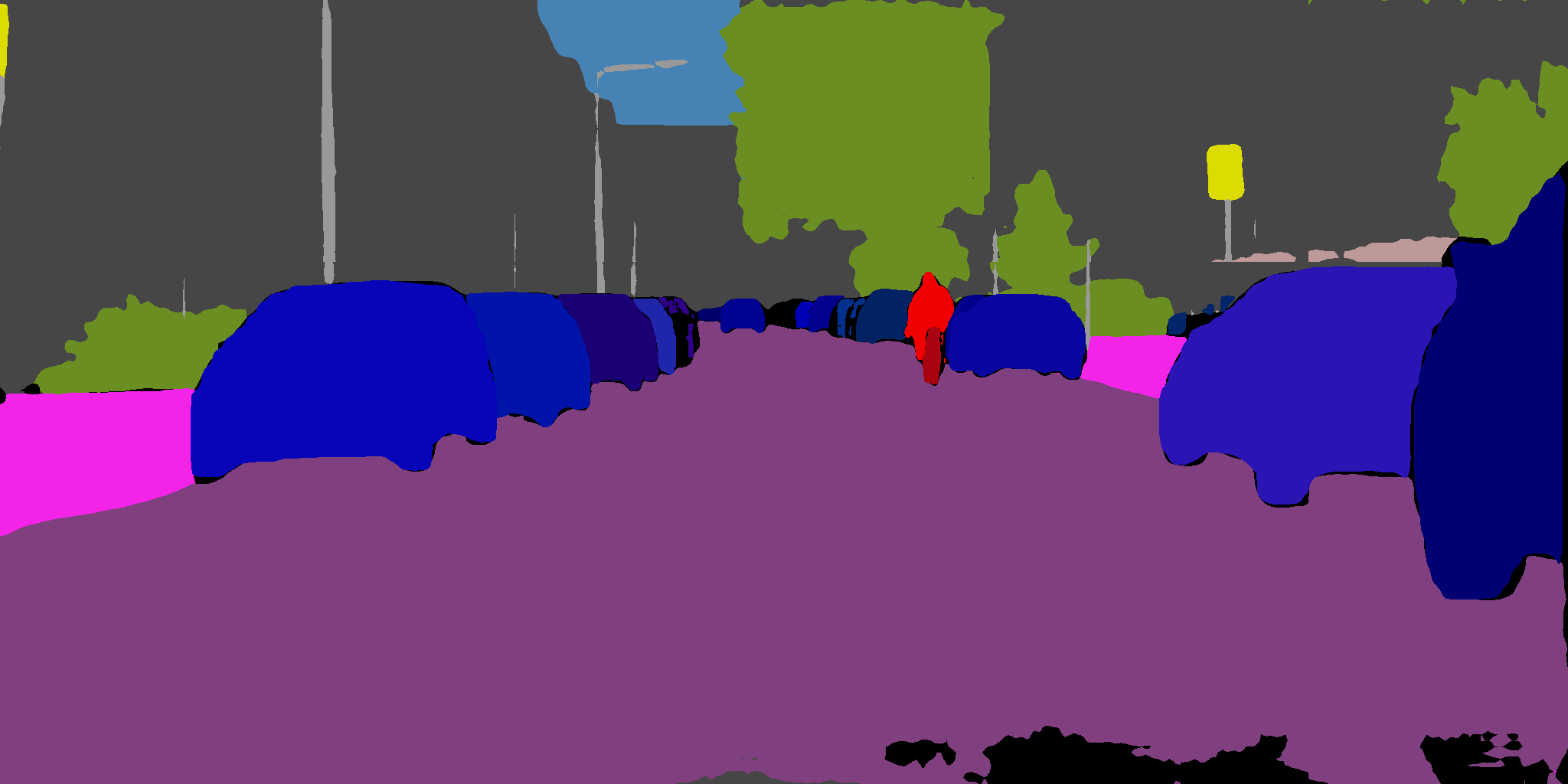}} \hfill
	\subfloat{\includegraphics[width = .249\linewidth]{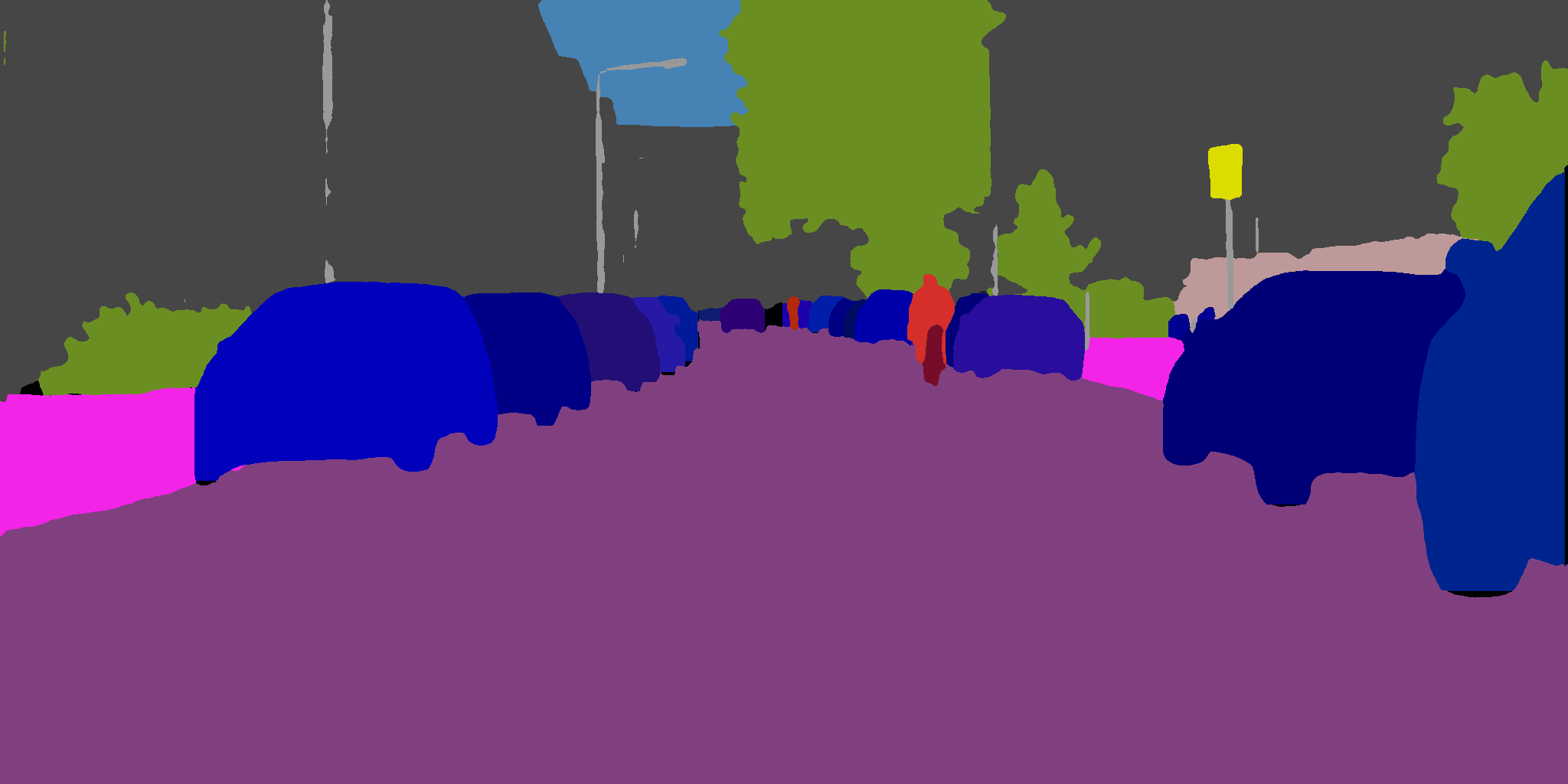}} \\ 
	\vspace{-0.35cm}	
	\subfloat{\includegraphics[width = .249\linewidth]{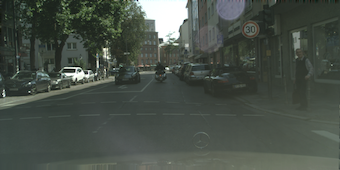}} \hfill
	\subfloat{\includegraphics[width = .249\linewidth]{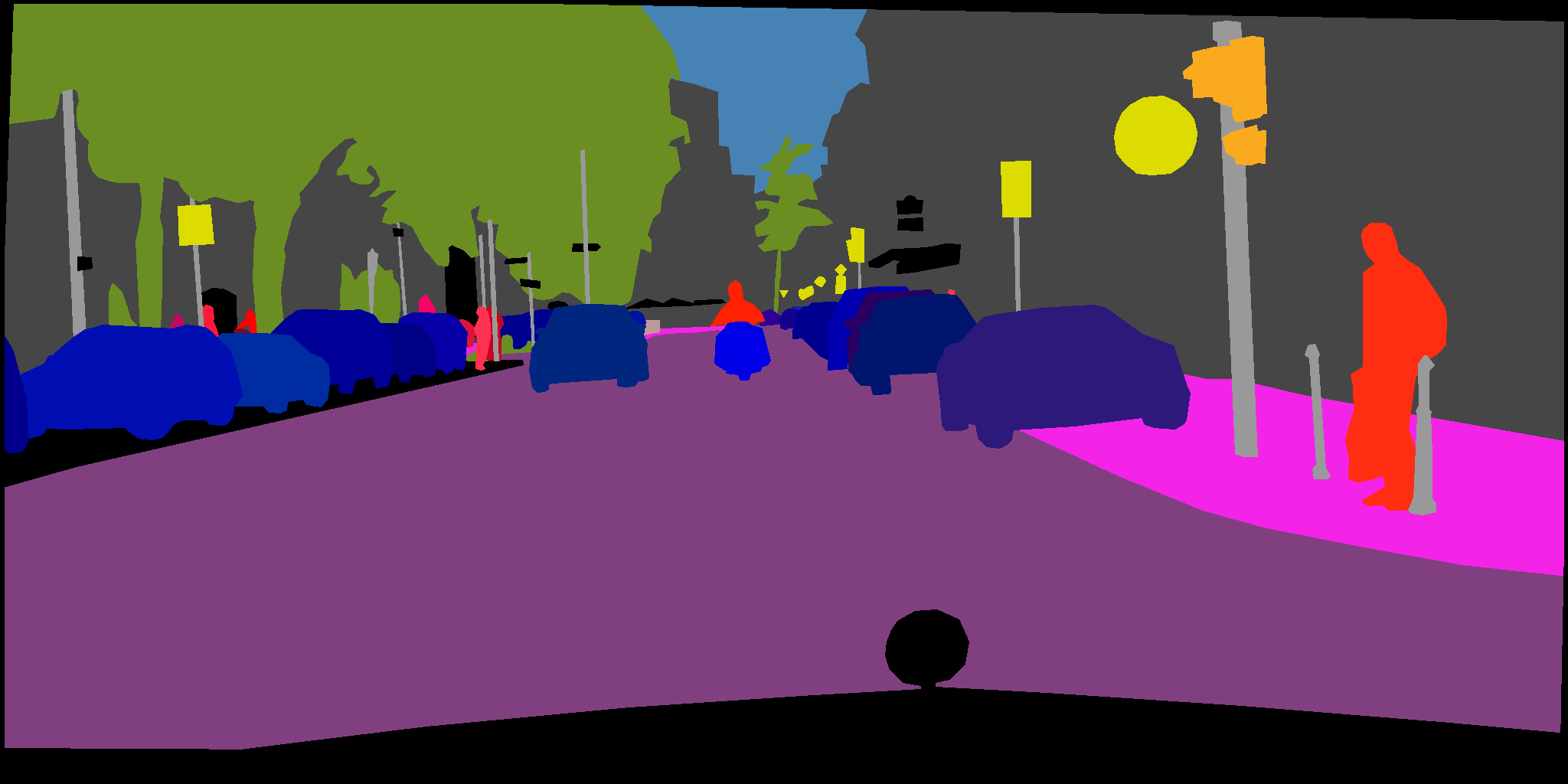}} \hfill
	\subfloat{\includegraphics[width = .249\linewidth]{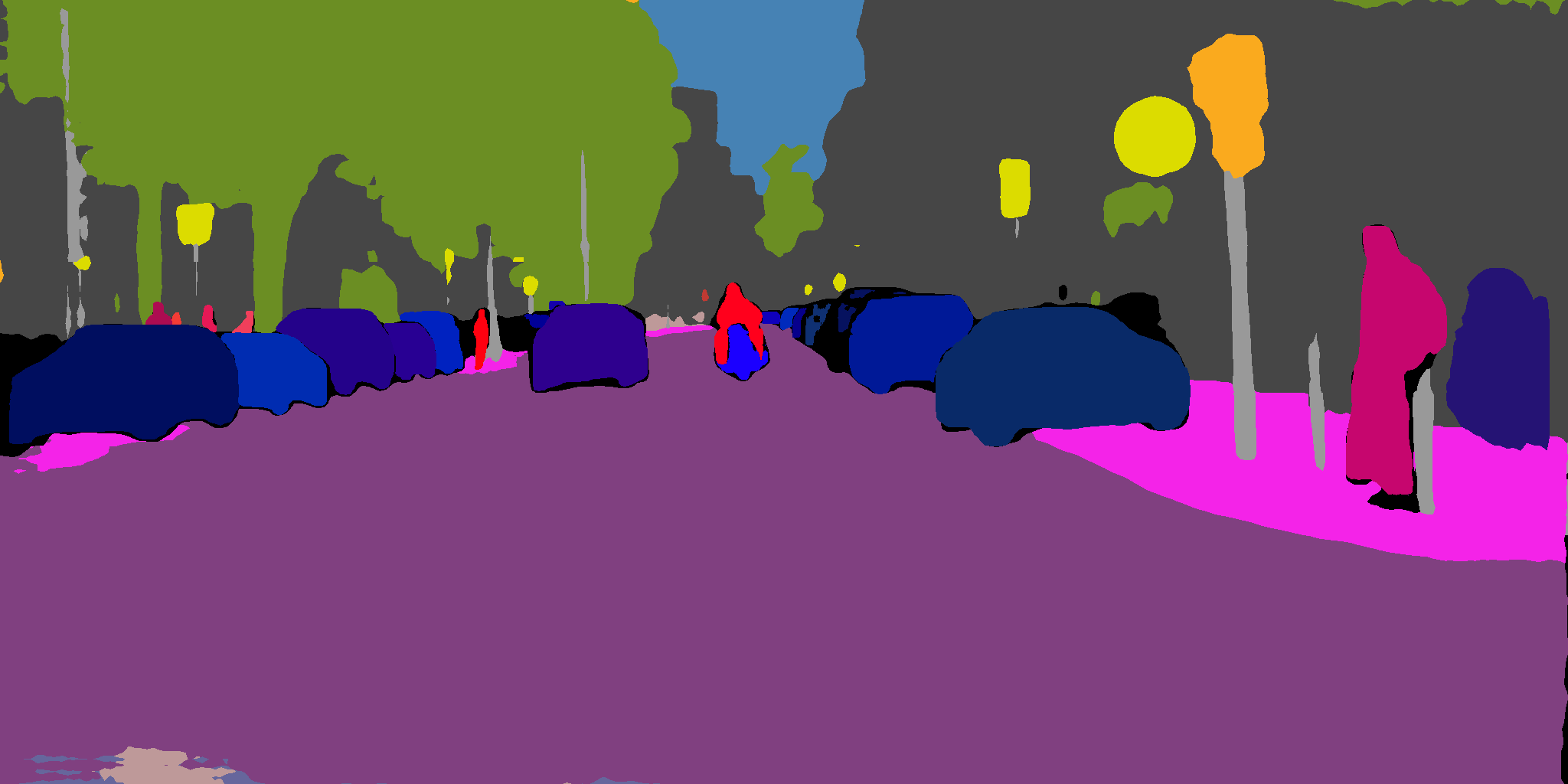}} \hfill
	\subfloat{\includegraphics[width = .249\linewidth]{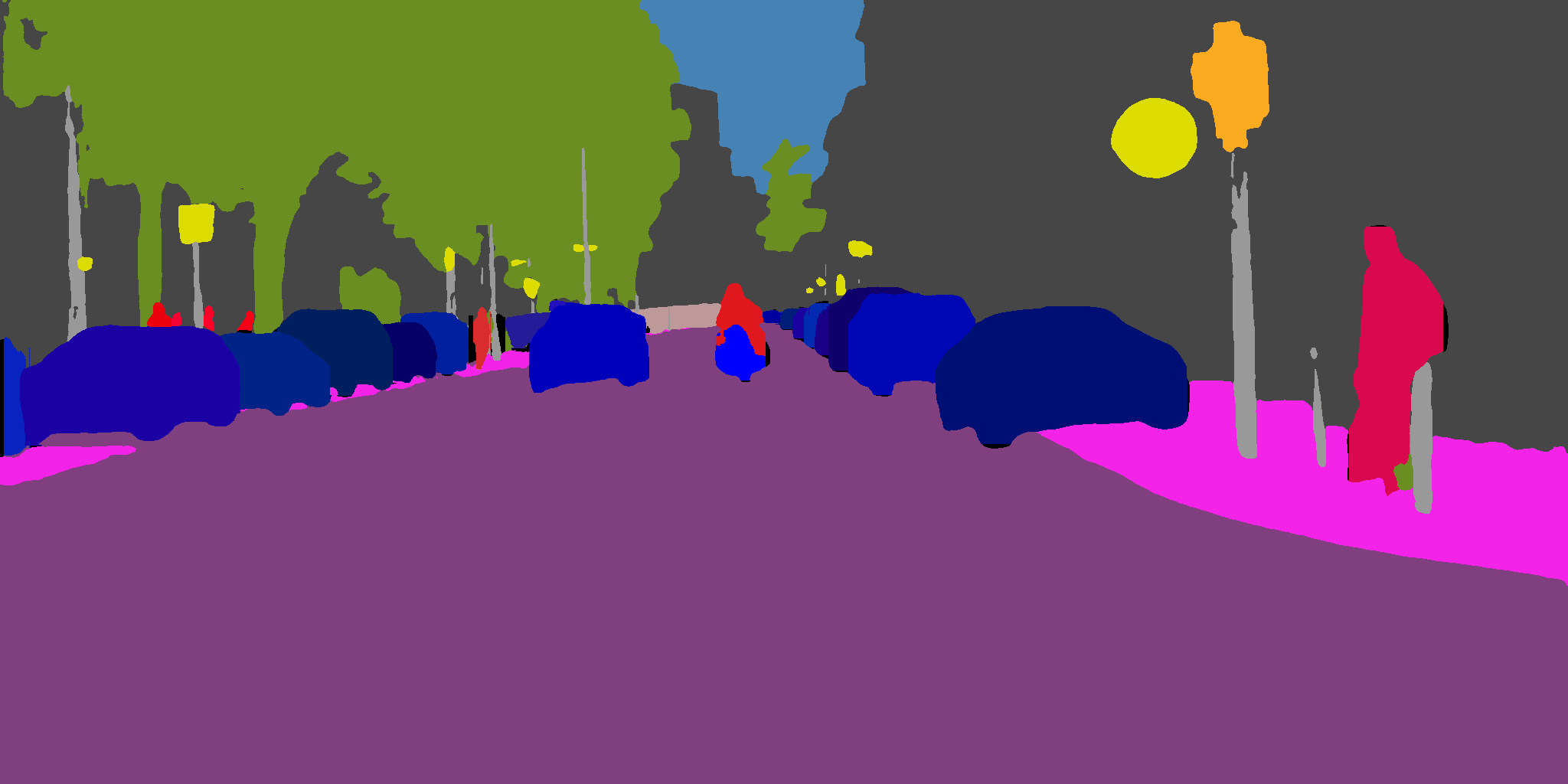}} \\ 
	\vspace{-0.35cm}
	\subfloat{\includegraphics[width = .249\linewidth]{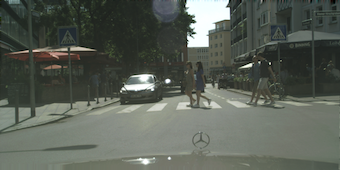}} \hfill
	\subfloat{\includegraphics[width = .249\linewidth]{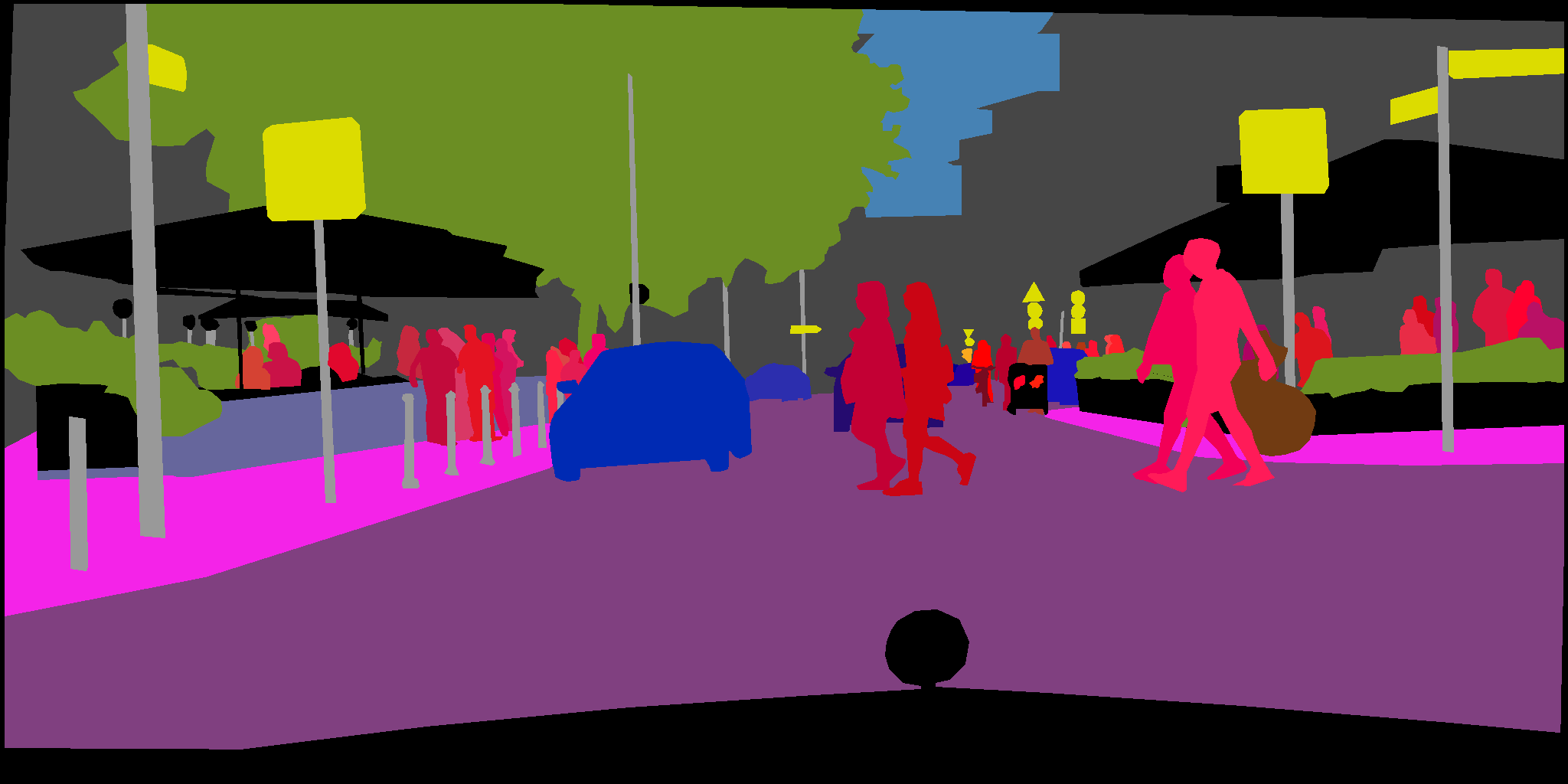}} \hfill
	\subfloat{\includegraphics[width = .249\linewidth]{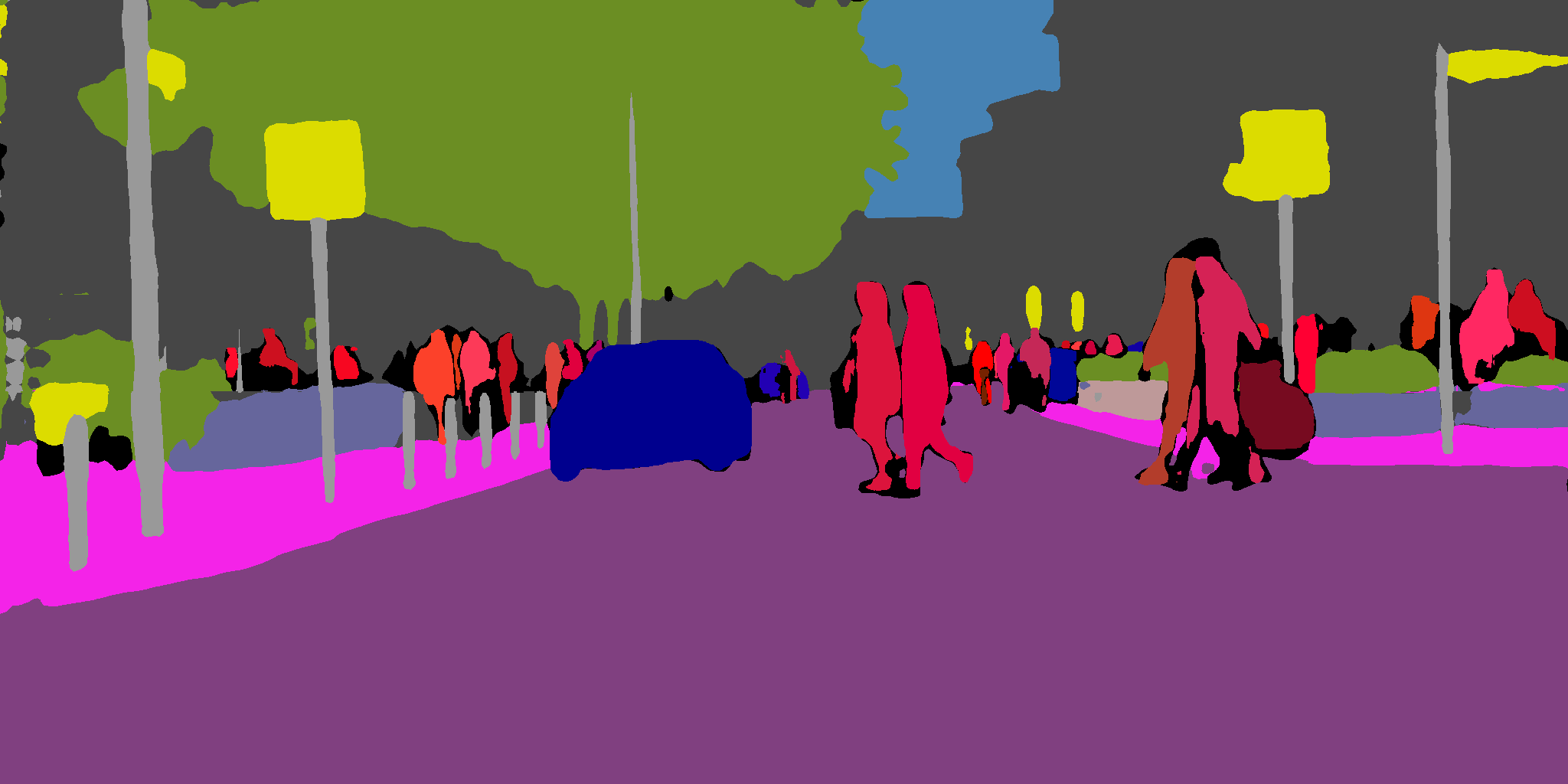}} \hfill
	\subfloat{\includegraphics[width = .249\linewidth]{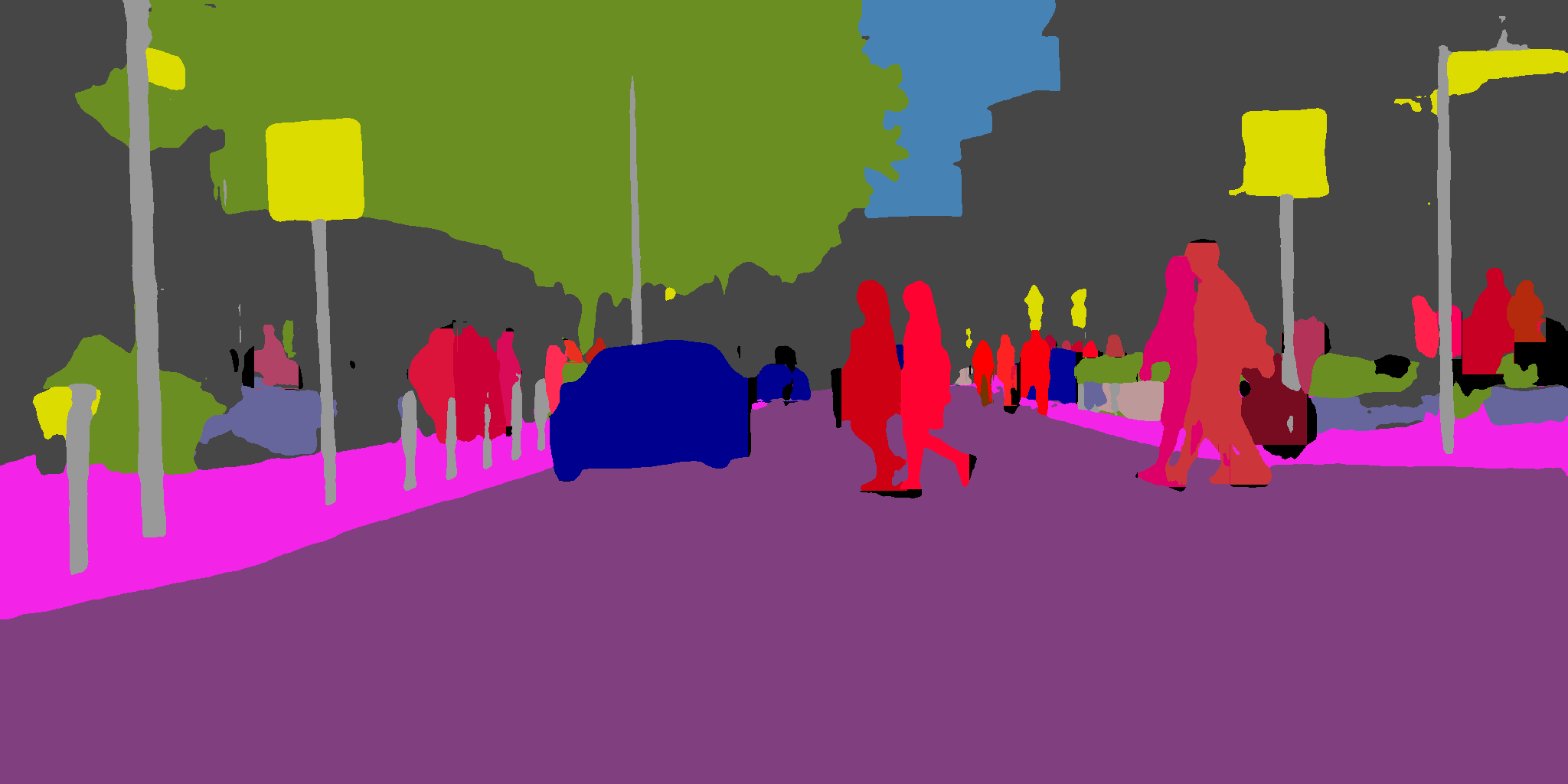}} \\ 	
	\vspace{-0.35cm}
	\subfloat{\includegraphics[width = .249\linewidth]{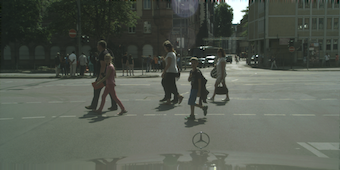}} \hfill
	\subfloat{\includegraphics[width = .249\linewidth]{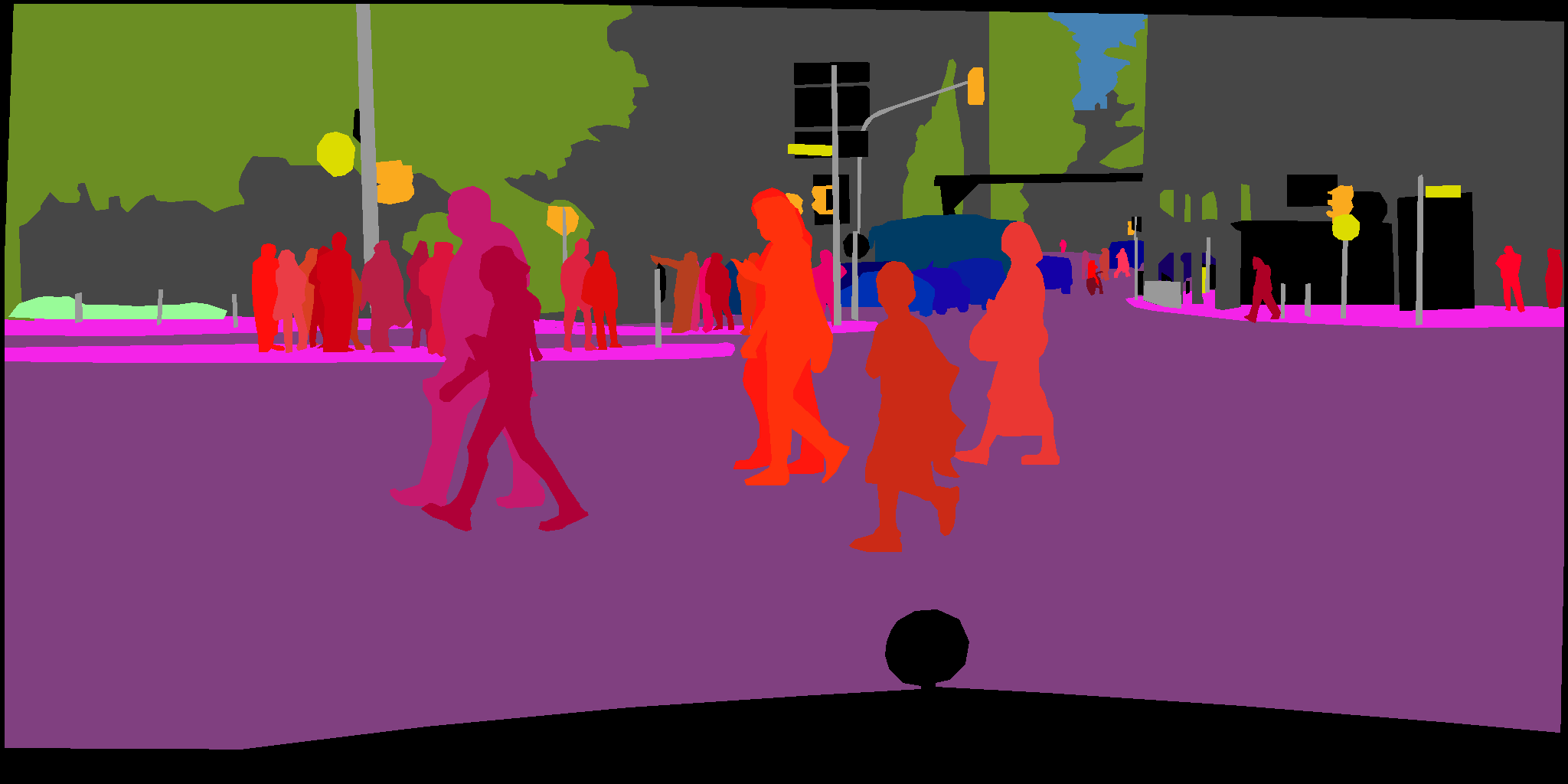}} \hfill
	\subfloat{\includegraphics[width = .249\linewidth]{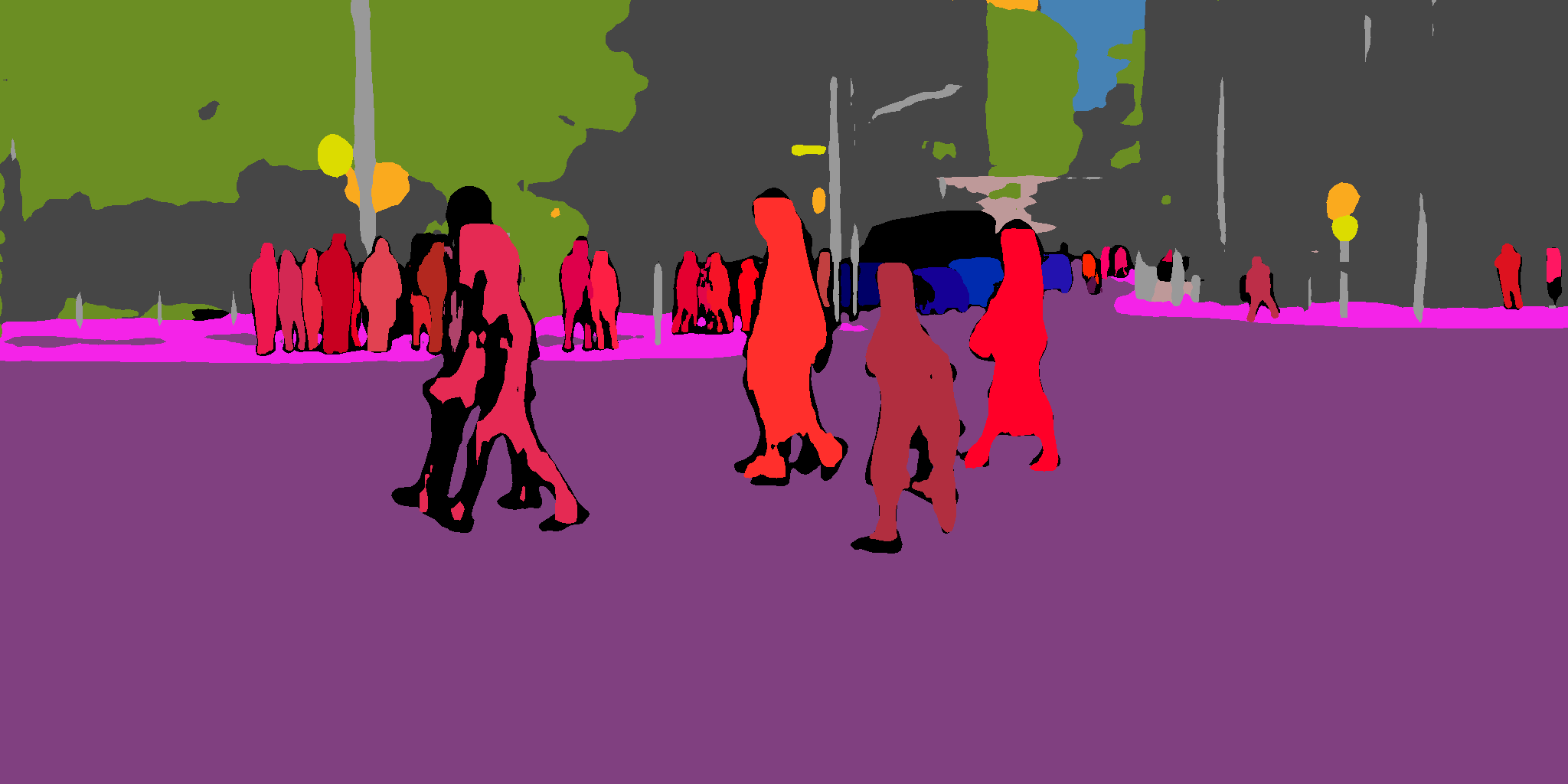}} \hfill
	\subfloat{\includegraphics[width = .249\linewidth]{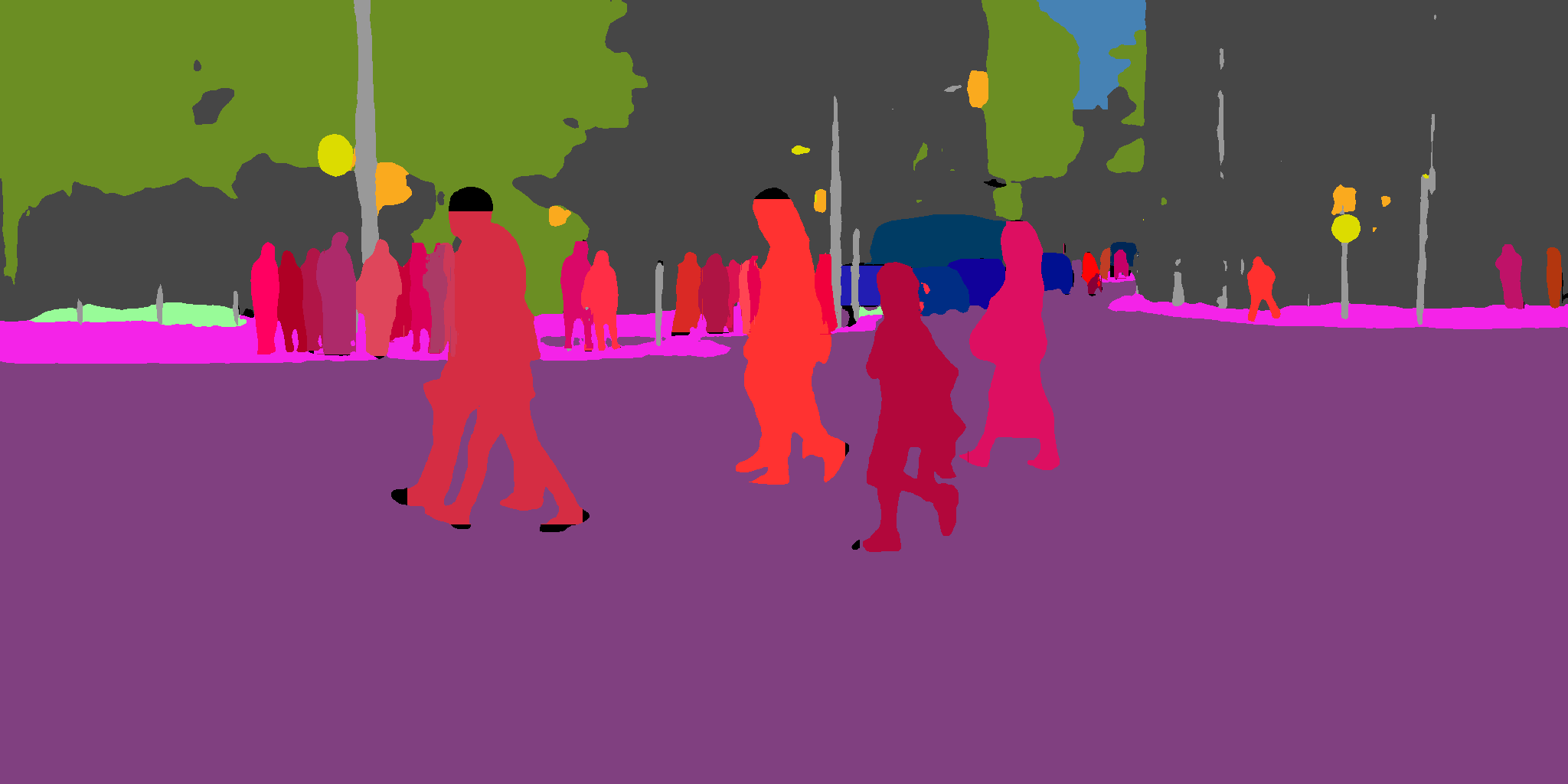}} \\ 
	\vspace{-0.35cm}
	\subfloat[Image]{\includegraphics[width = .249\linewidth]{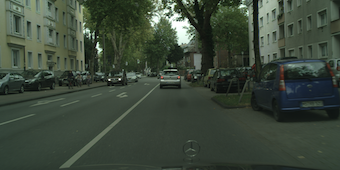}} \hfill
	\subfloat[Ground truth]{\includegraphics[width = .249\linewidth]{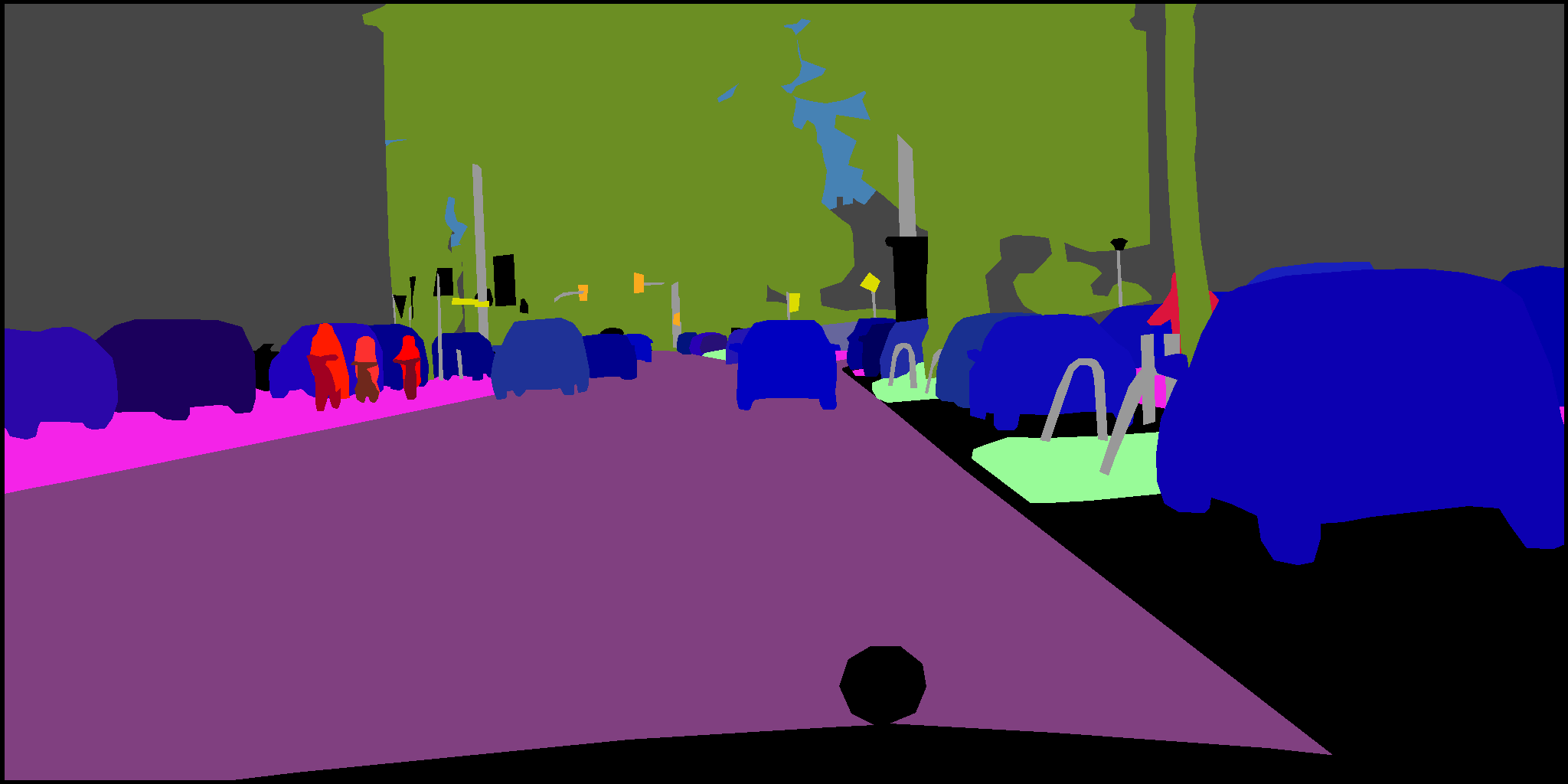}} \hfill
	\subfloat[Combined method~\cite{kirillov2018panoptic}]{\includegraphics[width = .249\linewidth]{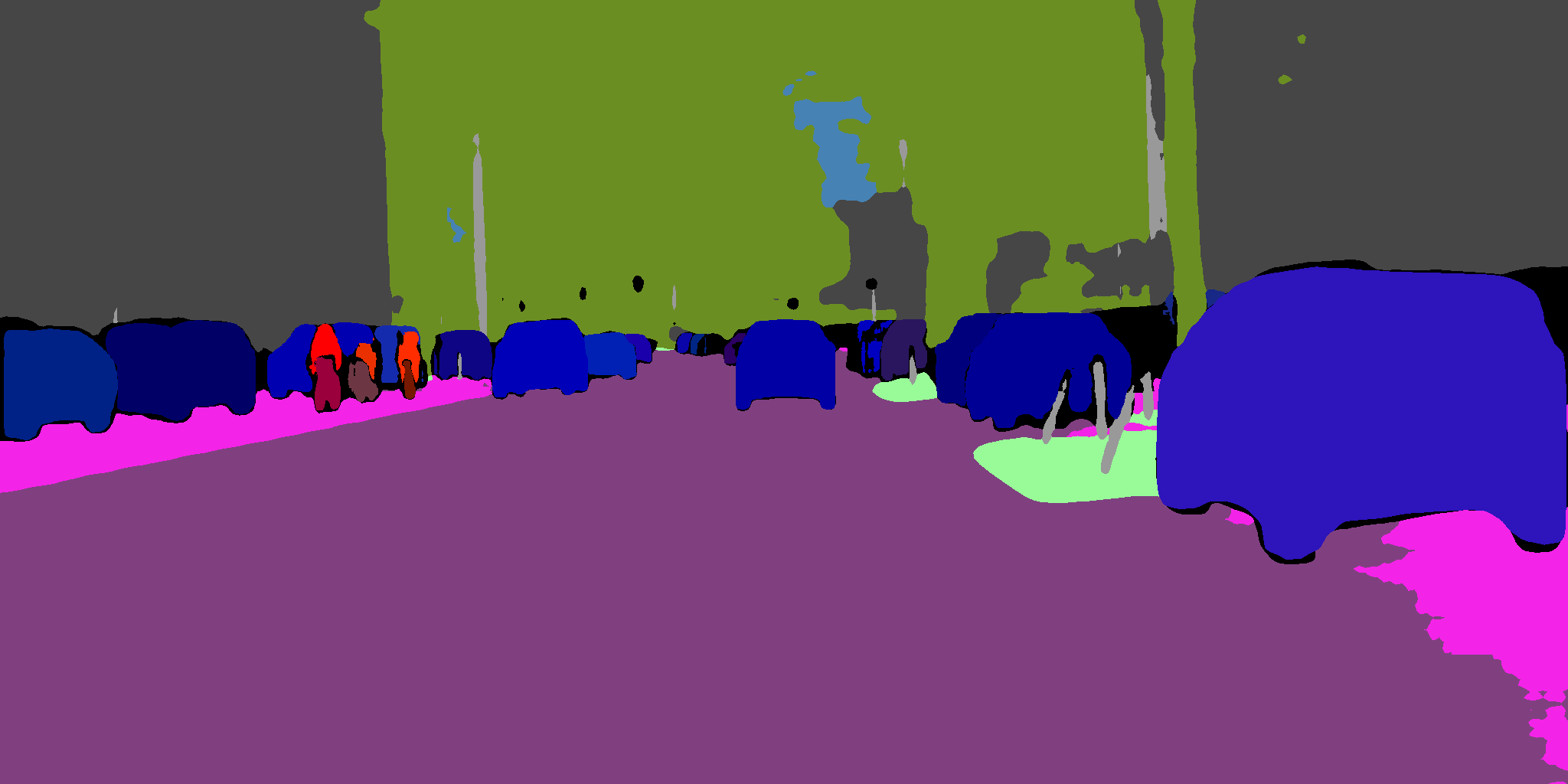}} \hfill
	\subfloat[Ours]{\includegraphics[width = .249\linewidth]{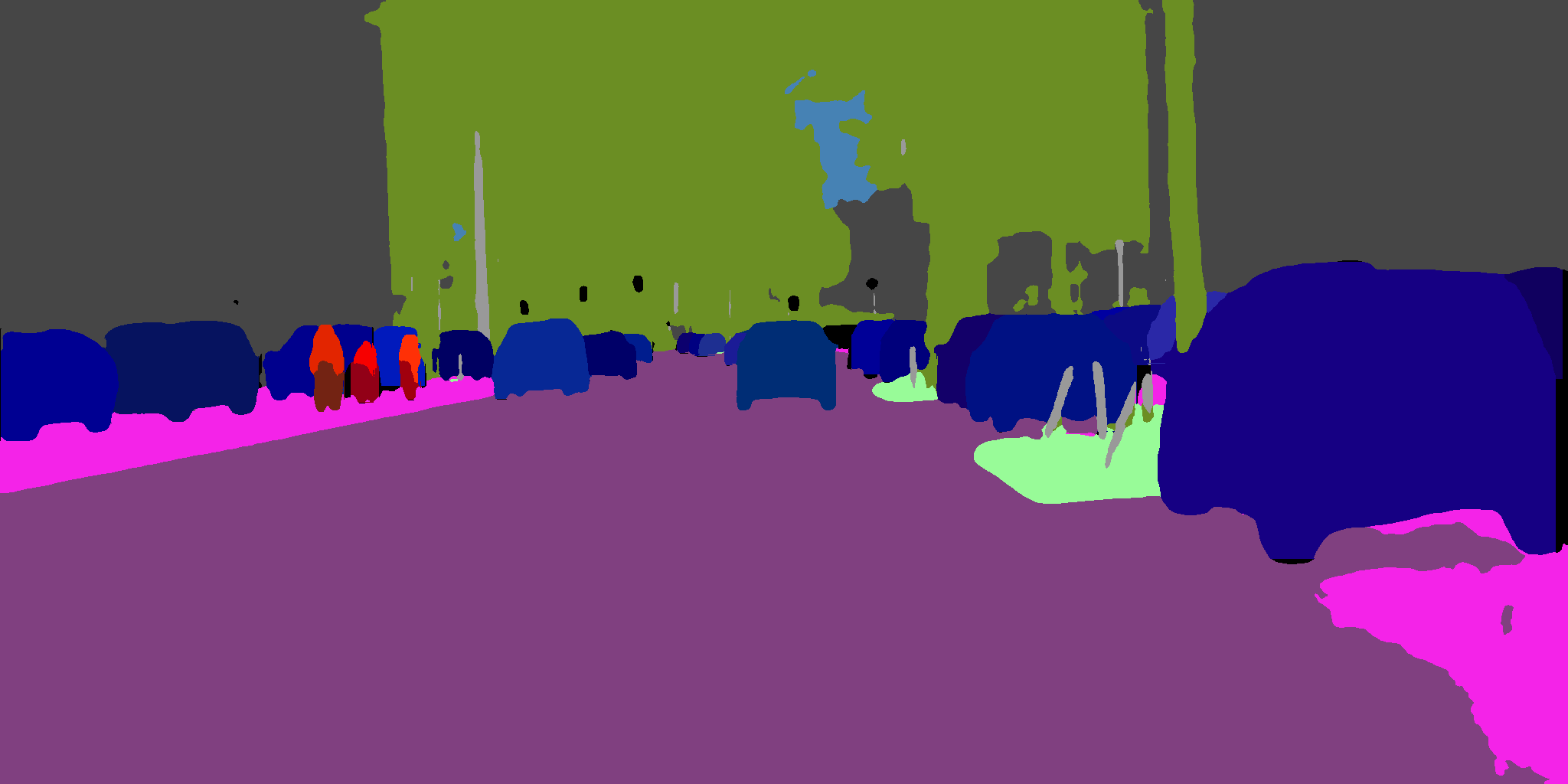}} \\
    \caption{Visual examples of panoptic segmentation on Cityscapes.}
   	\label{fig:results_cityscapes}
\end{figure*}

\begin{figure*}
	\centering	
	\subfloat{\includegraphics[width = .249\linewidth]{imgs/vis/image/ours_2_small.png}} \hfill
	\subfloat{\includegraphics[width = .249\linewidth]{imgs/vis/gt/ours_2.png}} \hfill
	\subfloat{\includegraphics[width = .249\linewidth]{imgs/vis/combined/ours_2.png}} \hfill
	\subfloat{\includegraphics[width = .249\linewidth]{imgs/vis/upsnet/ours_2.png}} \\
	\vspace{-0.35cm}
	\subfloat{\includegraphics[width = .249\linewidth]{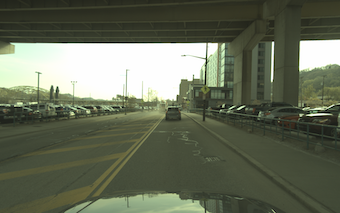}} \hfill
	\subfloat{\includegraphics[width = .249\linewidth]{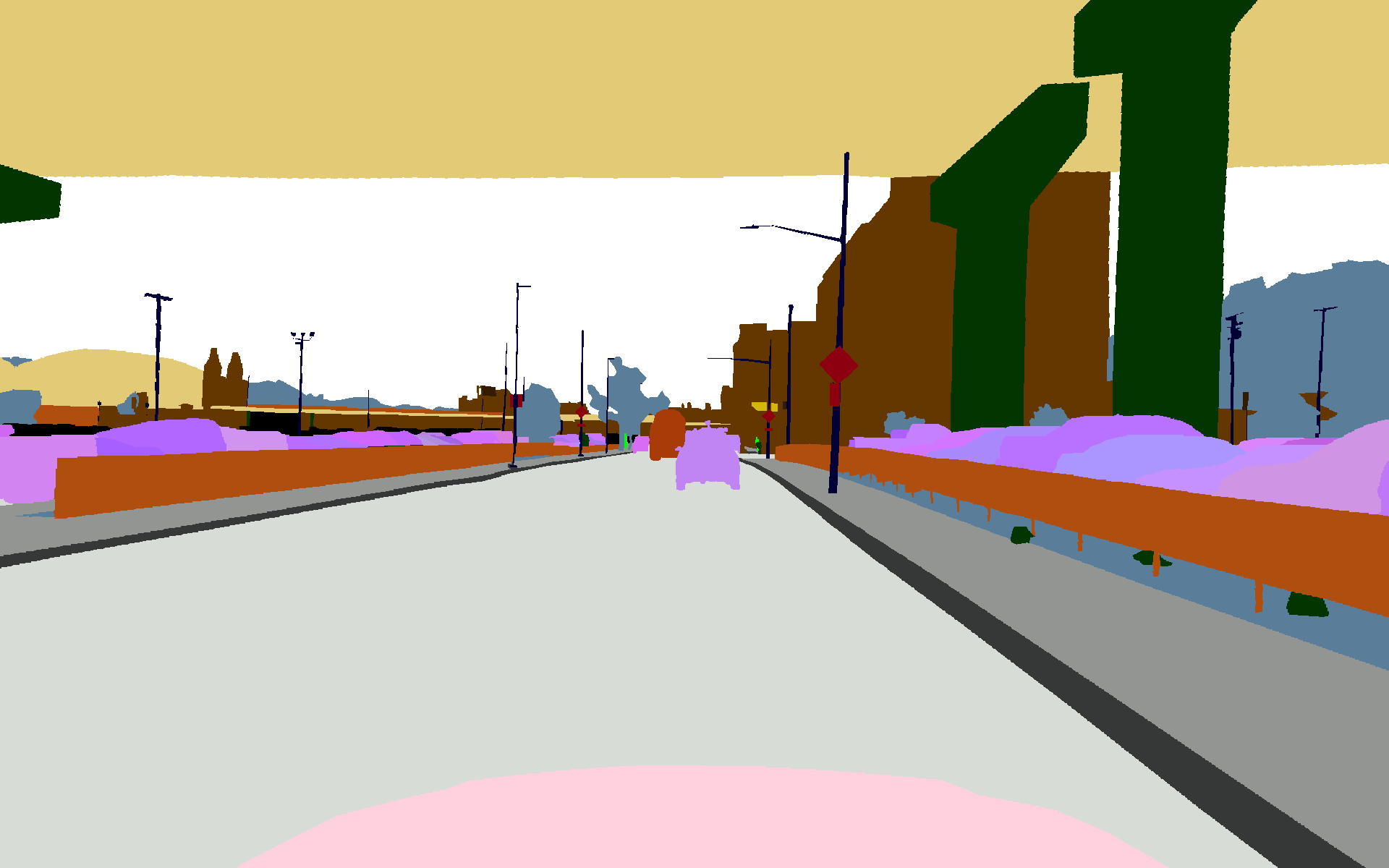}} \hfill
	\subfloat{\includegraphics[width = .249\linewidth]{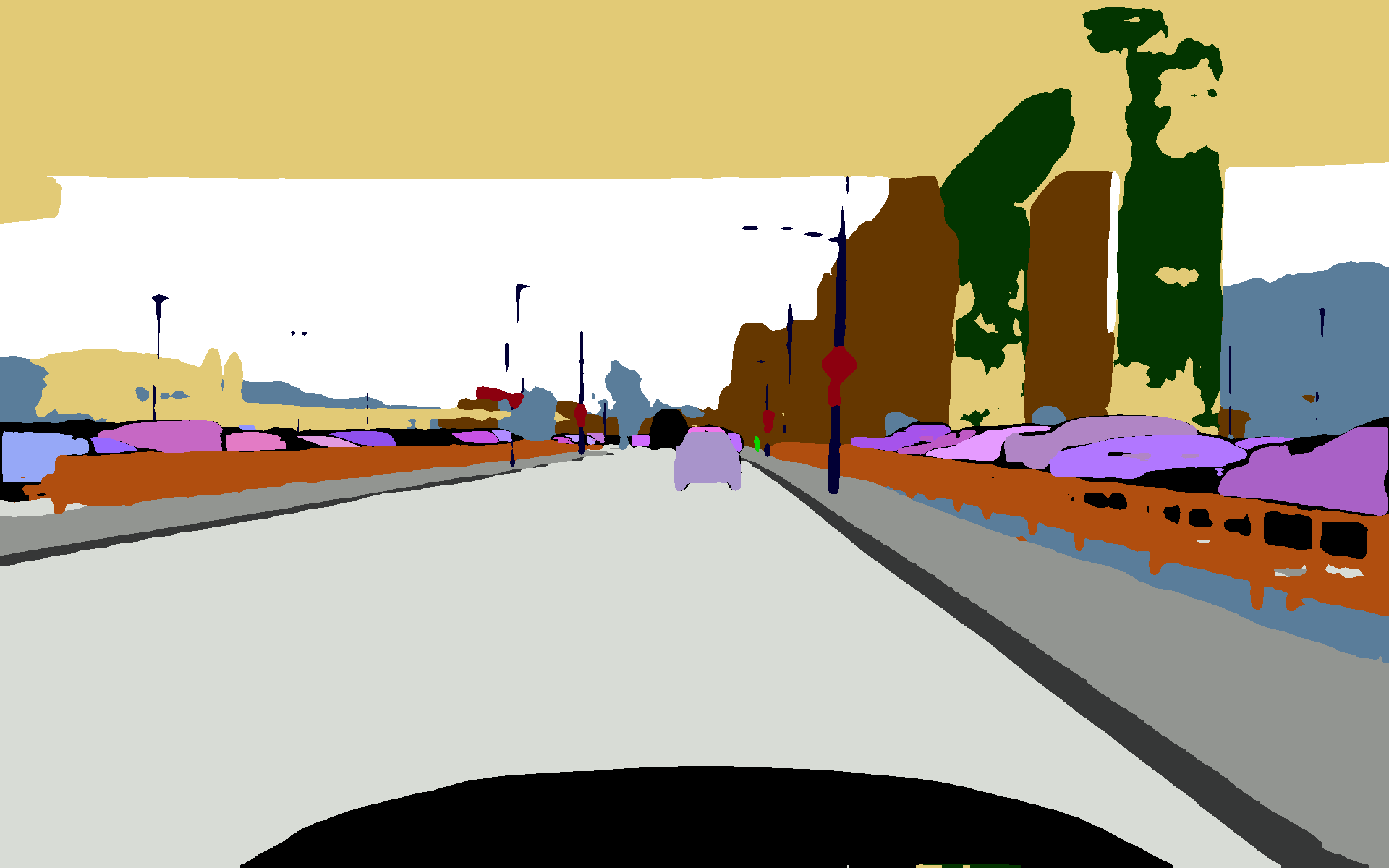}} \hfill
	\subfloat{\includegraphics[width = .249\linewidth]{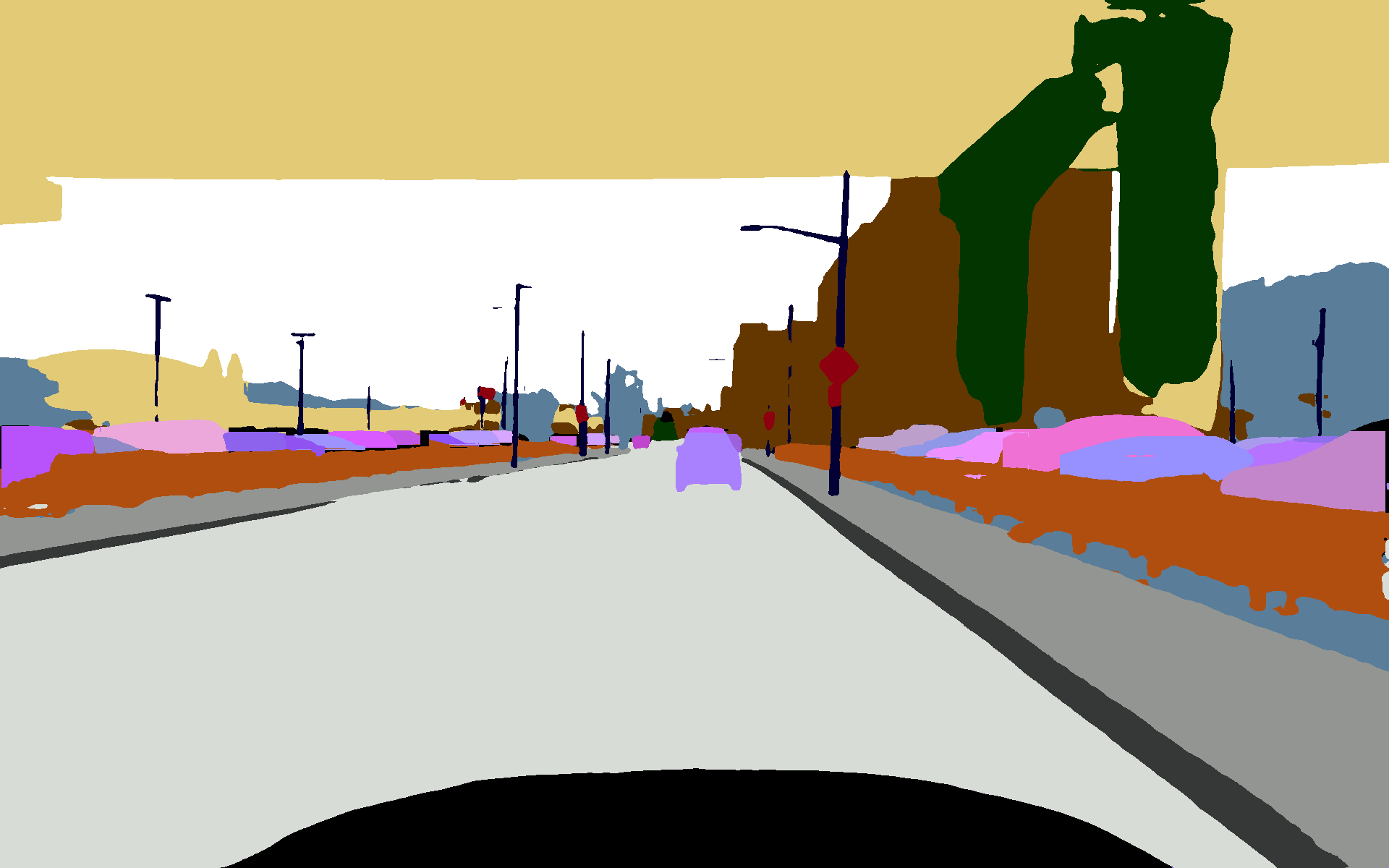}} \\
	\vspace{-0.35cm}
	\subfloat{\includegraphics[width = .249\linewidth]{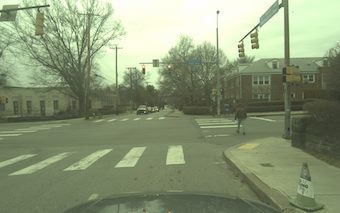}} \hfill
	\subfloat{\includegraphics[width = .249\linewidth]{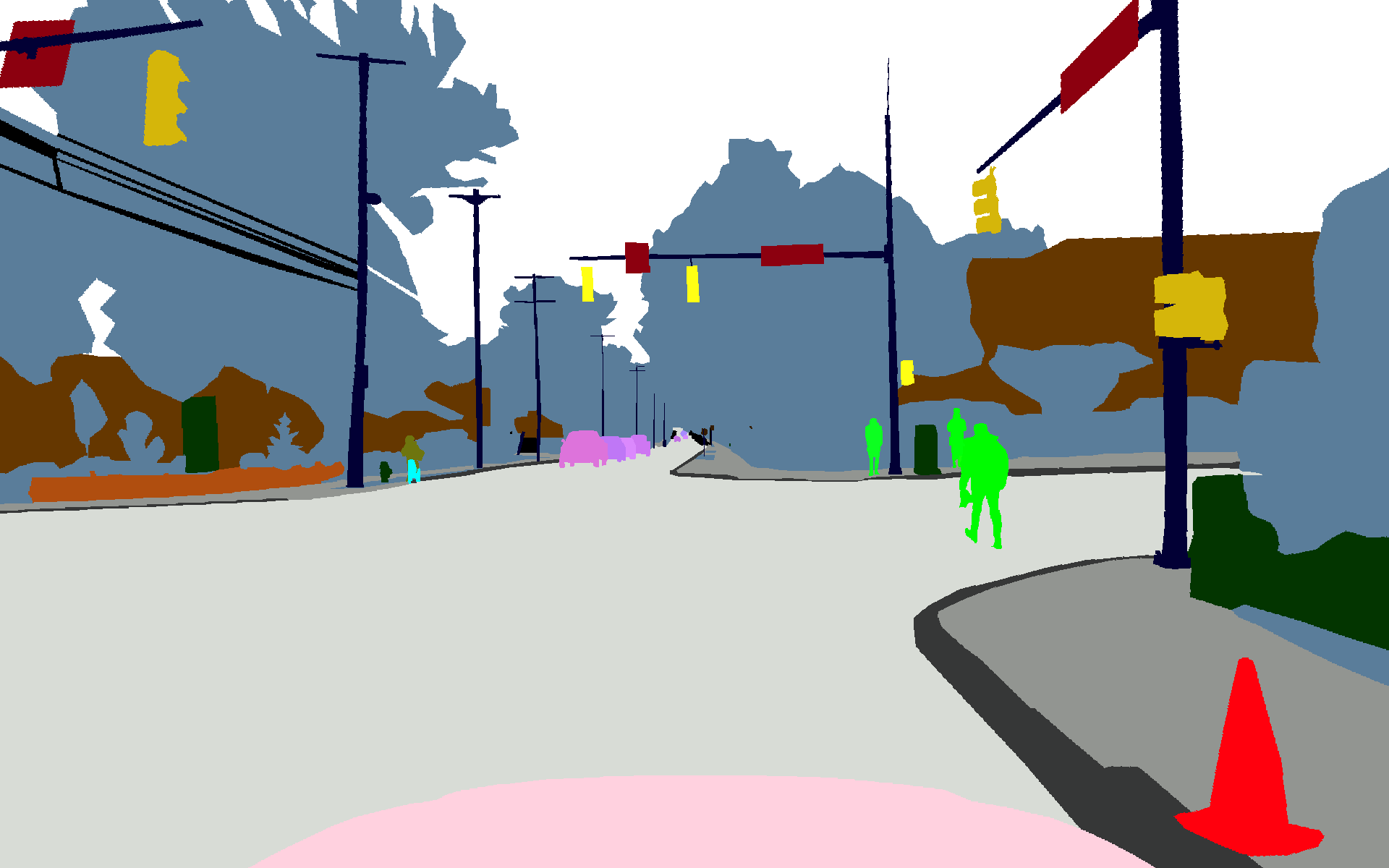}} \hfill
	\subfloat{\includegraphics[width = .249\linewidth]{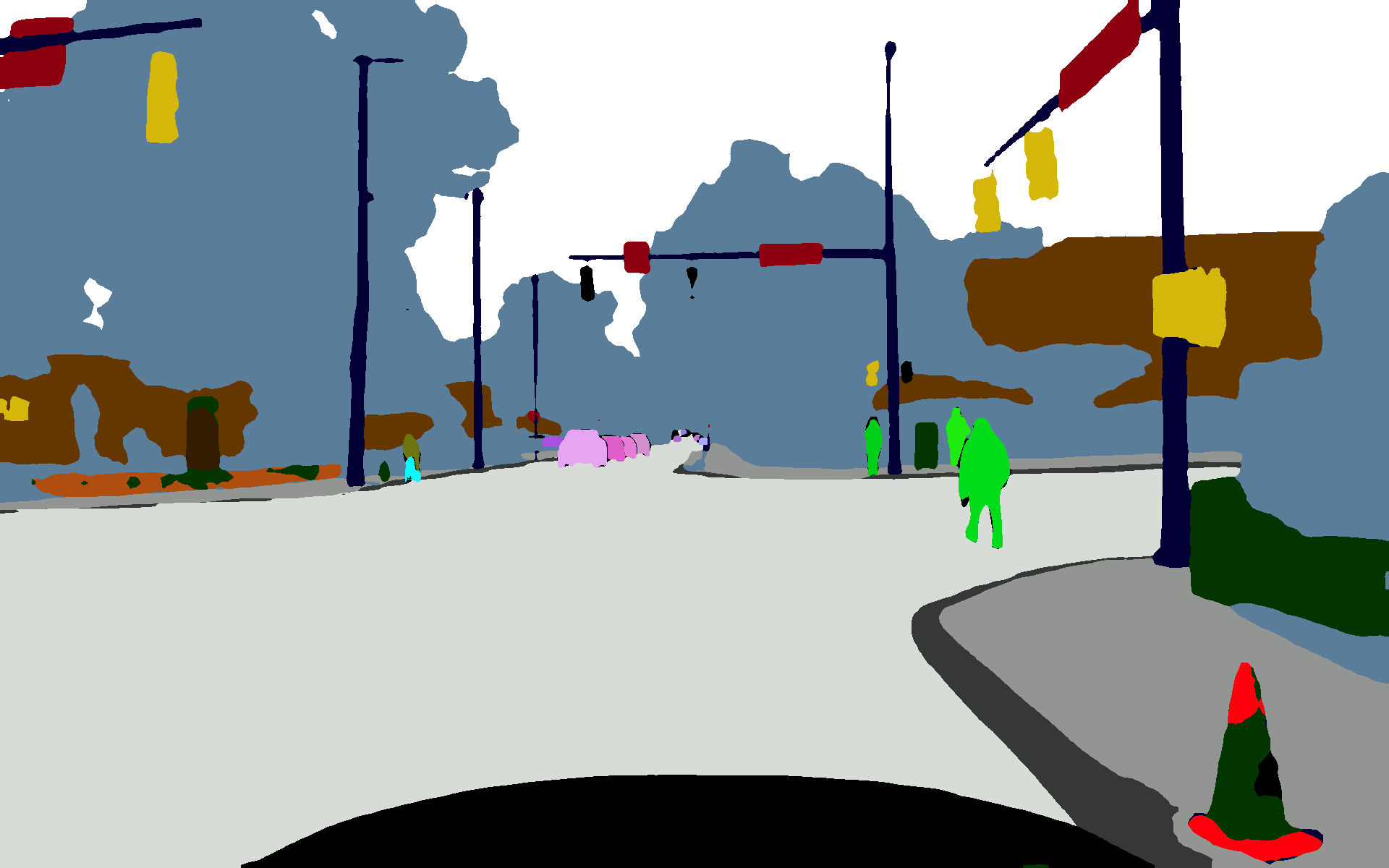}} \hfill
	\subfloat{\includegraphics[width = .249\linewidth]{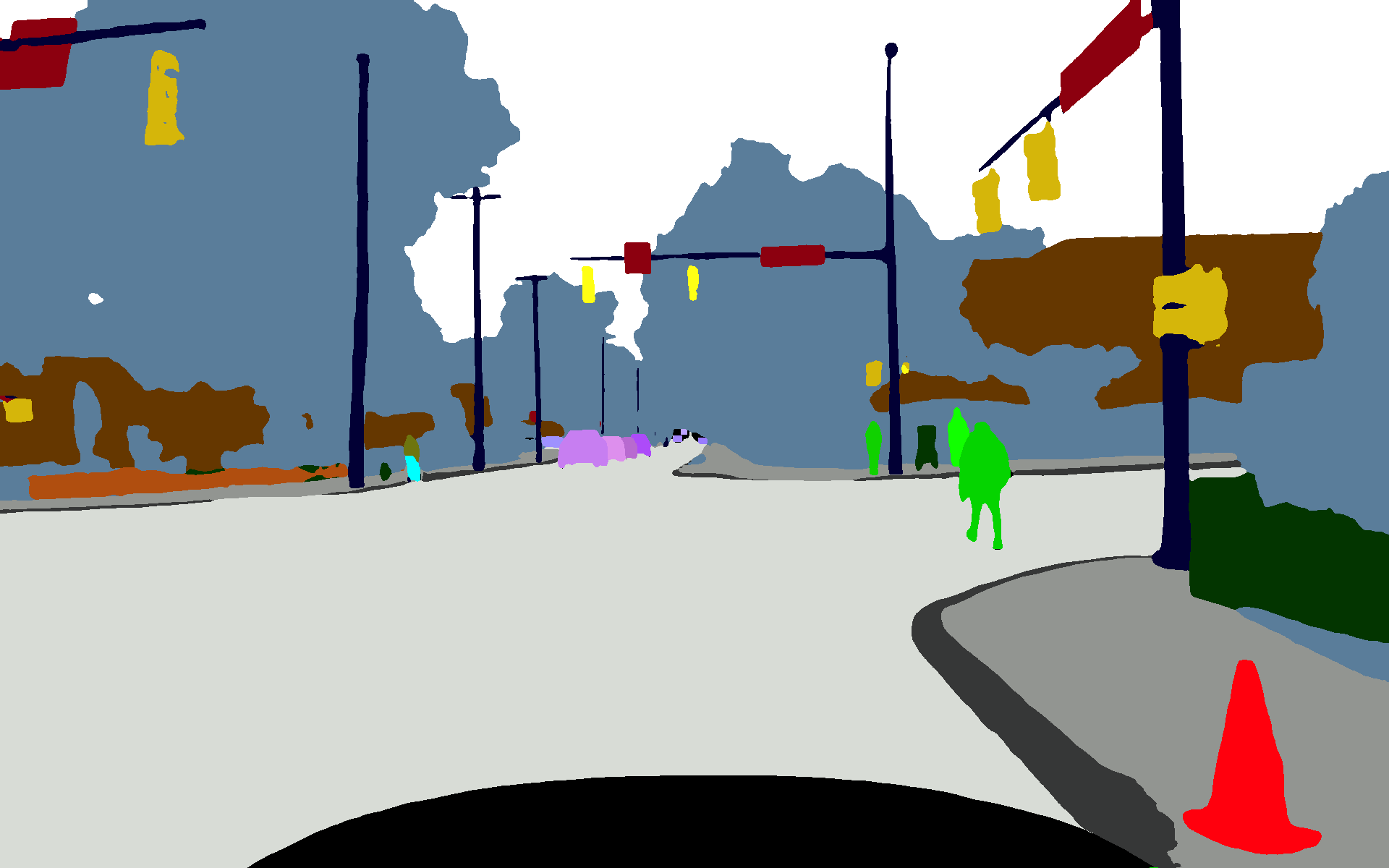}} \\
	\vspace{-0.35cm}
	\subfloat[Image]{\includegraphics[width = .249\linewidth]{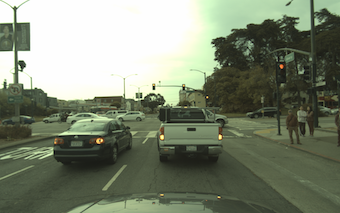}} \hfill
	\subfloat[Ground truth]{\includegraphics[width = .249\linewidth]{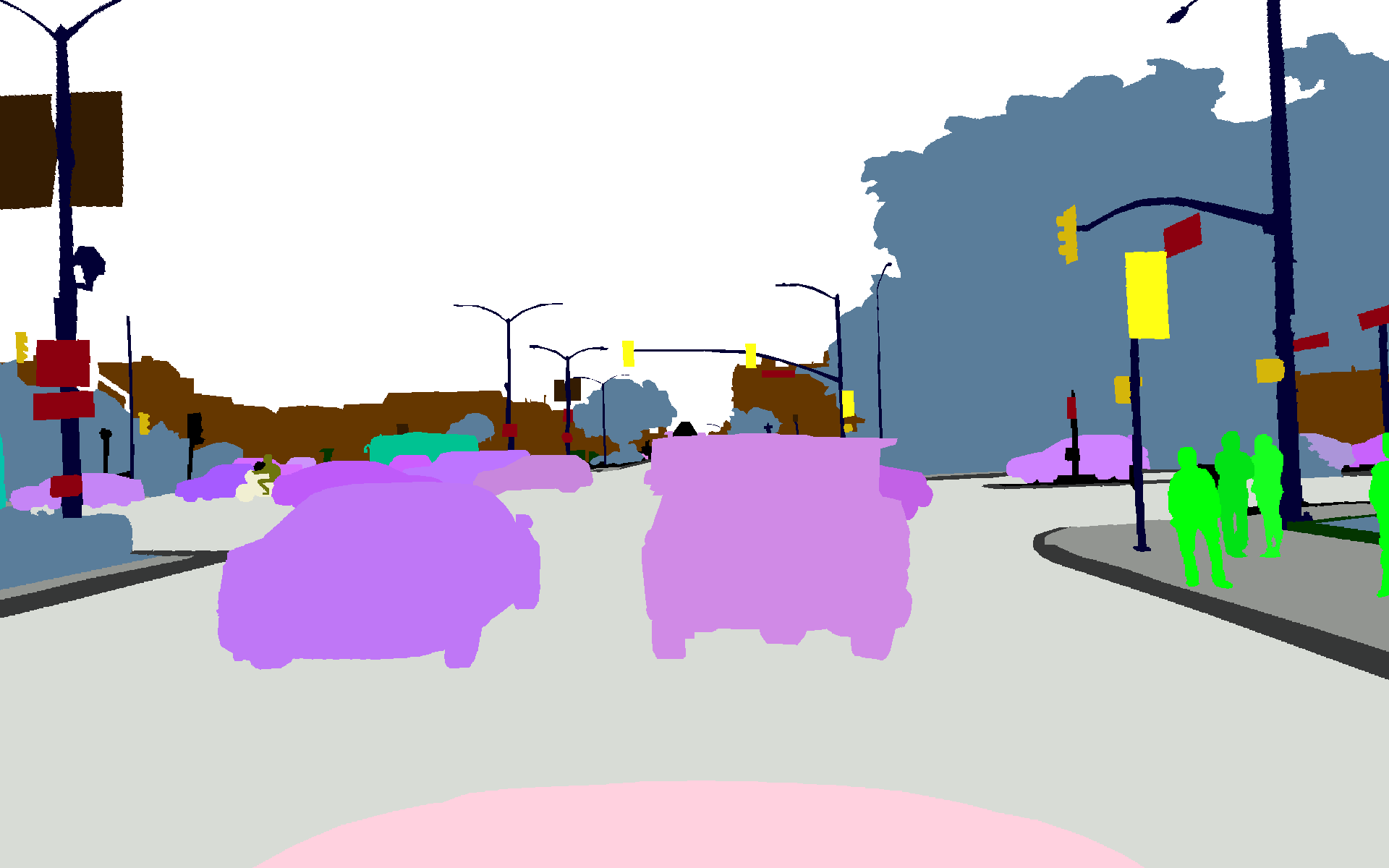}} \hfill
	\subfloat[Combined method~\cite{kirillov2018panoptic}]{\includegraphics[width = .249\linewidth]{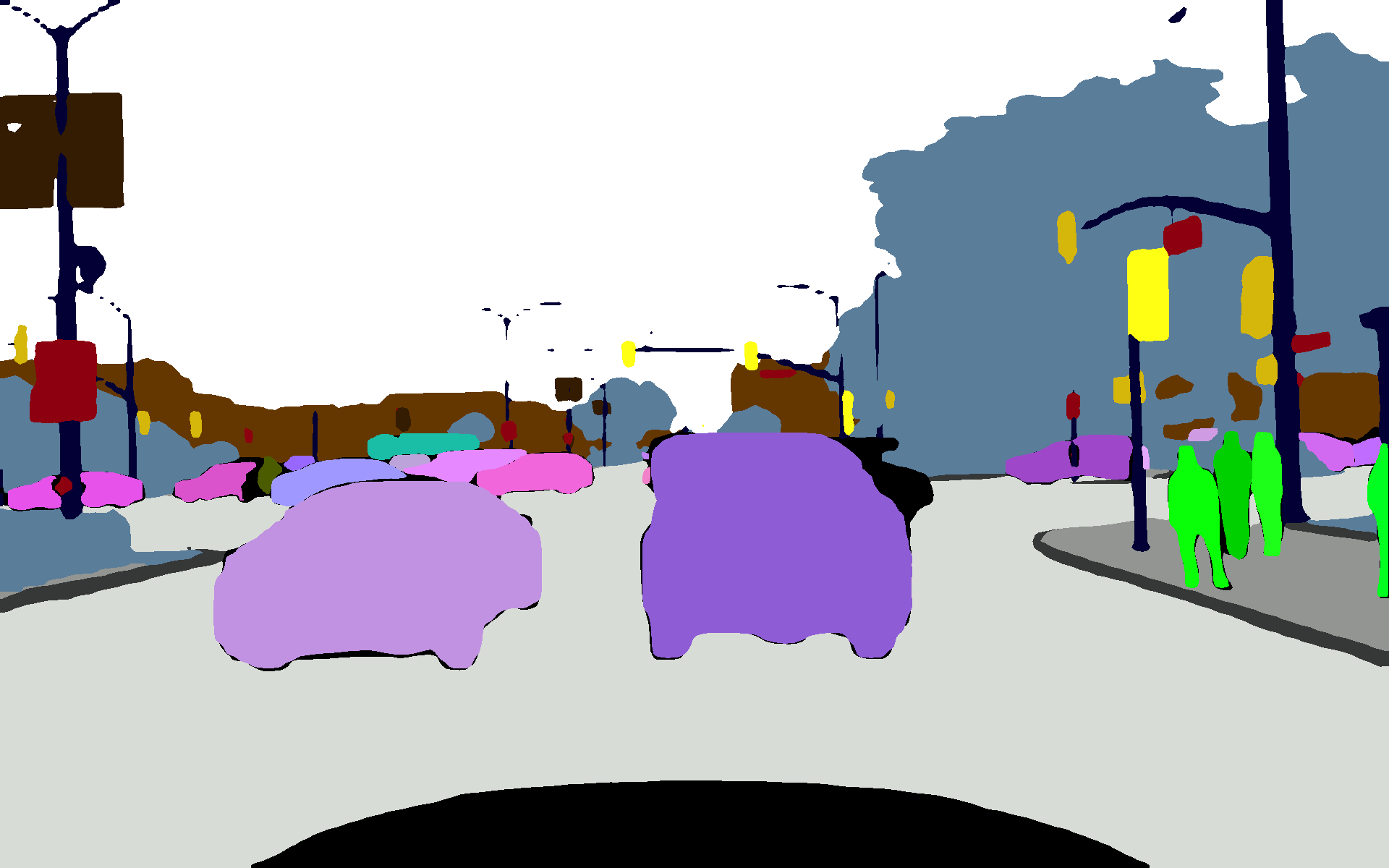}} \hfill
	\subfloat[Ours]{\includegraphics[width = .249\linewidth]{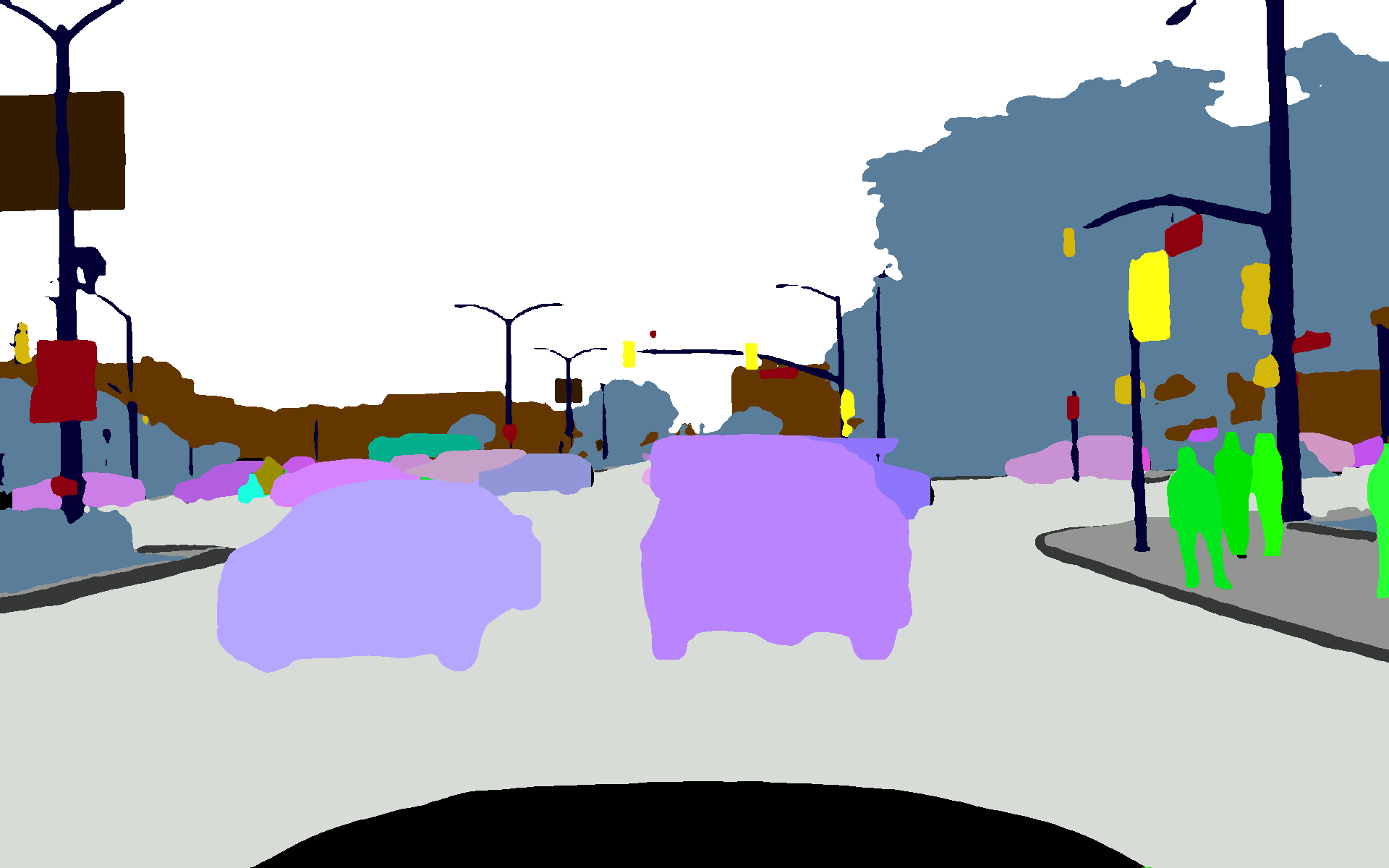}} \\
    \caption{Visual examples of panoptic segmentation on our internal dataset.}
   	\label{fig:results_ours}
\end{figure*}

We show more visual examples of our panoptic segmentation on COCO, Cityscapes and our dataset in Fig. \ref{fig:results_coco}, Fig. \ref{fig:results_cityscapes} and Fig. \ref{fig:results_ours} respectively.

\end{document}